\documentclass{article}

\usepackage[utf8]{inputenc} 
\usepackage[T1]{fontenc}    
\usepackage{booktabs}       
\usepackage{nicefrac}       
\usepackage{microtype}      
\usepackage{times}             

\usepackage{amssymb,amsmath,amsthm,bbm}
\usepackage[margin=1in]{geometry}
\usepackage{verbatim,float,url,dsfont}
\usepackage{graphicx,subfigure,psfrag}
\usepackage{algorithm,algorithmic}
\usepackage{mathtools,enumitem}
\usepackage[colorlinks,citecolor=blue]{hyperref}
\usepackage[utf8]{inputenc} 
\usepackage[T1]{fontenc}    
\usepackage{hyperref}       
\usepackage{url}            
\usepackage{booktabs}       
\usepackage{amsfonts}       
\usepackage{nicefrac}       
\usepackage{microtype}      

\usepackage{times}             
\usepackage{amsthm}

\usepackage{bbm}
\usepackage{verbatim,float,dsfont}
\usepackage{psfrag}
\usepackage{algorithm,algorithmic}
\usepackage{mathtools,enumitem}
\mathtoolsset{showonlyrefs}
\usepackage{tcolorbox}
\usepackage{breqn,lipsum}
\usepackage{thmtools,thm-restate}
\usepackage{graphicx,stackengine}
\usepackage{array}
\usepackage{caption}
\usepackage{boldline}
\usepackage{amssymb}
\usepackage{multirow}

\usepackage{graphicx}
\usepackage[nocomma]{optidef}
\usepackage{mdframed}

\usepackage{wrapfig}

\usepackage{tikz}
\usetikzlibrary{shapes,arrows}

\usetikzlibrary{positioning, quotes}
\usepackage{comment}

\newtheorem{theorem}{Theorem}
\newtheorem{lemma}[theorem]{Lemma}

\newtheorem{proposition}[theorem]{Proposition}
\newtheorem{remark}[theorem]{Remark}
\newtheorem{definition}[theorem]{Definition}

\newtheorem{example}[theorem]{Example}

\def\shownotes{1}  
\ifnum\shownotes=1
\newcommand{\authnote}[2]{$\ll$\textsf{\footnotesize #1 notes: #2}$\gg$}
\else
\newcommand{\authnote}[2]{}
\fi

\usepackage{accents}
\newcommand{\ubar}[1]{\underaccent{\bar}{#1}}

\makeatletter
\newcommand*\rel@kern[1]{\kern#1\dimexpr\macc@kerna}
\newcommand*\widebar[1]{%
  \begingroup
  \def\mathaccent##1##2{%
    \rel@kern{0.8}%
    \overline{\rel@kern{-0.8}\macc@nucleus\rel@kern{0.2}}%
    \rel@kern{-0.2}%
  }%
  \macc@depth\@ne
  \let\math@bgroup\@empty \let\math@egroup\macc@set@skewchar
  \mathsurround\z@ \frozen@everymath{\mathgroup\macc@group\relax}%
  \macc@set@skewchar\relax
  \let\mathaccentV\macc@nested@a
  \macc@nested@a\relax111{#1}%
  \endgroup
}
\makeatother

\newcommand{\argmin}{\mathop{\mathrm{argmin}}}

\makeatletter
\newcommand{\setword}[2]{%
  \phantomsection
  #1\def\@currentlabel{\unexpanded{#1}}\label{#2}%
}
\makeatother

\def\row{\mathrm{row}}

\def\sign{\mathrm{sign}}

\def\diag{\mathrm{diag}}

\def\cA{\mathcal{A}}

\def\cD{\mathcal{D}}

\def\cP{\mathcal{P}}
\def\cQ{\mathcal{Q}}
\def\cR{\mathcal{R}}
\def\cS{\mathcal{S}}

\def\cW{\mathcal{W}}

\def\bs{\ensuremath\boldsymbol}
\def\code#1{\texttt{#1}}

\newcommand{\grad}{\nabla}
\newcommand{\lamda}{\lambda}

\usepackage[numbers,sort&compress]{natbib}

\title{Optimal Dynamic Regret in Proper Online Learning with Strongly Convex Losses and Beyond}  

\author{Dheeraj Baby \\dheeraj@ucsb.edu \and Yu-Xiang Wang \\yuxiangw@cs.ucsb.edu}

\date{Dept. of Computer Science \\ UC Santa Barbara}

\begin{document}

\maketitle

\begin{abstract}
We study the framework of \emph{universal dynamic regret} minimization with \emph{strongly convex} losses. We answer an open problem in \citep{improperDynamic} by showing that in a \emph{proper learning} setup, Strongly Adaptive algorithms can achieve the near optimal dynamic regret of $\tilde O(d^{1/3} n^{1/3}\text{TV}[u_{1:n}]^{2/3} \vee d)$ against any comparator sequence $u_1,\ldots,u_n$ \emph{simultaneously}, where $n$ is the time horizon and $\text{TV}[u_{1:n}]$ is the Total Variation of comparator. These results are facilitated by exploiting a number of \emph{new} structures imposed by the KKT conditions that were not considered in \citep{improperDynamic} which also lead to other improvements over their results such as: (a) handling non-smooth losses  and (b) improving the dimension dependence on regret. Further, we also derive near optimal dynamic regret rates for the special case of proper online learning with exp-concave losses and an $L_\infty$ constrained decision set.
\end{abstract}

\section{Introduction}
Online Convex Optimization (OCO) \citep{hazan2016introduction} is a powerful learning paradigm for the task of sequential decision making. It is modelled as an interactive game between a learner and adversary as follows: For each time step $ t \in [n] := \{ 1,2,\ldots,n\}$, the learner plays a point $\bs x_t \in \mathbb{R}^d$. Then the adversary reveals a convex loss $f_t:\mathbb{R}^d \rightarrow \mathbb{R}$. A common objective in online learning is to minimize the learner's static regret against a convex set of benchmark points $\cD \subset \mathbb{R}^d$: $R_{\text{static}} = \sum_{t=1}^n f_t(\bs x_t) - \inf_{\bs w \in \cD} \sum_{t=1}^n f_t(\bs w)$.

However, the notion of static regret is not befitting to applications where the environment is non-stationary. To alleviate this issue, one may aim to control the dynamic regret against a sequence of comparators in $\cD$ (the comparator sequence may be potentially unknown to the learner):
\begin{align}
       R_{\text{dynamic}}(\bs w_{1:n}) = \sum_{t=1}^n f_t(\bs x_t) - \sum_{t=1}^n f_t(\bs w_t),
\end{align}
where we use the shorthand $\bs w_{1:n} := \{\bs w_1,\ldots,\bs w_n \}$. Here each $\bs w_t \in \cD$. Dynamic regret rates are usually expressed in terms of the time horizon $n$ and a regularity measure aka path length that captures the smoothness of the comparator sequence. For example, in \citep{zhang2018adaptive} a regularity measure $V_n(\bs w_{1:n}) = \sum_{t=2}^n \| \bs w_t - \bs w_{t-1}\|_2$ is defined. They propose an algorithm that attains a (near) optimal dynamic regret of $\tilde O(\sqrt{n(1+V_n(\bs w_{1:n}))})$ when the losses $f_t$ are convex ($\tilde O$ hides factors of $\log n$.) Such dynamic regret rates are sometimes referred as \emph{universal dynamic regret} rates as they are applicable to any comparator sequence $\bs w_{1:n}$.

However, optimal dynamic regret rates in terms of path length of the arbitrary comparator sequence when the loss functions have extra curvature properties such as strong convexity or exp-concavity, have been long eluded in the literature until a recent breakthrough by \citep{improperDynamic}. They define a path length in terms of the Total Variation (TV) of the comparator sequence as: $\text{TV}(\bs w_{1:n}) = \sum_{t=2}^n \| \bs w_t - \bs w_{t-1}\|_1$. They show that when the losses are strongly convex / exp concave and gradient Lipschitz, a Strongly Adaptive (SA) online learner (\citep{hazan2007adaptive,daniely2015strongly}) can attain a (near) optimal dynamic regret rate of $\tilde O^*(n^{1/3}C_n^{2/3} \vee 1)$\footnote{$\tilde O^*$ hides the dependence of $d$ and $\log n$; $a \vee b := \max\{ a, b\}$.} against all sequences with $\text{TV}(\bs w_{1:n}) \le C_n$ where $C_n$ is a quantity that may be unknown to the learner. However, this rate is attained using an improper SA algorithm whose decisions can lie outside $\cD$. A question that was left open was whether improper learning is strictly necessary to achieve the optimal rates for exp-concave optimization. In this work, we answer this in the negative by showing that a proper version of the SA algorithms can attain the optimal (modulo log factors and dimension dependencies) dynamic regret rates whenever the losses are strongly convex. 

We summarize our main contributions below.
\begin{itemize}
    \item We provide a new analysis that extends the results of \citep{improperDynamic} to proper strongly convex online learning to attain the near \emph{optimal} dynamic regret rate of $\tilde O(d^{1/3}n^{1/3}C_n^{2/3} \vee d)$ for Strongly Adaptive methods (see Corollary \ref{cor:sc}). In contrast to \citep{improperDynamic}, our results imply an important conclusion that improper learning is \emph{not strictly necessary} for attaining such fast rates with general strongly convex losses. To the best of our knowledge, this is the \emph{first} result that achieves near optimal dynamic regret in a setting of proper learning under strongly convex losses.
    
    \item For exp-concave losses, we prove an analogous result that Strongly Adaptive algorithms can attain a near optimal dynamic regret of $\tilde O^*((n^{1/3}C_n^{2/3} \vee 1))$ in the special case of $L_\infty$ (box) constrained decision set, $\cD =\{\bs x \in \mathbb{R}^d: \| \bs x\|_\infty \le B \}$ (see Theorem \ref{thm:ec-d}). 
    
    \item To facilitate these results we discover and exploit a number of new structures imposed by the KKT conditions that were not considered in \citep{improperDynamic}, which could be of independent interest.
    
\end{itemize}

\textbf{Notes on scope and relevance.}  Under exp-concave or strongly convex losses, the important question of finding an optimal (wrt universal dynamic regret) and proper algorithm has remained resistant to attacks in the non-stationary online learning literature for almost two decades since the work of \citep{zinkevich2003online}. In this work, we take the first steps in addressing this question by showing optimality of proper SA learner in proper learning settings. The fact that a proper version of Strongly Adaptive algorithms can lead to optimal rates was highly unclear from the analysis of \citep{improperDynamic}. Further, by lifting the gradient smoothness assumption for the revealed losses, we modestly enlarge the applicability of the results when compared to \citep{improperDynamic}. Though our proof techniques bear some semblance with that of \citep{improperDynamic} in terms of the usage of KKT conditions, this similarity is only superficial and we introduce several new non-trivial ideas in the analysis for attaining the new results (see Sections \ref{sec:road-sq} and \ref{sec:road-ec}).

\section{Related Work}
In this section, we compare and contrast our work with several existing lines of research.

\textbf{Dynamic regret minimization in non-stationary online learning}. Apart from \citep{improperDynamic}, our work fits into the broad literature of dynamic regret minimization in online learning such as \citep{zinkevich2003online,besbes2015non,jadbabaie2015online,yang2016tracking,Mokhtari2016OnlineOI,chen2018non,zhang2018adaptive,zhang2018dynamic,yuan2019dynamic,goel2019OBD,arrows2019,Zhao2020DynamicRO,cutkosky2020ParameterFree,baby2020higherTV,Zhao2021StronglyConvex,baby2021TVDenoise,Zhao2021Memory,Chang_Shahrampour_2021,baby2021KRR}. However, to the best of our knowledge none of these works are known to attain the optimal dynamic regret rate for our setting in terms of path length of the arbitrary comparator sequence.

\textbf{Adaptive online learning}. There is a complementary body of work on Strongly Adaptive regret minimization such as \citep{daniely2015strongly,jun2017CoinBetting,cutkosky2020ParameterFree,Zhang2021DualAA} and Adaptive regret minimization such as \citep{hazan2007adaptive,koolen2016specialist} (which are in fact Strongly Adaptive wrt exp-concave losses) that aims at controlling the static regret in any local time interval. This work focuses on developing new guarantees for algorithms that are Strongly Adaptive (SA) wrt strongly convex / exp-concave losses. The base learners we use for SA methods are the static regret minimizing algorithms from \citep{hazan2007logregret}.

\textbf{Locally adaptive non-parametric regression}. Our work is closely related to locally adaptive non-parametric regression literature from the statistics community such as \citep{mammen1991,vandegeer1990,donoho1998minimax,l1tf,tibshirani2014adaptive,wang2014falling, graphtf,  guntuboyina2018constrainedTF,ortelli2019prediction}. This work supplements them by removing the statistical assumptions and enabling to go beyond squared error losses for the non-parametric function class of TV bounded functions.

\textbf{Online non-parametric regression}. The results of \citep{rakhlin2014online} certifies that the minimax rate for competing against a reference class of TV bounded functions with squared error losses is $O(n^{1/3})$. However this bound doesn't capture the correct dependence on $C_n$ and is arrived via non-constructive arguments. On the other hand we arrive at the optimal dependence on both $n$ and $C_n$ via an efficient algorithm. Further, our results with squared error losses in Section \ref{sec:square-game} are more general than that of \citep{improperDynamic} (see Remark \ref{rmk:oracle_inequality}). Results on online non-parametric regression against reference class of Lipschitz functions, Sobolev functions and isotonic functions can be found in \citep{gaillard2015chaining,koolen2015minimax,kotlowski2016} respectively. However as noted in \citep{arrows2019}, these classes feature functions that are more regular than TV bounded functions. In fact they can be embedded inside a TV bounded function class. So the minimax optimally for TV class implies minimax optimality for the smoother function classes as well. 

We refer the reader to \citep{improperDynamic} and references therein for a more elaborate survey on existing literature.

\section{A gentle start: Squared loss games} \label{sec:square-game}

To start with, we consider the following squared loss game which will later play a pivotal role in the generalization to strongly convex losses.
\begin{itemize}
    \item At time $t \in [n] := \{1,\ldots,n \}$, player predicts $x_t \in [-B,B]$.
    \item Adversary reveals a label $y_t \in [-G,G]$
    \item Player suffers loss $(y_t - x_t)^2$.
\end{itemize}

We make the following assumption.

\textbf{Assumption A1: } \label{assm:wlog} We assume that $[-B,B] \subseteq [-G,G]$ with $B \ge 1$ without loss of generality.

Define a class of comparators as:
\begin{align}
\mathcal{TV}^B(C_n) := \Bigg \{ \Bigg. w_{1:n} \Bigg| \mathrm{TV}(w_{1:n}) &:= \sum_{t=2}^n |w_t - w_{t-1}| \le C_n, |w_t| \le B \: \forall t \in [n] \Bigg. \Bigg \}. \label{eq:tvb}
\end{align}

We are interested in simultaneously controlling the dynamic regret against all sequences in $\mathcal{TV}^B(C_n)$. The main algorithm we use for this task is the Follow-the-Leading-History (FLH) from \citep{hazan2007adaptive} with Online Gradient Descent (OGD) run on the decision set $[-B,B]$ as base learners. This algorithm will be referred as \emph{FLH-OGD} strategy henceforth. We provide a description of FLH in Appendix \ref{app:prelims} for completeness.  We have the following performance guarantee.

\begin{theorem} \label{thm:main-sq}
Suppose the labels $y_t$ generated by the adversary belong to $[-G,G]$. Let $x_t$ be the prediction at time $t$ of FLH with learning rate $\zeta = 1/(2(G+B)^2)$, base learners as OGD with step sizes $1/(2t)$ and decision set $[-B,B]$. Then for any comparator sequence $(w_1,\ldots,w_n) \in \mathcal{TV}^B(C_n)$
\begin{align}
    \sum_{t=1}^{n} (y_t - x_t)^2 - (y_t - w_t)^2
    &= \tilde O \left( n^{1/3}C_n^{2/3} \vee 1 \right),
\end{align}
where $\tilde O(\cdot)$ hides dependence on logarithmic factors of horizon $n,G,B$ and $a \vee b := \max \{a, b \}$.
\end{theorem}

\begin{remark}[\textbf{Adaptivity to }$C_n$ \textbf{and safe (non-stochastic) oracle inequality}]\label{rmk:oracle_inequality}
The FLH-OGD strategy does not require $C_n$ as an input. Further, Theorem \ref{thm:main-sq} has implications in non-parametric regression under safety constraints. When the non-parametric estimator for the $\mathcal{TV}^B$ sequence class is required to obey a safety constraint that the estimator's outputs $x_t$ must also lie in $[-B,B]$, Theorem \ref{thm:main-sq} implies the following oracle inequality:
 \begin{align}
 \sum_{t=1}^{n} (y_t - x_t)^2 + g(x_t) \leq \min_{w_{1:n}} \sum_{t=1}^n(y_t - w_t)^2 + g(w_t) + \tilde O \left( n^{1/3}\mathrm{TV}(w_{1:n})^{2/3} \vee 1 \right),
 \end{align}
 where $g(x)$ is a safety constraint such that $g(x) = \infty$ when $|x| > B$ and zero otherwise. This is a strict generalization of Remark 2 in \citep{improperDynamic}.
\end{remark}

\subsection{Key insight behind the proof of Theorem \ref{thm:main-sq}} \label{sec:key}
The insight we used in deriving regret rate in Theorem \ref{thm:main-sq} for a proper learning setup is based on the following idea: Suppose that we need to compete against a comparator sequence that incurs a Total Variation (TV) of $C_n$. We observe that, this comparator sequence of decisions in hindsight requires to obey the TV constraint while the decisions of the Strongly Adaptive (SA) learner need not obey any such constraints. Consider a time interval $I$  where the comparator sequence assumes a constant value (say $v_1$) in an \emph{arbitrary} convex decision set $D$. There could be some other point in $D$ (say $v_2$) which can incur better cumulative loss within that interval. Note that the comparator sequence may not assume the value $v_2$ in the interval $I$ due to the global TV constraint. Due to the strongly adaptive property, the regret (against $v_1$) of the SA learner in interval $I$ is then bounded by the regret (against $v_1$) of the static point $v_2$, which is less than or equal to zero,  plus an extra log term. The presence of such non-positive terms can delicately offset the effect of the positive log terms when summed across all such intervals to get favorable dynamic regret rates. How small the non-positive terms are, when summed across all intervals, depends on the magnitude of $C_n$ (and indirectly on $n$).


\subsection{Detailed road map for the proof of Theorem \ref{thm:main-sq}} \label{sec:road-sq}
In this section,  we focus on conveying the main ideas of our proof deferring the formal details to Appendix \ref{app:square-games}. We start by briefly reviewing the proof strategy of \citep{improperDynamic} and then intuitively capture the points of similarities and differences in our analysis. Throughout the proof we use the shorthand $[a,b] := \{a, a+1, \ldots, b \}$ for two natural numbers $a < b$.

We start by characterizing the offline optimal.
Define the sign function as $\sign{(x)} = 1 \text{ if } x > 0$; $ -1  \text{ if } x < 0$; and some $v \in [-1,1]  \text{ if } x = 0$.

\begin{restatable}{lemma}{lemkktsq} \label{lem:kkt-sq}
(\textbf{characterization of offline optimal}) 
Consider the following convex optimization problem (where $\tilde{z}_1,...,\tilde{z}_{n-1}$ are introduced as dummy variables)
\begin{mini!}|s|[2]                   
    {\tilde u_1,\ldots,\tilde u_n,
    \tilde{z_1},\ldots,\tilde z_{n-1}}                               
    {\frac{1}{2}\sum_{t=1}^{n} (y_t - \tilde u_t)^2}   
    {\label{eq:Example1}}             
    {}                                
    \addConstraint{\tilde z_t}{=\tilde u_{t+1} - \tilde u_{t} \: \forall t \in [n-1],}    
    \addConstraint{\sum_{t=1}^{n-1} |\tilde z_t|}{\le C_n, \label{eq:constr-1}}  
    \addConstraint{-B}{\le \tilde u_t \: \forall t \in [n],\label{eq:constr-2}}
    \addConstraint{\tilde u_t}{\le B \: \forall t \in [n],\label{eq:constr-3}}
\end{mini!}
Let $u_1,\ldots,u_n,z_1,\ldots,z_{n-1}$ be the optimal primal variables and let $\lambda \ge 0$ be the optimal dual variable corresponding to the constraint \eqref{eq:constr-1}. Further, let $\gamma_t^- \ge 0, \gamma_t^+ \ge 0$ be the optimal dual variables that correspond to constraints \eqref{eq:constr-2} and \eqref{eq:constr-3} respectively for all $t \in [n]$. By the KKT conditions, we have

\begin{itemize}
    \item \textbf{stationarity: } $u_t - y_t = \lambda \left ( s_t - s_{t-1} \right) +  \gamma^-_t -  \gamma^+_t$, where $s_t \in \partial|z_t|$ (a subgradient). Specifically, $s_t=\sign(u_{t+1}-u_t)$ if $|u_{t+1}-u_t|>0$ and $s_t$ is some value in $[-1,1]$ otherwise. For convenience of notations later, we also define 
    $s_n = s_0 = 0$.
    \item \textbf{complementary slackness: } (a) $\lamda \left(\sum_{t=2}^n |u_t - u_{t-1}| - C_n \right) = 0$; (b)  $ \gamma^-_t ( u_t + B) = 0$ and $ \gamma^+_t ( u_t - B) = 0$ for all $t \in [n]$
\end{itemize}
\end{restatable}

Let the optimal solution constructed by the offline oracle be denoted by $u_{1:n}$ (termed as offline optimal henceforth). In \citep{improperDynamic}, a partition $\cP = \{[i_s,i_t], i \in [M] \}$ of $[n]$ is formed with cardinality $| \cP | = M = O(n^{1/3}C_n^{2/3} \vee 1)$. The partition has an additional property that within each bin $[i_s,i_t] \in \cP$, we have $C_i := \sum_{j=i_s+1}^{i_t} |u_j - u_{j-1}| \le B/\sqrt{i_t-i_s+1}$ (see Lemma \ref{lem:part-sq}). Then for each bin, a three term regret decomposition is employed as follows:
\begin{align}
    \underbrace{\sum_{j=i_s}^{i_t} (y_j - x_j)^2 - (y_j - \bar{y}_i)^2}_{T_{1,i}} + \underbrace{\sum_{j=i_s}^{i_t} (y_j - \bar{y}_i)^2 - (y_j - \bar{u}_i)^2}_{T_{2,i}}+ \underbrace{\sum_{j=i_s}^{i_t} (y_j - \bar{u}_i)^2 - (y_j - u_j)^2,}_{T_{3,i}} \label{eq:reg-sq}
\end{align}
where $\bar u_i = \sum_{j=i_s}^{i_t} u_j / (i_t - i_s + 1)$ and $\bar y_i = \sum_{j=i_s}^{i_t} y_j / (i_t - i_s + 1)$ and $x_j$ are the predictions of the learner.
They use online averaging as base learners for FLH. By strong adaptivity, they show $T_{1,i} = O(\log n)$. They show that $T_{3,i}$ can be $O(\lamda C_i)$ in general  where $\lamda$ is the dual variable arising from the KKT conditions (see Lemma \ref{lem:kkt-sq}) which can be even $\Theta(n)$ in the worst case. Since $\bar y_i$ is the static minimizer of $g(x) = \sum_{j=i_s}^{i_t} (y_j - x)^2$, they bound $T_{2,i}$ by a non-positive term which when added to $T_{3,i}$ can diminish into an $O(1)$ quantity. Thus regret within the bin $[i_s,i_t]$ is $T_{1,i}+T_{2,i}+T_{3,i} = O(\log n)$. This regret bound is added across all $O(n^{1/3}C_n^{2/3} \vee 1)$ bins of $\cP$ to yield an $\tilde O(n^{1/3}C_n^{2/3} \vee 1 )$ dynamic regret.

\begin{figure}[h!]
\centering
\includegraphics[width=6cm,height=5cm]{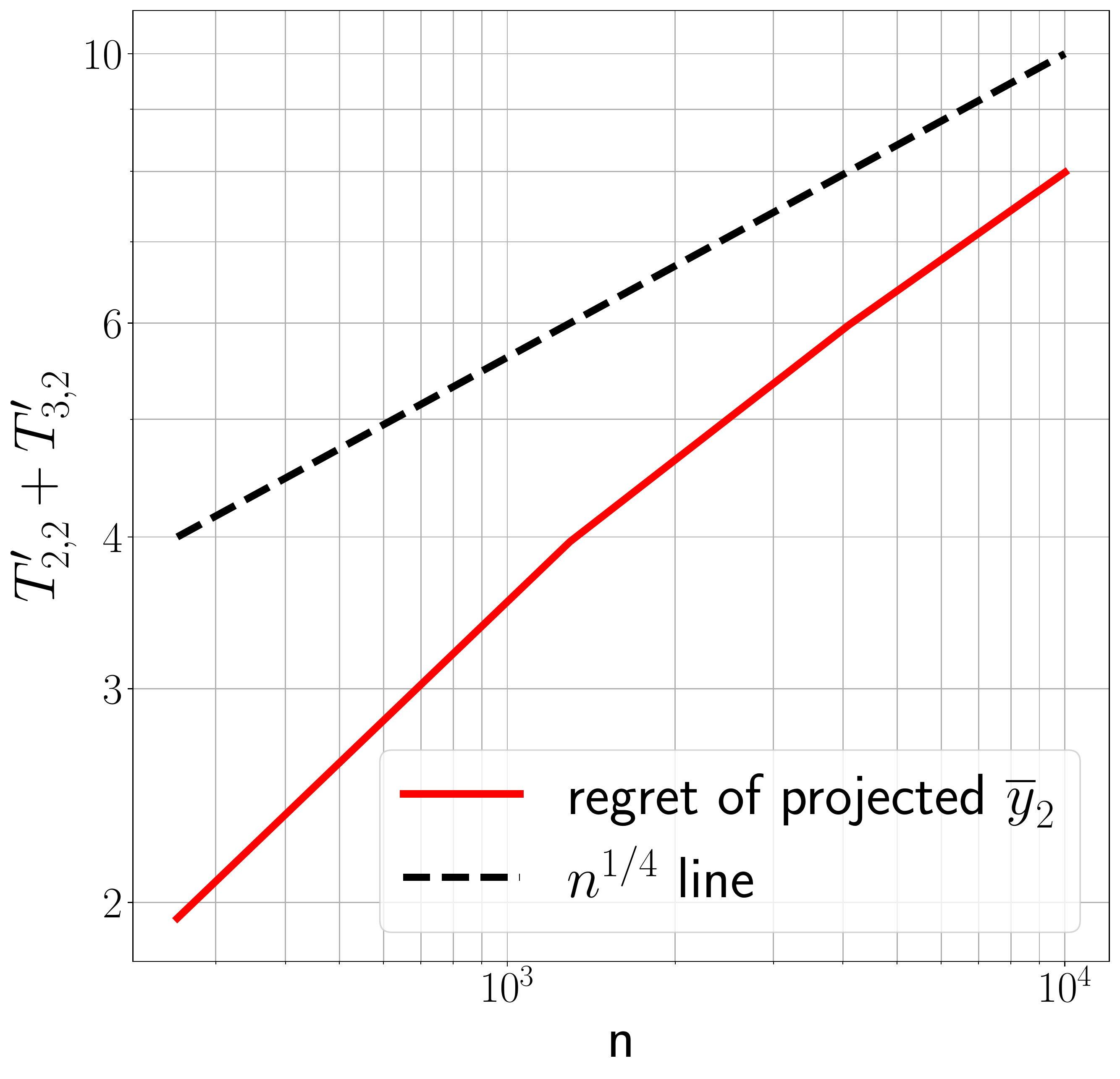}
\caption{\emph{Plot of $T'_{2,2}+T'_{3,2}$ (see Eq.\eqref{eq:reg-sq1} with $i=2$) for the Example \ref{ex} in Appendix \ref{app:square-games}. In this example, $C_n = O(1/\sqrt{n})$ and the partitioning procedure of \citep{improperDynamic} creates a partition $\cP$ of $[n]$ containing two bins. We see that $T'_{2,2}+T'_{3,2}$ in the second bin grows roughly as $O(n^{1/4})$. However for applying the analysis of \citep{improperDynamic}, we require this quantity for each bin in $\cP$ to grow as $O(1)$.  This makes the direct extension of the techniques in \citep{improperDynamic} with $\bar y_i$ replaced by $\Pi(\bar y_i)$ as in Eq.\eqref{eq:reg-sq1} inapplicable for the proper learning setting we study.}}    
\label{fig:large-t2}
\end{figure}

In our protocol of squared loss games, the labels $y_t \in [-G,G] \supseteq [-B,B]$. So we can't use online averages as base learner for constructing a proper learning algorithm. So in this work we use projected OGD as base learners with decision set $[-B,B]$. With such an algorithm, we may attempt to work with a slightly modified version of the three term regret decomposition of \eqref{eq:reg-sq} as:

\begin{align}
    \underbrace{\sum_{j=i_s}^{i_t} (y_j - x_j)^2 - (y_j - \Pi(\bar{y}_i))^2}_{T'_{1,i}} + \underbrace{\sum_{j=i_s}^{i_t} (y_j - \Pi(\bar{y}_i))^2 - (y_j - \bar{u}_i)^2}_{T'_{2,i}} + \underbrace{\sum_{j=i_s}^{i_t} (y_j - \bar{u}_i)^2 - (y_j - u_j)^2}_{T'_{3,i}}, \label{eq:reg-sq1}
\end{align}
where $\Pi(x)$ is the projection of $x \in \mathbb{R}$ to the interval $[-B,B]$. Unfortunately while doing so, the term $T'_{2,i}$ can be not negative enough to diminish $T'_{3,i}$ to an $O(1)$ quantity. We provide an empirical demonstration of this phenomenon in Fig.\ref{fig:large-t2}. At this point, we hope that we have made a clear case on why the analysis of \citep{improperDynamic} cannot be directly extended to handle proper learning.

To get around this issue, we first identify two regimes for the dual variable $\lamda$. We show that when $\lamda = O(n^{1/3}/C_n^{1/3})$, one can still work with the same partitioning $\cP$ of \citep{improperDynamic} (see Lemma \ref{lem:part-sq}) and use a decomposition similar to Eq.\eqref{eq:reg-sq1} to get the desired regret bound (see Lemma \ref{lem:low-lamda}).

Before explaining the details of the regime $\lamda = \Omega(n^{1/3}/C_n^{1/3})$, we introduce the following definitions for convenience:

\begin{restatable}{definition}{defoned} \label{def:struct-gap-1d}\
\begin{itemize}
    \item For a bin $[a,b] \subseteq \{2,\ldots,n-1 \}$, the offline optimal solution is said to assume Structure 1 if $u_j =u_a \in (-B,B)$ for all $j \in [a,b]$ and $u_b >  u_{b+1}$ and $ u_a >  u_{a-1}$.

    \item For a bin $[a,b] \subseteq \{2,\ldots,n-1 \}$, the offline optimal solution is said to assume Structure 2 if $u_j =u_a \in (-B,B)$ for all $j \in [a,b]$ and $u_b <  u_{b+1}$ and $ u_a <  u_{a-1}$.
    
    \item For a bin $[a,b]$, we define $\text{gap}_{\text{min}}(\beta,[a,b]) := \min_{j \in [a,b]} |u_j - \beta|$ where $\beta \in \mathbb{R}$.
\end{itemize}
\end{restatable}

Consider the following two conditions.

\textbf{Condition 1:} For a bin $[i_s,i_t] \in \cP$, the offline optimal satisfies $\text{gap}_{\text{min}}(-B,[i_s,i_t]) \ge \text{gap}_{\text{min}}(B,[i_s,i_t])$ and within at-least one sub-interval $[r,s] \subseteq [i_s,i_t]$, the offline optimal assumes the form of Structure 2.

\textbf{Condition 2:} For a bin $[i_s,i_t] \in \cP$, the offline optimal satisfies $\text{gap}_{\text{min}}(-B,[i_s,i_t]) < \text{gap}_{\text{min}}(B,[i_s,i_t])$ and within at-least one sub-interval $[r,s] \subseteq [i_s,i_t]$, the offline optimal assumes the form of Structure 1.

Define:

$$\cQ := \{[i_s,i_t] \in \cP : \text{ the offline optimal satisfies Condition 1 or 2 in } [i_s,i_t] \}.$$

\begin{figure}[h!]
\includegraphics[width=\linewidth,keepaspectratio=true]{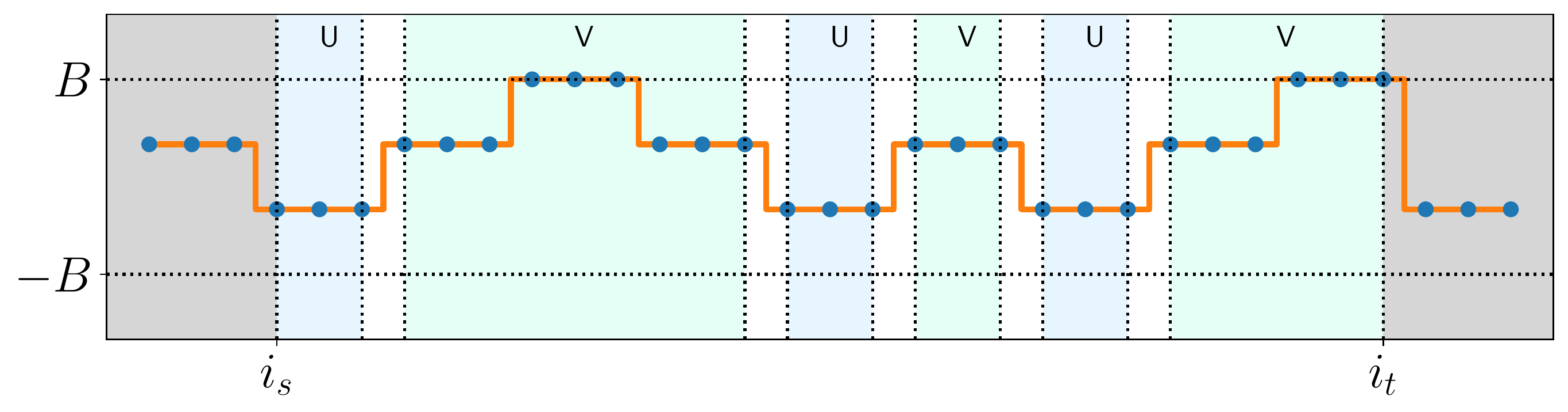}
    \caption{\emph{Refinement of a bin $[i_s,i_t] \in \cP$ that satisfy Condition 1 in Section \ref{sec:road-sq} into smaller style U and style V sub-intervals. Blue dots represent the optimal sequence}}    
    \label{fig:splits-1d-ex}
\end{figure}

We refine a bin $[i_s,i_t] \in \cQ$ that satisfy Condition 1 into smaller sub-intervals as shown in Fig.\ref{fig:splits-1d-ex}, such that: for a style U sub-interval, the offline optimal takes the form of Structure 2 and for a style V sub-interval, the offline optimal has a non-decreasing section followed by an optional decreasing section. A similar refinement is also performed for bins in $\cQ$ that satisfy Condition 2.

Our strategy is to bound:
\begin{align}
    \text{regret in style U sub-intervals}
    &= O(\log n) +  \text{a negative term}. \label{eq:style-u}
\end{align}

This is accomplished by a two term regret decomposition. Suppose $[a,b]$ is a style U sub-interval. We use the decomposition:
\begin{align}
 \underbrace{\sum_{j=a}^b (y_j - x_j)^2 - (y_j - w)^2}_{T_1} + \underbrace{\sum_{j=a}^b (y_j - w)^2 - (y_j - u_j)^2,}_{T_2} \label{eq:sketch}
\end{align}
with $w = \Pi \left ( \sum_{j=a}^b y_j/(b-a+1) \right)$.

Next, we bound
\begin{align}
    \text{regret in style V sub-intervals}
    &= O(\log n), \label{eq:style-v}
\end{align}

using a similar two term regret decomposition as in Eq.\eqref{eq:sketch} with $w$ replaced by a carefully chosen $w_j \in [(u_a \wedge \ldots \wedge u_b), (u_a \vee \ldots \vee u_b)]$ such that $\sum_{j=a+1}^b \mathbb{I} \{w_j \neq w_{j-1} \} \le 6$ where $\mathbb{I} \{ \cdot \}$ is the indicator function taking values in $\{0,1 \}$. We use the notation $x \wedge y = \min \{x, y \}$

We perform this task of refinement for every interval $[i_s,i_t]$ in $\cQ$. Then we bound the regret in the resulting sub-intervals (as per Eq.\eqref{eq:style-u} or \eqref{eq:style-v}) and add the regret bounds across all such sub-intervals. Note that the total number of sub-intervals after refinement can be much larger than $|\cP|  = O(n^{1/3}C_n^{2/3} \vee 1)$. So if the bound in Eq.\eqref{eq:style-u} is not tight enough, then there is a possibility that the resulting regret bound can be highly sub-optimal. This poses a major challenge in contrast to the analysis of \citep{improperDynamic} where they only need to work with a partition of size $O(n^{1/3}C_n^{2/3} \vee 1)$ and bound the regret in each interval of the partition by an $\tilde O(1)$ quantity. 

To address this issue, we form tight bounds for Eq.\eqref{eq:style-u} by exploiting certain structures in the KKT conditions that were previously unexplored in \citep{improperDynamic} via Lemmas \ref{lem:lamda-len}, \ref{lem:outside1d}, \ref{lem:large-margin} and \ref{lem:high-lamda}. Of particular interest is Lemma \ref{lem:lamda-len} which highlights a fundamental way in which the adversary is constrained. Then we prove that if every bin $[i_s,i_t] \in \cQ$ satisfies $\text{gap}_{\text{min}}(-B,[i_s,i_t]) \vee \text{gap}_{\text{min}}(B,[i_s,i_t]) \ge \mu_{\text{th}}$ where $\mu_{\text{th}}$ is as defined in Lemma \ref{lem:large-margin}, then the culmination of the negative terms in Eq.\eqref{eq:style-u} can gracefully offset the effect of the positive $O(\log n)$ terms in Eq.\eqref{eq:style-u} and Eq.\eqref{eq:style-v} when summed across all refined intervals to obtain an $O(n^{1/3}C_n^{2/3} \vee 1)$ bound overall for $\sum_{[i_s,i_t] \in \cQ} \sum_{j=i_s}^{i_t} (y_j - x_j)^2 - (y_j - u_j)^2$ (see proof of Lemma \ref{lem:high-lamda}). 

Further we show in Lemma \ref{lem:large-margin} that when $\lamda  = \Omega(n^{1/3}/C_n^{1/3})$ and $C_n = \tilde O(n)$, the criterion $\text{gap}_{\text{min}}(-B,[i_s,i_t]) \vee \text{gap}_{\text{min}}(B,[i_s,i_t]) \ge \mu_{\text{th}}$ is always satisfied for every bin $[i_s,i_t] \in \cQ$. This can be seen informally as follows. Recall that the TV of the offline optimal within the bin $[i_s,i_t]$ is a ``small'' quantity that is at-most $(B/\sqrt{i_t-i_s+1}) \le B$. So if $\text{gap}_{\text{min}}(-B,[i_s,i_t])$ is small, then due to this small TV constraint, we expect the quantity $\text{gap}_{\text{min}}(B,[i_s,i_t])$ to be sufficiently large and vice versa.

Finally, for each bin in $\cR := \cP \setminus \cQ$ we show (by using Lemma \ref{lem:mono-1d}) that its regret contribution can be bounded by $O(\log n)$. Since $|\cR| = O(n^{1/3}C_n^{2/3} \vee 1)$, such regret bounds lead to $\tilde O(n^{1/3}C_n^{2/3} \vee 1)$ bound overall when summed across all bins in $\cR$.

Before closing this section, we capture the intuition behind the importance of the criterion $\text{gap}_{\text{min}}(-B,[i_s,i_t]) \vee \text{gap}_{\text{min}}(B,[i_s,i_t]) \ge \mu_{\text{th}}$ and why it can produce a sufficiently negative term in Eq.\eqref{eq:style-u}.  Let's consider a style U sub-interval $[a,b]$ obtained by refining a bin $[i_s,i_t] \in \cQ$ which satisfy Condition 1. Since $[a,b]$ is style U sub-interval, the offline optimal takes the form of Structure 2 in $[a,b]$. Suppose that $|B + u_a| \ge \text{gap}_{\text{min}}(-B,[i_s,i_t]) \ge \text{gap}_{\text{min}}(B,[i_s,i_t]) \ge \mu_{\text{th}}$. Here the first inequality holds by the definition of $\text{gap}_{\text{min}}(-B,[i_s,i_t])$. Also, note that $u_j = u_a$ for all $j \in [a,b]$ by the definition of Structure 2. Let $\bar y_{a \rightarrow b} := \sum_{j=a}^b y_j/(b-a+1)$. From the KKT conditions it can be shown that $\bar y_{a \rightarrow b} < u_a$. We provide intuitive explanation for the case $\Pi(\bar y_{a \rightarrow b}) = -B$. This can happen only when $\bar y_{a \rightarrow b} \le -B$. Qualitatively in such a scenario, we expect the decision $-B$ to be much better than playing the decision $u_a$ which is bigger than $-B$. Whenever there is sufficient gap (more formally a gap of at-least $\mu_{\text{th}}$) between $-B$ and $u_a$, one can expect that $u_a$ can be very sub-optimal in comparison to $-B$ ($ = \Pi(\bar y_{a \rightarrow b})$) which makes the term $T_2$ in Eq.\eqref{eq:sketch} (with $w = -B$ and $u_j = u_a$) sufficiently negative. 

When $\bar y_{a \rightarrow b} \in (-B,B)$, $T_2$ with $w = \bar y_{a \rightarrow b}$ can be shown to be sufficiently negative using the arguments of \citep{improperDynamic}. However, the interplay of this negative term with the sum of regret bounds in all refined intervals is more delicate as described in the proof of Lemma \ref{lem:high-lamda}.

\section{Performance guarantees for strongly convex losses}\label{sec:sc} \label{sec:box}
In this section, we extend the results on squared error losses to general strongly convex losses.

\subsection{Strongly convex losses and box decision set}
In this section, we show that the style of analysis presented for squared error losses directly generalizes to strongly convex losses in multi-dimensions whenever the decision set is an $L_\infty$ norm ball. The main idea is to provide a reduction to the uni-variate squared loss games via standard surrogate loss tricks \citep{hazan2007logregret} and instantiate FLH-OGD appropriately. All unspecified proofs for this section are deferred to Appendix \ref{app:sc}. We consider the following protocol:

\begin{itemize}
    \item At time $t \in [n]$ learner predicts $\bs x_t \in \mathbb{R}^d$ with $\| \bs x_t\|_\infty \le B$.
    \item Adversary reveals loss $f_t$.
    \item Learner suffers loss $f_t(\bs x_t)$.
\end{itemize}

We have the following Corollary due to Theorem \ref{thm:main-sq}.

\begin{restatable}{corollary}{corSc} \label{cor:sc}
Let the loss functions $f_t$ be $H$ strongly convex in $L_2$ norm across the (box) domain $\cD = \{ \bs x \in \mathbb{R}^d : \| \bs x\|_\infty \le B\}$. i.e, $f_t(\bs y) \ge f_t(\bs x) + \grad f_t(\bs x)^T (\bs y- \bs x) + \frac{H}{2} \| \bs y - \bs x \|_2^2$ for all $\bs x, \bs y \in \cD$. Suppose $\|\grad f_t(\bs x) \|_\infty \le G_\infty$ for all $\bs x \in \cD$. For each $i \in [d]$, construct surrogate losses $\ell_t^{(i)}:\mathbb{R} \rightarrow \mathbb{R}$ as $\ell_t^{(i)}(x) = \left( x - (\bs x_t[i] - \grad f_t(\bs x_t)[i]/H) \right)^2$ where $\bs x_t$ is the prediction of the learner at time $t$. By running $d$ instances of uni-variate FLH-OGD (Fig.\ref{fig:flh} in Appendix \ref{app:prelims}) with decision set $[-B,B]$ and learning rate $\zeta = 1/(2(2B+G_\infty/H)^2)$ where instance $i$ predicts $\bs x_t[i]$ at time $t$ and suffers losses $\ell_t^{(i)}$, we have
\begin{align}
    \sum_{t=1}^{n} f_t(\bs x_t) - f_t(\bs w_t) = \tilde O\left ( d^{1/3}n^{1/3} C_n^{2/3} \vee d \right),
\end{align}
for any comparator sequence $\bs w_{1:n}$ with $ TV(\bs w_{1:n}) := \sum_{t=2}^{n} \|\bs w_t - \bs w_{t-1} \|_1 \le C_n$. $\tilde O(\cdot)$ hides the dependence on factors of $\log n, B, H, G_\infty$.
\end{restatable}

When compared with the information theoretic lower bound of \citep{improperDynamic} (Proposition 11 there), we see that the rate of Theorem \ref{thm:main-sq} is optimal (modulo log factors) wrt to $n,C_n$ and $d$. The dependence of $\tilde O(d)$ for low $C_n$ regimes is due to the fact that we only assume $\|\grad f_t(\bs x) \|_\infty = O(1)$ as opposed to assuming $\|\grad f_t(\bs x) \|_2 = O(1)$.

\begin{remark} \textbf{(relaxed assumptions \& improvements)}
Unlike \citep{improperDynamic}, we do not assume gradient Lipschitzness of the losses $f_t$. Further, for the box decision set, our results attain an optimal $O(d^{1/3})$ dimension dependence on regret in the non-trivial regime of $C_n \ge 1/n$ in comparison to the $O(d^2)$ dependence of \citep{improperDynamic} for strongly convex losses.
\end{remark}

\begin{remark}
We emphasize that the theory developed in Section \ref{sec:square-game} is vital for extending the results with the surrogate losses as in Corollary \ref{cor:sc}. Consider squared losses $\ell_t(x) = (x - y_t)^2$ with labels $y_t$ such that $|y_t| \le Y$ for all $t$.  \citep{improperDynamic} requires that the predictions $x_t$ obey $x_t \in [-Y,Y]$. In our use case with surrogate losses $\ell_t^{(i)}(x) = \left( x - (\bs x_t[i] - \grad f_t(\bs x_t)[i]/H) \right)^2$ such a requirement can be not well defined. Here the labels can be regarded $y_t = \bs x_t[i] - \grad f_t(\bs x_t)[i]/H$ which depends on $\bs x_t[i]$. As per the setup of Corollary \ref{cor:sc}, the $i^{\text{th}}$ FLH-OGD instance uses losses $\ell_t^{(i)}, \: t \in [n]$ and its prediction at time $t$ is $\bs x_t[i]$. So constructing a uniform bound $Y$ to contain the predictions $\bs x_t[i]$ requires a uniform bound on the predictions $\bs x_t[i]$ itself for all $t$ which is self conflicting. Hence the strategy of \citep{improperDynamic} for squared error losses is incompatible for using the surrogate losses $\ell_t^{(i)}$.

\end{remark}

\subsection{Strongly convex losses and general convex decision sets}

In this section, we show how to convert an optimal algorithm  described in Section \ref{sec:box} for the box decision set to an optimal (modulo factors of $\log n$ and dimensions dependencies) algorithm for any convex decision set via a black box reduction. This reduction is essentially due to the seminal work of  \citep{Cutkosky2018BlackBoxRF}.

\begin{figure}[h!]
	\centering
	\fbox{
		\begin{minipage}{12 cm}
		Box to general convex set reduction: Inputs - Decision set $\cW$, $G > 0$
    \begin{enumerate}
        \item Let $\cD$ be the tightest box that circumscribes $\cW$. i.e, $\cD = \{\bs x \in \mathbb{R}^d: \| \bs x\|_\infty \le \sup_{\bs w \in \cW} \| \bs w \|_\infty \}$.
        
        \item Let $\cA$ be the algorithm attaining the guarantee in Corollary \ref{cor:sc} with decision set $\cD$ and $G_\infty = 2G$.
        
        \item At round $t$, get iterate $\bs x_t$ from $\cA$.
        \item Play $\hat{\bs x}_t = \Pi_{\cW}(\bs x_t) := \argmin_{\bs y \in \cW} \|\bs x_t - \bs y \|_1$.
        \item Get loss $f_t$.
        \item Construct surrogate loss $\ell_t(\bs x) = f_t(\bs x) + G \cdot S(\bs x)$, where $S(\bs x) := \| \bs x - \Pi_{\cW}(\bs x)\|_1$.
        \item Send $\ell_t(\bs x)$ to $\cA$.
    \end{enumerate}
		\end{minipage}
	}
	\caption{Black box reduction from box to arbitrary convex decision set. This technique is due to \citep{Cutkosky2018BlackBoxRF}.}
	\label{fig:bbox}
\end{figure}

We have the following guarantee for the scheme in Fig. \ref{fig:bbox}.

\begin{theorem}
Assume the notations in Fig. \ref{fig:bbox}. Let the input decision set be $\cW$. Let the losses be $H$ strongly convex in $L_2$ norm across $\cD$ and satisfy $\|\grad f_t(\bs x) \|_\infty \le G$ for all $\bs x \in \cD$. Then the reduction scheme in Fig. \ref{fig:bbox} guarantees that
\begin{align}
    \sum_{t=1}^n f_t(\hat{\bs x}_t) - f_t(\bs w_t)
    &= \tilde O\left ( d^{1/3}n^{1/3} C_n^{2/3} \vee d \right),
\end{align}
for any comparator sequence $\bs w_{1:n} \in \cW$ with $ TV(\bs w_{1:n}) := \sum_{t=2}^{n} \|\bs w_t - \bs w_{t-1} \|_1 \le C_n$. $\tilde O(\cdot)$ hides the dependence on factors of $\log n, H, G_\infty$.
\end{theorem}
\begin{proof}
We start by listing several observations. First, note that the function $S(\bs x)$ is convex and 1-Lipschitz across $\mathbb{R}^d$. (Proposition 1 in \citep{Cutkosky2018BlackBoxRF}). 

Also, the sub-gradient $\partial S(\bs x) = \{ \bs y \in \mathbb{R}^d : \bs y[j] = \sign \left (x[j] - \Pi_{\cW}(\bs x)[j] \right) \: j \in [d] \}$ (due to Theorem 4 in \citep{Cutkosky2018BlackBoxRF}). Here $\sign(a) = a/|a|$ if $|a| > 0$ and any number between $[-1,1]$ otherwise.

Finally the surrogate losses $\ell_t$ are $H$ strongly convex in $L2$ norm across $\cD$, as adding a convex function to strongly convex function preserves strong convexity. However, $\ell_t$ are \emph{not} gradient Lipschitz due to the component $G \| \bs x - \Pi_{\cW}(\bs x)\|_1$ being not smooth.

We have that for any $\bs x \in \cD$,
\begin{align}
    \|\grad \ell_t(\bs x) \|_\infty
    &\le \| \grad f_t(\bs x) \|_\infty + G \| \partial S(\bs x) \|_\infty\\
    &\le 2G,
\end{align}
where the last line is due to the assumption that $\|\grad f_t(\bs x) \|_\infty \le G$ and $\partial S(\bs x)$ is just a vector of signs as established before.

Hence we have that the losses $\ell_t$ sent to algorithm $\cA$ satisfy the conditions of Corollary \ref{cor:sc} with $G_\infty = 2G$. Hence we have that
\begin{align}
    \sum_{t=1}^n \ell_t(\bs x_t) - \ell_t(\bs w_t)
    &= \tilde O\left ( d^{1/3}n^{1/3} C_n^{2/3} \vee d \right), \label{eq:acontrol}
\end{align}
where $\bs w_{1:n}$ is as mentioned in the theorem statement.

By Taylor's theorem, we have that for some $\bs z$ in the line segment joining $\bs x_t$ and $\hat {\bs x}_t$
\begin{align}
    f_t(\hat {\bs x}_t)
    &=f_t(\bs x_t) + \grad f_t(\bs z)^T (\hat {\bs x}_t - \bs x_t)\\
    &\le f_t(\bs x_t) + G \|\hat {\bs x}_t - \bs x_t \|_1\\
    &= \ell_t(\bs x_t)
\end{align}
where the inequality is due to Holder's inequality and the assumption that $\|\grad f_t(\bs x) \|_\infty \le G$ for all $\bs x \in \cD$.

Further for any $\bs w_t \in \cW$, we have that $f_t(\bs w_t) = \ell_t(\bs w_t)$. Thus overall we obtain,

\begin{align}
    \sum_{t=1}^n f_t(\bs {\hat x}_t) - f_t(\bs w_t)
    &\le \sum_{t=1}^n \ell_t(\bs x_t) - \ell_t(\bs w_t). \label{eq:bcontrol}
\end{align}

Combining Eq.\eqref{eq:acontrol} and \eqref{eq:bcontrol} now yeilds the theorem.
\end{proof}

\begin{remark}
We emphasize that the removal of gradient smoothness assumption for strongly convex losses (from \citep{improperDynamic}) as done in the current work was important to apply the reduction scheme of Fig.\ref{fig:bbox} as the losses $\ell_t$ are not gradient smooth.
\end{remark}

\section{Performance guarantees for exp-concave losses} \label{sec:ec}
In this section, we control the dynamic regret with exp-concave and gradient smooth losses when the decision set is an $L_\infty$ ball. All unspecified lemma statements and proofs are deferred to Appendix \ref{app:ec}. We make the following assumptions:

\textbf{Assumption B1:} The loss functions $\ell_t$ are $\alpha$ exp-concave in the box decision set $\cD=\{\bs x \in \mathbb{R}^d: \| \bs x\|_\infty \le B \}$ .ie, $\ell_t(\bs y) \ge \ell_t(\bs x) + \grad \ell_t(\bs x)^T (\bs y- \bs x)+ \frac{\alpha}{2} \left( \grad \ell_t(\bs x)^T (\bs y- \bs x) \right)^2$ for all $\bs x, \bs y \in \cD$.

\textbf{Assumption B2:} The loss functions $\ell_t$ satisfy $\| \grad \ell_t(\bs x)\|_2 \le G$ and $\| \grad \ell_t(\bs x)\|_\infty \le G_\infty$ for all $\bs x \in \cD$. Without loss of generality, we let $G \wedge G_\infty \wedge B \ge 1$, where $a \wedge b := \min \{a , b \}$.

We consider the following protocol:
\begin{itemize}
    \item At time $t \in [n]$ learner predicts $\bs x_t \in \mathbb{R}^d$ with $\| \bs x_t\|_\infty \le B$.
    \item Adversary reveals the loss function $\ell_t$.
\end{itemize}

In view of Assumption B1, following \citep{hazan2007logregret}, one can define the surrogate losses:
\begin{align}
    f_t(\bs x) = \left(\sqrt{\alpha/2} \grad \ell_t(\bs x_t)^T (\bs x- \bs x_t) + 1/\sqrt{2\alpha} \right)^2. \label{eq:exp-surrogate}
\end{align}

It follows that
\begin{align}
    \sum_{t=1}^n \ell_t(\bs x_t) - \ell_t(\bs w_t)
    &\le \sum_{t=1}^n f_t(\bs x_t) - f_t(\bs w_t),
\end{align}
where $\bs x_t, \bs w_t \in \cD$.

Further, we make two useful observations about surrogate losses $f_t$.

First for $\bs x \in \cD$, since $\left|\sqrt{\alpha/2} \grad \ell_t(\bs x_t)^T (\bs x- \bs x_t) + 1/\sqrt{2\alpha} \right| \le 2GB\sqrt{\alpha d/2} + 1/\sqrt{2 \alpha} := \gamma$, we have that $f_t$  are $1/(2\gamma^2)$ exp-concave over $\cD$ (see Section 3.3 in \citep{BianchiBook2006}). 

Second, since $\grad^2 f_t(\bs x) = \grad \ell_t(\bs x_t)\grad \ell_t(\bs x_t)^T \preccurlyeq G^2 \bs I$, we have that the losses $f_t$ are $G^2$ gradient Lipschitz over $\cD$.

We are interested in controlling the regret:
\begin{align}
    R_n(C_n)
    &:= \sup_{\substack{\bs w_1,\ldots,\bs w_n \in \cD \\ \sum_{t=2}^n \|\bs w_t - \bs w_{t-1} \|_1 \le C_n}}
    \sum_{t=1}^n \ell_t(\bs x_t) - \ell_t(\bs w_t),
\end{align}
where $\bs x_t$ is the decisions of the algorithm.

We have the following performance guarantee when the losses are exp-concave.

\begin{restatable}{theorem}{main}\label{thm:ec-d}
Suppose Assumptions B1-B2 are satisfied. Define $\gamma := 2GB\sqrt{\alpha d/2} + 1/\sqrt{2 \alpha}$. By using the base learner as ONS with parameter $\zeta = \min \left \{\frac{1}{16GB\sqrt{d}}, 1/(4 \gamma^2) \right \}$, decision set $\cD$, loss at time $t$ to be $f_t$ and  choosing learning rate of FLH as $\eta = 1/(2 \gamma^2)$, FLH-ONS (Fig.\ref{fig:flh} in Appendix \ref{app:prelims}) obeys
\begin{align}
    R_n(C_n)
    &\le \sup_{\substack{\bs w_1,\ldots,\bs w_n \in \cD \\ \sum_{t=2}^n \|\bs w_t - \bs w_{t-1} \|_1 \le C_n}}
    \sum_{t=1}^n f_t(\bs x_t) - f_t(\bs w_t)\\
    &= \tilde O \left( 140 d^2(8G^2B^2\alpha d + G^2B^2 + 1/\alpha) (n^{1/3}C_n^{2/3} \vee 1) \right)\mathbb{I}\{ C_n > 1/n\}\\
    &\quad + \tilde O\left( d(8G^2B^2\alpha d + 1/\alpha \right) \mathbb{I}\{ C_n \le 1/n\},
\end{align}
where $\bs x_t$ is the decision of the algorithm at time $t$ and $\tilde O(\cdot)$ hides polynomial factors of $\log n$. $\mathbb{I}\{\cdot \}$ is the boolean indicator function assuming values in $\{0,1 \}$.
\end{restatable}

\begin{remark}(\textbf{relaxed assumptions \& improvements})
In \citep{improperDynamic}, it is assumed that the losses are gradient Lipschitz and exp-concave over an enlarged set $\cD^\dagger = \{\bs x: \| \bs x\|_\infty \le B+G\}$ where $B$ and $G$ are as in Assumptions B1-B2. While our proper learning results doesn't require gradient Lipschitzness and require exp-concavity to hold in the smaller constraint set $\cD$ as in Assumption B1. Further \citep{improperDynamic} attains a worse dependence of $O(d^{3.5})$ in the non-trivial regime $C_n \ge 1/n$.
\end{remark}

Further, we show in Appendix \ref{app:reparam} that when the decision set is a polytope satisfying certain conditions, we can reparametrize the original problem into the framework of box constrained online learning with exp-concave losses.

\subsection{Road map for the proof of Theorem \ref{thm:ec-d}} \label{sec:road-ec}
The proof of Theorem \ref{thm:ec-d} is facilitated by generalising the arguments used for proving Theorem \ref{thm:main-sq}. We first form a coarse partition of $[n]$ namely $\cP$ in Lemma \ref{lem:key-part-multi} by a direct extension of Lemma \ref{lem:part-sq}. For the regime where dual varaible $\lamda = O(d^{1.25} n^{1/3}/C_n^{1/3})$, we employ a two term regret decomposition for each bin $[i_s,i_t] \in \cP$ as follows:
\begin{align}
    \underbrace{\sum_{j=i_s}^{i_t} f_j(\bs x_j) - f_j(\check{\bs u}_i)}_{T_{1,i}} +  \underbrace{\sum_{j=i_s}^{i_t} f_j(\check{\bs u}_i) - f_j(\bs u_j)}_{T_{2,i}},
\end{align}
where $\bs x_j$ is the prediction of the FLH-ONS algorithm and $\bs u_{1:n}$ is the offline optimal sequence in Lemma \ref{lem:kkt-ec-d}.
We exhibit a choice of $\check{\bs u}_i \in \cD$ in Lemma \ref{lem:low-lamda-multi} so that $T_{1,i} + T_{2,i}$ when summed across all bins $[i_s,i_t] \in \cP$ yield a total regret of $\tilde O^*(n^{1/3}C_n^{2/3} \vee 1)$.

For handling the alternate regime $\lamda  = \Omega(d^{1.25} n^{1/3}/C_n^{1/3})$, we provide a refinement scheme \code{fineSplit} in Fig.\ref{fig:split} in Appendix \ref{app:ec}. Specifically let $\cR$ be the set of all intervals in $\cP$ that satisfy the prerequisite of \code{fineSplit} procedure. Let $\cS := \cP \setminus \cR$.

For each interval in $\cR$, we invoke \code{fineSplit}. This refinement scheme splits the original interval into sub-bins that satisfy either the properties in Lemma \ref{lem:split-mono1} (which can be regarded as a generalization of style U sub-bins in Section \ref{sec:road-sq}) or Lemma \ref{lem:split-mono2} (which can be regarded as a generalization of style V sub-bins in Section \ref{sec:road-sq}). Sub-bins that satisfy condition in Lemma \ref{lem:split-mono1} is termed as style U\textsuperscript{+} sub-bins and those that satisfy condition in Lemma \ref{lem:split-mono2} is termed as style V\textsuperscript{+} sub-bins henceforth for brevity. Sub-bins satisfying conditions of both Lemmas \ref{lem:split-mono1} and \ref{lem:split-mono2} are regarded as style U\textsuperscript{+} sub-bins. For each such sub-bin $[a,b]$, we employ a two term regret decomposition as follows:

\begin{align}
    \underbrace{\sum_{j=a}^b f_j(\bs x_j) - f_j(\check{\bs u}_j)}_{T_1} + \underbrace{\sum_{j=a}^b f_j(\check{\bs u}_j) - f_j(\bs u_j)}_{T_2}. \label{eq:refine-ec}
\end{align}

We term the sequence $\check{\bs u}_{a:b}$ as the \emph{ghost sequence} as they are fictitious intermediate comparator sequence introduced solely for the purpose of analysis. We provide a mechanical way of generating an appropriate ghost sequence in the \code{generateGhostSequence} procedure in Fig.\ref{fig:ghost} which satisfies the properties stated in Lemma \ref{lem:ghost}. Of particular interest is how we choose the ghost sequence for style U\textsuperscript{+} sub-bins. Suppose for a style U\textsuperscript{+} sub-bin $[a,b]$, let $k \in [d]$ be the coordinate where the offline optimal takes the form of Structure 1 or Structure 2 (see Definition \ref{def:struct-gap}). Then we set for all $j \in [a,b]$:
\begin{align}
 \check{\bs u}_j[k] =  \Pi \left(\bs u_a[k] - \frac{1}{(b-a+1)\beta} \sum_{j=a}^b \grad f_j(\bs u_j)[k] \right),   
\end{align}
where $\Pi(\cdot)$ is the projection to $[-B,B]$ and $\beta := G^2$. This choice is very different from the unprojected gradient descent update used in \citep{improperDynamic}. It can be viewed as a lazy projected gradient descent like update (with step size $1/((b-a+1)\beta)$) where the update operation is performed only across coordinate $k$. Note that it is not exactly gradient descent across coordinate $k$ since in the second term above we are using $\grad f_j(\bs u_j)[k]$ instead of $\grad f_j(\bs u_a)[k]$.

The choice of $\check{\bs u}_j[k']$ for  $k' \neq k$ is more involved and is accomplished by carefully selecting a sequence that switches only $O(1)$ times and assumes values in $[(\bs u_a[k'] \wedge \ldots \wedge \bs u_b[k']), (\bs u_a[k'] \vee \ldots \vee \bs u_b[k'])]$ as mentioned in \code{generateGhostSequence} procedure in Fig.\ref{fig:ghost} in Appendix \ref{app:ec}.

Next, by using similar gap criteria used in Section \ref{sec:road-sq} and exploiting gradient Lipschitzness, we show that $T_1 + T_2$ in Eq.\eqref{eq:refine-ec} can be bounded by $O^*(\log n) + \text{ a negative term}$ for each style U\textsuperscript{+} sub-bin obtained by refining bins in $\cR$. For each style V\textsuperscript{+} sub-bin, the regret is bounded by $O^*(\log n)$ (see Lemma \ref{lem:mono-d}). When such bounds are added for all sub-bins generated by invoking \code{fineSplit} on every interval in $\cR$, we show that the negative terms gracefully offset the  culmination of $O^*(\log n)$ terms to result in a regret bound of $\tilde O^*(n^{1/3}C_n^{2/3} \vee 1)$ (see Proof of Lemma \ref{lem:high-lamda-multi}).

The regret contribution from all bins in $\cS$ is bounded by $\tilde O^*(n^{1/3}C_n^{2/3} \vee 1)$ using Lemma \ref{lem:mono-d}. Finally summing the regret contributions from bins in $\cR$ and $\cS$ yield the theorem.

\section{Conclusion and future work}
In this work we presented a new analysis that extends the results of \citep{improperDynamic} and showed near optimal universal dynamic regret in a proper learning setting for strongly convex losses. Results on the special case of exp-concave losses and box decision set are also derived. Further we relaxed the gradient Lipschitzness assumption for losses revealed and derived regret rates with improved dependence on $d$.

An important open problem is to extend these results for exp-concave losses with general convex decision sets.

\section*{Acknowledgments}
The research was partially supported by NSF Award \#2007117 and a start-up grant from UCSB CS
department.

\bibliography{tf,yx}
\bibliographystyle{plainnat}

\newpage

\onecolumn
\appendix

\section{Preliminaries}\label{app:prelims}
For the sake of completeness, we recall the description of Follow-the-Leading-History (FLH) algorithm from \citep{hazan2007adaptive}.

\begin{figure}[h!]
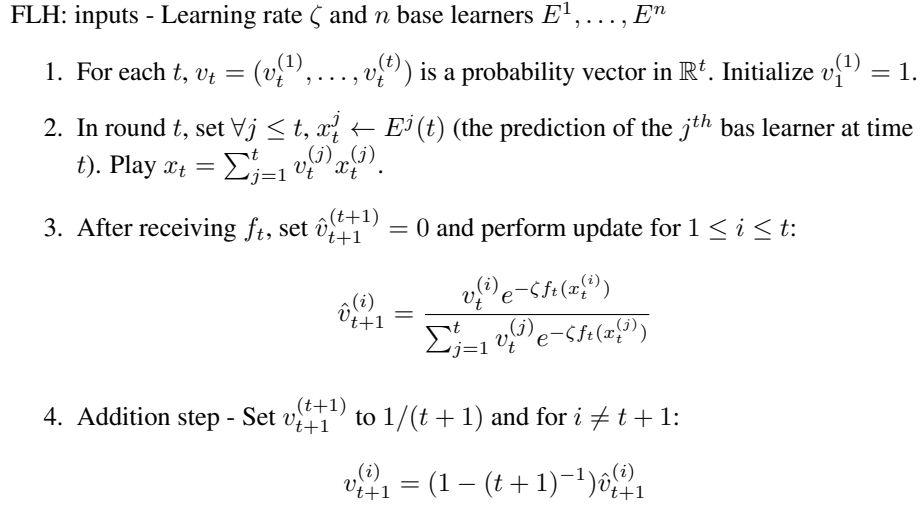

	\centering
	\fbox{
		\begin{minipage}{12 cm}
		FLH: inputs - Learning rate $\zeta$ and $n$ base learners $E^1,\ldots,E^n$
            \begin{enumerate}
                \item For each $t$, $v_t = (v_t^{(1)},\ldots,v_t^{(t)})$ is a probability vector in $\mathbb{R}^t$. Initialize $v_1^{(1)} = 1$.
                \item In round $t$, set $\forall j \le t$, $x_t^j \leftarrow E^j(t)$ (the prediction of the $j^{th}$ bas learner at time $t$). Play $x_t =  \sum_{j=1}^t v_t^{(j)}x_t^{(j)}$.
                \item After receiving $f_t$, set $\hat v_{t+1}^{(t+1)} = 0$ and perform update for $1 \le i \le t$:
                \begin{align}
                    \hat v_{t+1}^{(i)}
                    &= \frac{v_t^{(i)}e^{-\zeta f_t(x_t^{(i)})}}{\sum_{j=1}^t v_t^{(j)}e^{-\zeta f_t(x_t^{(j)})}}
                \end{align}
                \item Addition step - Set $v_{t+1}^{(t+1)}$ to $1/(t+1)$ and for $i \neq t+1$:
                \begin{align}
                    v_{t+1}^{(i)}
                    &= (1-(t+1)^{-1}) \hat v_{t+1}^{(i)}
                \end{align}
            \end{enumerate}
		\end{minipage}
	}
	\caption{FLH algorithm}
	\label{fig:flh}
\end{figure}

FLH enjoys the following guarantee against any base learner.
\begin{proposition}\label{prop:flh} \citep{hazan2007adaptive}
Suppose the loss functions are exp-concave with parameter $\alpha$. For any interval $I = [r,s]$ in time, the algorithm FLH Fig.\ref{fig:flh} with learning rate $\zeta = \alpha$ gives $O(\alpha^{-1}( \log r + \log |I|))$ regret against the base learner in hindsight.
\end{proposition}

\begin{definition} (\citep{daniely2015strongly}) \label{def:sa}
An algorithm is said to be Strongly Adaptive (SA) if for every contiguous interval $I \subseteq [n]$, the static regret incurred by the algorithm is $O(\text{poly}(\log n) \Gamma^*(|I|))$ where $\Gamma^*(|I|)$ is the value of minimax static regret incurred in an interval of length $|I|$.
\end{definition}

It is known from \citep{hazan2007logregret} that OGD and ONS achieves static regret of $O(\log n)$ and $O(d \log n)$ for strongly convex and exp-concave losses respectively. Hence in view of Proposition \ref{prop:flh} and Definition \ref{def:sa}, we can conclude that:
\begin{itemize}
    \item FLH with OGD as base learners is an SA algorithm for strongly convex losses.
    \item FLH with ONS as base learners is an SA algorithm for exp-concave losses. (We treat dimension $d$ as a constant problem parameter and consider minimaxity only wrt $n$.)
\end{itemize}

We have the following guarantee on runtime.
\begin{proposition} \citep{hazan2007adaptive}
Let $\rho$ be the per round run time of base learners and $r_n$ be the static regret suffered by the base learners over $n$ rounds. Then FLH procedure has a runtime of $O(\rho n)$ per round. To improve the runtime one can use AFLH procedure from \citep{hazan2007adaptive} that incurs $O(\rho \log n)$ runtime overhead per round and suffers $O(r_n \log n)$ static regret in any interval.
\end{proposition}

Similar runtime improvements at the expense of blowing up the regret by a factor of $\log n$ can also be obtained from the IFLH algorithm of \citep{zhang2018dynamic}.

\section{Proofs for Section \ref{sec:square-game}} \label{app:square-games}

We start by characterizing the offline optimal.
Define the sign function as $\sign{(x)} = 1 \text{ if } x > 0$; $ -1  \text{ if } x < 0$; and some $v \in [-1,1]  \text{ if } x = 0$. We start by presenting a sequence of useful lemmas.

\lemkktsq*
\begin{proof}
We can form the Lagrangian of the optimization problem as:
\begin{align}
    \mathcal{L}(\tilde{u}_{1:n} , \tilde{z}_{1:n-1}, \tilde{\bs v}, \tilde{\lambda}, \tilde{\gamma}_{1:n}^+, \tilde{\gamma}_{1:n}^-)
    &= \frac{1}{2}\sum_{t=1}^{n} (y_t - \tilde u_t)^2 + \tilde \lambda \left( \sum_{t=1}^{n-1} |\tilde z_t| - C_n\right) + \sum_{t=1}^{n-1} \tilde v_t( \tilde u_{t+1} - \tilde u_t - \tilde z_t)\\
    &\quad + \sum_{t=1}^n \tilde \gamma_t^- (-B - \tilde u_t) + \tilde \gamma_t^+ (\tilde u_t -B),
\end{align}
for dual variables $\tilde \lambda > 0$, $\tilde {v}_{1:n}$ unconstrained, $\tilde \gamma_{1:n}^- \ge 0$ and $\tilde \gamma_{1:n}^+ \ge 0$. Let $(u_{1:n}, z_{1:n}, v_{1:n}, \lambda, \gamma_{1:n}^-, \gamma_{1:n}^+)$ be the optimal primal and dual variables. By stationarity conditions (via the derivative wrt $u_t$), we have:
\begin{align}
    u_t - y_t + v_{t-1} - v_t - \gamma_t^- + \gamma_t^+ = 0,
\end{align}
where we take $v_0 = v_n = 0$. Stationarity conditions via derivative wrt $z_t$ yields
\begin{align}
    v_t = \lambda s_t.
\end{align}
Combining the above two equations and the complementary slackness rules yields the lemma.
\end{proof}

\begin{example}\label{ex}
We describe the example used to create Fig.\ref{fig:large-t2}. We adopt the notations of Lemma \ref{lem:kkt-sq}.
\begin{itemize}
    \item $G = 4$ and $B = 2$.

    \item For each $k \in [0,\frac{n^{1/4}}{2}-1]$, $u_j = B- \frac{1}{2n^{3/4}}$ for all $j \in [2kn^{3/4}+1, (2k+1)n^{3/4}]$.
    
    \item For each $k \in [0,\frac{n^{1/4}}{2}-1]$, $u_j = B$ for all $j \in [(2k+1)n^{3/4}+1, (2k+2)n^{3/4}]$.
    
    \item $y_1 = y_{n^{3/4}} = B - \frac{1}{2n^{3/4}} - \frac{n^{3/4}-2}{n}$. $y_j = B - \frac{1}{2n^{3/4}} - (1 - 2/n)$ for all $ j \in [2,n^{3/4}-1]$.
    
    \item For each $k \in [1,\frac{n^{1/4}}{2}-1]$, $y_{2kn^{3/4}+1} = y_{(2k+1)n^{3/4}} = B -\frac{1}{2n^{3/4}} - \frac{n^{3/4}-2}{n}$. $y_j = B -\frac{1}{2n^{3/4}} - (1 - 2/n)$ for all $j \in [2kn^{3/4}+2, (2k+1)n^{3/4}-1]$.
    
    \item For each $k \in [0,\frac{n^{1/4}}{2}-1]$, $y_j = G$ for all $j \in [(2k+1)n^{3/4}+1, (2k+2)n^{3/4}]$.
    
    \item $\gamma_j^- = 0$ for all $ j \in [n]$.
    
    \item  For each $k \in [0,\frac{n^{1/4}}{2}-1]$, $\gamma_j^+ = 0$ for all $j \in [2kn^{3/4}+1, (2k+1)n^{3/4}]$.
    
    \item For each $k \in  [0,\frac{n^{1/4}}{2}-2]$, $\gamma_{(2k+1)n^{3/4}+1}^+ = \gamma_{(2k+2)n^{3/4}}^+ = G - B - \frac{n^{3/4}-2}{n}$. $\gamma_j^+ = G - B - 2(1 - 1/n)$ for all $j \in [(2k+1)n^{3/4}+2, (2k+2)n^{3/4}-1]$.
    
    \item $\gamma_{n-n^{3/4}+1}^+ = \gamma_n^+ = G - B - \frac{n^{3/4}-2}{n}$.
    
    \item $\lamda = n^{3/4}-2$.
    
    \item $s_t = 1/n + (t-1) \frac{1 - 2/n}{n^{3/4}-2}$ for $1 \le t \le n^{3/4}-1$. $s_{n^{3/4}} = 1$.
    
    \item For each $k \in [0, \frac{n^{1/4}}{2} - 2]$, $s_t = 1 - 1/n + (t-1 - (2k+1)n^{3/4}) \frac{2/n - 2}{n^{3/4}-2}$ for $(2k+1)n^{3/4} + 1 \le t \le (2k+2)n^{3/4}-1$. $s_{(2k+2)n^{3/4}} = -1$.
    
    \item For each $k \in [1, \frac{n^{1/4}}{2} - 1]$, $s_t = -1 + 1/n + (t-1-2kn^{3/4}) \frac{2 - 2/n}{n^{3/4} - 2}$. $s_{(2k+1)n^{3/4}} = 1$.
    
    \item $s_t = 1-1/n + (t-1-n+n^{3/4}) \frac{2/n-1}{n^{3/4}-1}$ for $n - n^{3/4} +1 \le t \le n-1$. $s_n = 0$.
    
\end{itemize}
\end{example}

\textbf{Terminology.} We will refer to the optimal primal variables $u_1,\ldots,u_n$ in Lemma \ref{lem:kkt-sq} as the \emph{offline optimal solution} in this section. For two natural numbers $a < b$, we denote $[a,b] = \{a, a+1,\ldots,b \}$.

\defoned*

The following Lemma plays a central role in the analysis. Qualitatively, it captures a fundamental way in which the adversary is constrained.

\begin{restatable}{lemma}{lemLamdaLen} ($\bs \lamda$\textbf{-length lemma}) \label{lem:lamda-len}
Suppose that the offline optimal solution sequence takes the form of Structure 1 or Structure 2 in an interval $[j,j+\ell - 1]$ for some $\ell > 0$ and $j \in \{2,\ldots,n-1\}$. Then $\lamda \le \frac{(B+G)\ell}{2}$.
\end{restatable}
\begin{proof}
We consider the case of Structure 2. Arguments are similar for case  of Structure 1. Let the optimal sign assignments be written as $s_{j+k-1} = -1 + \epsilon_k$ where $\epsilon_k \in [0,2]$ for all $k \in [\ell-1]$. From the KKT conditions, we have

\begin{align*}
    y_j &= u - \lamda \epsilon_1\\
    y_{j+1} &= u - \lamda (\epsilon_2 - \epsilon_1)\\
    &\vdots\\
    y_{j+\ell-2} &= u - \lamda (\epsilon_{\ell-1} - \epsilon_{\ell-2})\\
    y_{j+\ell-1} &= u - \lamda (2-\epsilon_{\ell-1})
\end{align*}

Consider a vector $\bs z = [\epsilon_1,\epsilon_2 - \epsilon_1, \ldots, 2-\epsilon_{\ell-1}]^T$. Note that the condition $\| \bs z\|_\infty > 0$ is always satisfied. Otherwise we must have $2 = \epsilon_{\ell-1} = \ldots = \epsilon_1$. But $\epsilon_1 = 2$ makes $\| \bs z\|_\infty > 0$ yielding a contradiction.

Let $k^*$ be such that $|\bs z[k^*]| = \| \bs z\|_\infty$.  Since $\lamda \ge 0$, we can write $\lamda = \frac{|y_{j+k^*-1} - u|}{\| \bs z\|_\infty}$. Since $|y_{j+k^*-1} - u|$ is bounded, a lower bound on ${\|\bs z\|_\infty}$ will yield an upper bound on $\lamda$. To this end, we consider the following optimization problem:

\begin{mini!}|s|[2]                   
    {t,\epsilon_{1},\ldots,\epsilon_{\ell-1}}                               
    {t}   
    {\label{eq:Example1}}             
    {}                                
    \addConstraint{0 \le \epsilon_i \le 2 \: \forall i \in [\ell-1],}    
    \addConstraint{\epsilon_1 \le t,}  
    \addConstraint{|\epsilon_{i+1} - \epsilon_{i} | \le t \: \forall i \in [\ell-2],}
    \addConstraint{2 - \epsilon_{\ell-1} \le t}
\end{mini!}

We can form the Lagrangian as:

\begin{align}
    \mathcal L(t,\epsilon_{1:\ell-1},a_{1:\ell-1}, b_{1:\ell-1}, c_{1:\ell-2}, d_{1:\ell-2}, e_1, e_{\ell-1})
    &= t - \sum_{i=1}^{\ell-1} a_i \epsilon_i + \sum_{i=1}^{\ell-1} b_i (\epsilon_i - 2) \\
    &+\sum_{i=1}^{\ell-2} c_i (-t - \epsilon_{i+1} + \epsilon_i) + \sum_{i=1}^{\ell-2} d_i (\epsilon_{i+1} - \epsilon_i - t)\\
    &+ e_1(\epsilon_1 - t) + e_2(2-\epsilon_{\ell-1} - t)
\end{align}

Stationarity conditions are:

\begin{align}
    \frac{\partial {\mathcal L}}{ \partial t} = 0 & \implies 1 + \sum_{i=1}^{\ell-2} - c_i - d_i - e_1 - e_2 = 0\\
    \frac{\partial {\mathcal L}}{ \partial \epsilon_1} = 0 & \implies -a_1 + b_1 + c_1 - d_1 + e_1 = 0\\
    \frac{\partial {\mathcal L}}{ \partial \epsilon_{\ell-1}} = 0 & \implies -a_{\ell-1} + b_{\ell-1} - c_{\ell-2} + d_{\ell-2} - e_2 = 0\\
    \frac{\partial {\mathcal L}}{ \partial \epsilon_i} = 0 & \implies -a_i + b_i - c_{i-1} + c_i + d_{i-1} - d_i = 0, \: \text{where } i \in \{2,\ldots, \ell-2 \}
\end{align}

Complementary slackness conditions are:
\begin{align}
    a_i \epsilon_i &= 0, \: i \in [\ell-1]\\
    b_i (\epsilon_i - 2) &= 0, \: i \in [\ell-1]\\
    c_i (-t - \epsilon_{i+1} + \epsilon_i) &= 0, \: i \in [\ell-2] \\
    d_i (\epsilon_{i+1} - \epsilon_i - t) &= 0, \: i \in [\ell-2] \\
    e_1(\epsilon_1 - t) &= 0\\
    e_2(2-\epsilon_{\ell-1} - t) &= 0
\end{align}

Dual feasibility conditions are $a_i \ge 0,\: b_i \ge 0$ for $i \in [\ell-1]$ and $c_i \ge  0,\: d_i \ge 0$ for $i \in [\ell-2]$ and $e_1 \ge 0, \: e_2 \ge 0$.

Primal feasibility conditions are given by the constraint set of the optimization problem.

Now we form a guess for optimal primal and dual variables as $t = 2/\ell$ and $\epsilon_i = 2i/\ell$ for $i \in [\ell-1]$ and $a_i = b_i = 0$ for $i \in [\ell-1]$ and $c_i = 0$ for $i \in [\ell-2]$ and $e_1 = e_2 = d_1 = \ldots = d_{\ell-2} = 1/\ell$. All the KKT conditions can be readily verified for this solution guess.

Recall that, earlier we defined $\bs z = [\epsilon_1,\epsilon_2 - \epsilon_1, \ldots, 2-\epsilon_{\ell-1}]^T$ and $\lamda = \frac{|y_{j+k^*-1} - u|}{\| \bs z\|_\infty}$ where $k^*$ is such that $|\bs z[k^*]| = \| \bs z\|_\infty$. By the previous optimization problem we deduce that $\|\bs z \|_\infty \ge 2/\ell$. Since $|y_{j+k^*-1} - u| \le B+G$, we conclude that $\lamda \le (B+G)\ell/2$

\end{proof}

Next, we exhibit a useful partitioning scheme of the interval $[n]$.
\begin{restatable}{lemma}{lempartsq}\label{lem:part-sq}(\citep{improperDynamic})(\textbf{key partition}) 
Initialize $\cP \leftarrow \Phi$. Starting from time 1, spawn a new bin $[i_s,i_t]$ whenever $\sum_{j=i_s+1}^{i_t+1} | u_j -  u_{j-1} | > B/\sqrt{n_i}$, where $n_i = i_t - i_s + 2$. Add the spawned bin $[i_s,i_t]$ to $\cP$.

Let $M:=|\cP|$. We have $M = O\left (1 \vee n^{1/3}C_n^{2/3} B^{-2/3} \right)$.
\end{restatable}

\textbf{Notations.} For bin $[i_s,i_t] \in \cP$ we define: $n_i = i_t - i_s + 1$, $\bar u_i = \frac{1}{n_i}\sum_{j=i_s}^{i_t} u_j$, $\bar y_i = \frac{1}{n_i}\sum_{j=i_s}^{i_t} y_j$, $\Gamma_i^+ = \sum_{j=i_s}^{i_t} \gamma_j^+$, $\Gamma_i^- = \sum_{j=i_s}^{i_t} \gamma_j^-$, $\Delta s_i = s_{i_t} - s_{i_s-1}$, $C_i = \sum_{j=i_s+1}^{i_t} |u_j - u_{j-1}|$.

For any general bin $[a,b]$ define the quantities $n_{a \rightarrow b},\bar u_{a \rightarrow b}, \bar y_{a \rightarrow b}, \Gamma^+_{a \rightarrow b}, \Gamma^-_{a \rightarrow b},\Delta s_{a \rightarrow b},C_{a \rightarrow b}$ analogously as above.

Next we calculate the static regret guarantee of the FLH-ONS strategy.
\begin{restatable}{lemma}{lemStatOgd}\label{lem:stat-ogd} (\citep{hazan2007logregret}, \citep{hazan2007adaptive})
Consider a bin $[a,b] \subseteq [n]$ and a point $w \in [-B,B]$. Under the setting of Theorem \ref{thm:main-sq} we have
\begin{align}
    \sum_{t=a}^b (y_t - x_t)^2 - (y_t - w)^2
    &\le 10(B+G)^2 \log n\\
    &= \tilde O(1),
\end{align}
where $x_t$ are the predictions of FLH-OGD.
\end{restatable}
\begin{proof}
The losses $(y_t- x)^2$ are strongly convex with parameter 2. Further the gradients are bounded by $2(G+B)$. Hence by Theorem 1 in \citep{hazan2007logregret} we have the static regret guarantee of OGD being $4(G+B)^2 \cdot (2\log n) / 4 = 2(G+B)^2 \log n$.

The losses $(y_t- x)^2$ are $1/(2(G+B)^2)$ exp-concave. So by applying Theorem 3.2 in \citep{hazan2007adaptive} we have the regret of FLH against any base experts bounded as $8 (G+B)^2 \log n$. 

Adding these regret bounds yields the lemma.

\end{proof}

\begin{restatable}{lemma}{lemLowLamda} (\textbf{low} $\bs \lamda$ \textbf{regime})\label{lem:low-lamda}
If the optimal dual variable $\lamda = O\left( \frac{n^{1/3}}{C_n^{1/3}} \right)$, we have the regret of FLH-OGD strategy bounded as
\begin{align}
    \sum_{t=1}^{n} (y_t - x_t)^2 - (y_t - u_t)^2 
    &= \tilde O(n^{1/3}C_n^{2/3} \vee 1),
\end{align}
where $x_t$ is the prediction of FLH-OGD at time $t$.
\end{restatable}
\begin{proof}

Throughout this proof, the bins $[i_s,i_t]$ we consider belong to the partition $\mathcal P$.

\textbf{Case 1: } When the offline optimal solution touches the boundary $B$ within a bin $[i_s,i_t]$. We use a three term regret decomposition as follows. 

\begin{align}
    \underbrace{\sum_{j=i_s}^{i_t} (y_j - x_j)^2 - (y_j - B)^2}_{T_{1,i}} + \underbrace{\sum_{j=i_s}^{i_t} (y_j - B)^2 - (y_j - \bar u_i)^2}_{T_{2,i}} + \underbrace{\sum_{j=i_s}^{i_t} (y_j - \bar u_i)^2 - (y_j -  u_j)^2}_{T_{3,i}}
\end{align}

Now $T_{1,i} = O(\log n)$ by strong adaptivity of FLH. Observe that due to complementary slackness, $\gamma_j^- = 0$ uniformly within the bin since the TV within the bin is at-most $B/\sqrt{n_i} < 2B$ and hence the solution never touches $-B$ boundary within this bin. By using the KKT conditions, we have $y_j = u_j - \lamda (s_j - s_{j-1}) + \gamma_j^+$. So

\begin{align}
    T_{2,i}
    &= \sum_{j=i_s}^{i_t} (\bar u_i - B)^2 + 2(y_j - \bar u_i) (\bar u_i - B)\\
    &= n_i (\bar u_i - B)^2 + 2n_i (\bar y_i - \bar u_i) (\bar u_i - B)\\
    &\le_{(a)} B^2 + 2 (\bar u_i - B) (\Gamma_i^+ - \lamda \Delta s_i)\\
    &\le_{(b)} B^2 + 4 \lamda C_i + 2 \Gamma_i^+ (\bar u_i - B) \label{eq:low-lamda-1}
\end{align}
where in line (a) we used KKT conditions and $|\bar u_i - B| \le B/\sqrt{n_i}$ due to the TV constraint within bin and in line (b) we used: (i) $|\bar u_i - B| \le C_i$ as the optimal solution assumes the value $B$ at some time point in $[i_s,i_t]$ (ii) $|\Delta s_i | \le 2$.

We have
\begin{align}
    T_{3,i}
    &= \sum_{j=i_s}^{i_t} (u_j - \bar u_i)^2 + 2(y_j - u_j)(u_j - \bar u_i)\\
    &\le n_iC_i^2 + 2\sum_{j=i_s}^{i_t} (-\lamda(s_j - s_{j-1}) + \gamma_j^+)(u_j - \bar u_i)\\
    &=_{(a)} n_iC_i^2 + 2\lamda(s_{i_s-1}(u_{i_s} - \bar u_i) - s_{i_t}(u_{i_t} - \bar u_i)) + 2\lamda C_i +  2\sum_{j=i_s}^{i_t} \gamma_j^+ (u_j - \bar u_i)\\
    &=_{(b)} n_iC_i^2 + 2\lamda(s_{i_s-1}(u_{i_s} - \bar u_i) - s_{i_t}(u_{i_t} - \bar u_i)) + 2\lamda C_i + 2 \Gamma_i^+ (B-\bar u_i)\\
    &\le_{(c)} B^2 + 6 \lamda C_i + 2 \Gamma_i^+ (B-\bar u_i), \label{eq:last-term}
\end{align}
where line (a) is obtained by a rearrangement of the sum and line (b) is obtained by the complementary slackness condition which states that $\gamma_j^+ = 0$ if $u_j < B$. Line (c) is obtained by $|u_j - \bar u_i| \le C_i$ for any $j \in [i_s,i_t]$ and by applying triangle inequality.

So overall we can bound the regret within this bin by adding Eq.\eqref{eq:low-lamda-1} and \eqref{eq:last-term} with $T_{1,i} = O(\log n)$ as
\begin{align}
    T_{1,i} + T_{2,i} + T_{3,i}
    &\le O(\log n) + 2B^2 + 10 \lamda C_i. \label{eq:1}
\end{align}

\textbf{Case 2:} When the offline optimal solution touches boundary $-B$ within a bin $[i_s,i_t]$. This case can be treated similar to Case 1.

\textbf{Case 3:} When the offline optimal solution doesn't touch either boundaries within a bin $[i_s,i_t]$. Here we use a two term regret decomposition as 
\begin{align}
    \underbrace{\sum_{j=i_s}^{i_t} (y_j - x_j)^2 - (y_j - \bar u_i)^2}_{T_{1,i}} + \underbrace{\sum_{j=i_s}^{i_t} (y_j - \bar u_i)^2 - (y_j -  u_j)^2}_{T_{2,i}}.
\end{align}

By following the analysis used in obtaining the bound of Eq.\eqref{eq:last-term} (where we use $\gamma_j^- = \gamma_j^+ = 0$ due to complementary slackness), we obtain
\begin{align}
    T_{1,i} + T_{2,i} 
    &\le O(\log n) + B^2 + 6 \lamda C_i \label{eq:2}
\end{align}

By summing up the regret bounds which assumes the form in Eq.\eqref{eq:1} (for Case 1 and 2) or Eq.\eqref{eq:2} (for Case 3) across all bins in the partition $\cP$, we obtain the overall regret as
\begin{align}
    \sum_{t=1}^{n} (y_t - x_t) ^2 - (y_t - u_t)^2
    &\le O(|\cP| \log n B^2) + 2B^2 |\cP| + 10 \lamda C_n\\
    &= \tilde O(n^{1/3}C_n^{2/3} \vee 1),
\end{align}
where in the last line we used the fact that $|\cP| = O(n^{1/3}C_n^{2/3} \vee 1)$ and $\lamda = O((n/C_n)^{1/3})$ by the premise of the lemma.
\end{proof}

\begin{figure}[h!]
\centering
\includegraphics[width=10cm,height=4cm]{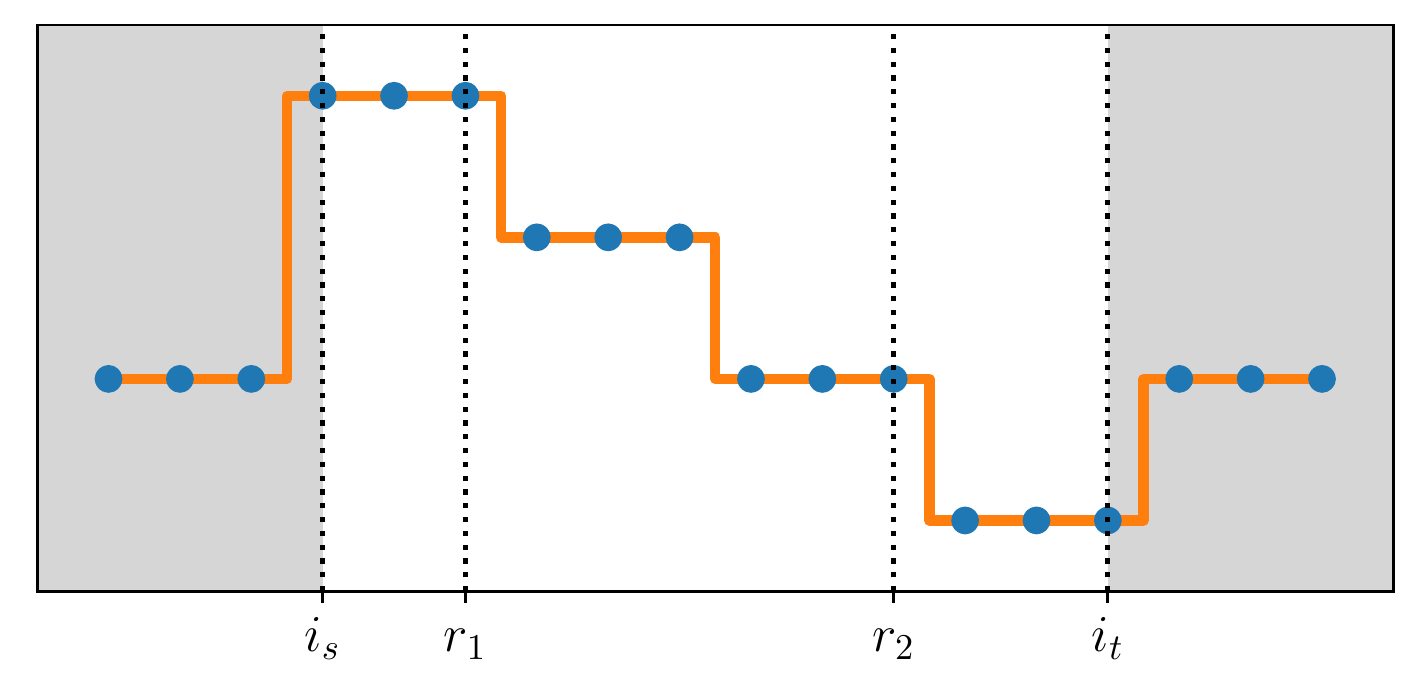}
    \caption{\emph{An example of a configuration referred in the proof of Lemma \ref{lem:mono-1d}. Here $s_{r_1} = s_{r_2} = -1$ and the sequence is non-increasing within $[r_1+1,r_2]$.}}    
    \label{fig:mono}
\end{figure}

\begin{restatable}{lemma}{lemMonoOned} (\textbf{monotonic sequence}) \label{lem:mono-1d}
Consider a bin $[i_s,i_t] \in \cP$ such that the offline optimal solution is monotonic within this bin. Then the regret of FLH-OGD strategy within this bin is at-most $31(B+G)^2 \log n = O(\log n)$.
\end{restatable}
\begin{proof}
When the optimal sequence is monotonic within a bin $[i_s,i_t] \in \cP$, it is always possible to form \emph{at-most} 3 bins: $[i_s,r_1]$, $[r_1+1,r_2]$, $[r_2+1,i_t]$ such that the offline optimal solution is constant within bins $[i_s,r_1]$ and $[r_2+1,i_t]$ alongside the condition that the bin $[r_1+1,r_2]$ satisfies one of the following properties: a) $s_{r_1} = s_{r_2} = 1$ and the offline optimal solution is non-decreasing within bin $[r_1+1,r_2]$ or b) $s_{r_1} = s_{r_2} = -1$ and the offline optimal solution is non-increasing within bin $[r_1+1,r_2]$. (see for eg. Fig.\ref{fig:mono}).

Due to Lemma \ref{lem:stat-ogd}, the regret within bins $[i_s,r_1]$ and $[r_2+1,i_t]$ is at-most $ 10(B+G)^2 \log n$ each. Note that this three sub-bin refinement can make sure that the offline optimal solution doesn't touch the boundaries $\pm B$ within the bin $[r_1+1, r_2]$. We bound the regret within bin $[r_1+1, r_2]$ via a two term regret decomposition as follows. 

\begin{align}
    \underbrace{\sum_{j=r1+1}^{r_2} (y_j - x_j)^2 - (y_j - \bar u_{r1+1 \rightarrow r_2})^2}_{T_{1}} + \underbrace{\sum_{j=r1+1}^{r_2} (y_j - \bar u_{r1+1 \rightarrow r_2})^2 - (y_j -  u_j)^2}_{T_{2}}.
\end{align}

We have $T_{1} \le 10(B+G)^2 \log n $. Further due to KKT conditions we have,

\begin{align}
    T_2
    &= \sum_{j=r1+1}^{r_2} (u_j - \bar u_{r1+1 \rightarrow r_2})(2y_j - u_j - \bar u_{r1+1 \rightarrow r_2})\\
    &= \sum_{j=r1+1}^{r_2} (u_j - \bar u_{r1+1 \rightarrow r_2})(2y_j - 2 u_j + u_j - \bar u_{r1+1 \rightarrow r_2})\\
    &= \sum_{j=r1+1}^{r_2} (u_j - \bar u_{r1+1 \rightarrow r_2})^2 + 2 \lamda (u_j - \bar u_{r1+1 \rightarrow r_2}) (s_{j-1} - s_j)\\
    &\le n_i C^2_i + \sum_{j=r1+1}^{r_2} 2 \lamda (u_j - \bar u_{r1+1 \rightarrow r_2}) (s_{j-1} - s_j),
\end{align}
where in the last line we used $|u_j - \bar u_{r1+1 \rightarrow r_2}| \le C_i$. We also have $n_i C_i^2 \le B^2$ by the construction in Lemma \ref{lem:part-sq}.

By expanding the second term followed by a regrouping of terms in the summation, we can write
\begin{align}
    \sum_{j=r1+1}^{r_2} 2 \lamda (u_j - \bar u_{r1+1 \rightarrow r_2}) (s_{j-1} - s_j)
    &=2\lamda \left(s_{r_1}(u_{r_1+1} - \bar u_{r1+1 \rightarrow r_2}) - s_{r_2}(u_{r_2} - \bar u_{r1+1 \rightarrow r_2}) \right)\\
    &\quad+ 2\lamda \sum_{j=r_1+2}^{r_2} |u_j - u_{j-1}| \\
    &= 2\lamda C_{r_1+1 \rightarrow r_2}\\
    &\quad + 2\lamda \left(s_{r_1}(u_{r_1+1} - \bar u_{r1+1 \rightarrow r_2}) - s_{r_2}(u_{r_2} - \bar u_{r1+1 \rightarrow r_2}) \right). \label{eq:mono-inter}
\end{align}

Since $s_{r_1} = s_{r_2} = 1$ if the offline optimal is non-decreasing in $[r_1+1,r_2]$ or $s_{r_1} = s_{r_2} = -1$ if the offline optimal is non-increasing in $[r_1+1,r_2]$, we have $s_{r_1}u_{r_1+1} - s_{r_2}u_{r_2} = -|u_{r_1+1} - u_{r_2}| = -C_{r_1+1 \rightarrow r_2}$. Hence we see that the second term exactly cancels with the first term in Eq.\eqref{eq:mono-inter}.

Thus overall we have shown that the total regret in $[i_s,i_t]$ is at-most $31(B+G)^2 \log n$.
\end{proof}

\begin{restatable}{lemma}{lemWithinOned} \label{lem:within1d}
Suppose there exists an interval $[a,b]$ (which may not belong to $\cP$) with length $\ell$ such that the optimal sequence takes the form of Structure 1 or Structure 2 within $[a,b]$. Assume that $\bar y_{a \rightarrow b} \in [-B,B]$. Then the regret of FLH-OGD within the bin $[a,b]$ at-most $10(B+G)^2 \log n  - \frac{4 \lamda^2}{\ell}$.
\end{restatable}
\begin{proof}
We use a two term regret decomposition as follows:
\begin{align}
    \underbrace{\sum_{j=a}^{b} (y_j - x_j)^2 - (y_j - \bar y_{a \rightarrow b})^2}_{T_{1}} + \underbrace{\sum_{j=a}^{b} (y_j - \bar y_{a \rightarrow b})^2 - (y_j -  u_j)^2}_{T_{2}}.
\end{align}

By the Definition \ref{def:struct-gap-1d} of Structure 1 and 2, the offline optimal solution is constant within bin $[a,b]$. We denote $u_j = u$ for all $j \in [a,b]$. Further $|\Delta s_{a \rightarrow b} = 2|$. We have,
\begin{align}
    T_2
    &= -\ell (\bar y_{a \rightarrow b} - u)^2 - 2 \sum_{j=a}^b (y_j - \bar y_{a \rightarrow b}) (\bar y_{a \rightarrow b} - u)\\
    &=  -\ell (\bar y_{a \rightarrow b} - u)^2 \\
    &=_{(a)} -\frac{-\lamda^2 (\Delta s_{a \rightarrow b})^2}{\ell}\\
    &= - \frac{4 \lamda^2}{\ell},
\end{align}
where line (a) is obtained by the KKT conditions $y_j = u - \lamda (s_j - s_{j-1})$ for all $j \in [a,b]$ and hence $\bar y_{a \rightarrow b} = u - \frac{\lamda \Delta s_{a \rightarrow b}}{\ell}$.

Due to Lemma \ref{lem:stat-ogd}, we have $T_1 \le 10(B+G)^2 \log n$. Combining both bounds yields the lemma.
\end{proof}

\begin{restatable}{lemma}{lemOutsideOned} \label{lem:outside1d}
Consider a bin $[a,b]$ with length $\ell$.

\textbf{Case 1:} When offline optimal takes the form of Structure 1 within this bin and $\bar y_{a \rightarrow b} \ge B$, then
\begin{align}
    \sum_{j=a}^b (y_j - x_j)^2 - (y_j - u_j)^2
    &\le 10(B+G)^2 \log n - \ell (B-u_a)^2,
\end{align}
and

\textbf{Case 2:} When offline optimal takes the form of Structure 2 within this bin and $\bar y_{a \rightarrow b} \le -B$, then
\begin{align}
    \sum_{j=a}^b (y_j - x_j)^2 - (y_j - u_j)^2
    &\le 10(B+G)^2 \log n - \ell (B+u_a)^2,
\end{align}
where $x_j$ are the predictions of the FLH-OGD algorithm.
\end{restatable}
\begin{proof}
We consider Case 2. Arguments for Case 1 are similar. We employ a two term regret decomposition as follows.

\begin{align}
    \underbrace{\sum_{j=a}^{b} (y_j - x_j)^2 - (y_j + B)^2}_{T_{1}} + \underbrace{\sum_{j=a}^{b} (y_j + B)^2 - (y_j -  u_j)^2}_{T_{2}}.
\end{align}

By Definition \ref{def:struct-gap-1d}, the offline optimal solution is constant within bin $[a,b]$. So we have $u_j = u_a$ for all $j \in [a,b]$. From the KKT conditions, we have
\begin{align}
    T_2
    &= \sum_{j=a}^b (u_a + B)^2 + 2 (y_j - u_a) (u_a+B)\\
    &= \ell (u_a+B)^2 - 2 \lamda \Delta s_{a \rightarrow b} (u_a+ B)\\
    &= \ell (u_a+B)^2 - 4 \lamda (u_a+B), \label{eq:gap1}
\end{align}
where in the last line we used $\Delta s_{a \rightarrow b} = 2$ for Structure 2. From the premise of the lemma for Case 2, we have $\bar y_{a \rightarrow b} \le -B$. Since $\bar y_{a \rightarrow b} = u_a - 2\lamda/\ell $, we must have
\begin{align}
    \bar y_{a \rightarrow b} \le -B &\implies \lamda \ge \frac{\ell}{2}(u_a+B).
\end{align}

Plugging this lower bound to Eq.\eqref{eq:gap1} and noting that $u_a+B \ge 0$, we get
\begin{align}
    T_2 &\le -\ell (u_a+B)^2.
\end{align}

By Lemma \ref{lem:stat-ogd}, we have $T_1 \le 10(B+G)^2 \log n$. Now summing $T_1$ and $T_2$ results in the lemma.

\end{proof}

\begin{restatable}{lemma}{lemLargeMargin} (\textbf{large margin bins}) \label{lem:large-margin}
Assume that $\lamda \ge \phi \frac{n^{1/3}}{c_n^{1/3}}$ for some constant $\phi$ that do not depend on $n$ and $C_n$. Consider a bin $[i_s,i_t] \in \cP$ within which the offline optimal solution takes the form of Structure 1 or Structure 2 (or both) for some appropriate sub-intervals of $[i_s,i_t]$. Let $\mu_{\text{th}} = \sqrt{\frac{36(B+G)^3 C_n^{1/3} \log n}{ \phi n^{1/3}}}$. Then $\text{gap}_{\text{min}}(-B,[i_s,i_t]) \vee \text{gap}_{\text{min}}(B,[i_s,i_t]) \ge \mu_{\text{th}}$ whenever $C_n \le \left(\frac{B^2 \phi}{144 (B+G)^3 \log n} \right)^3 n = \tilde O(n)$.
\end{restatable}
\begin{proof}
Suppose $\text{gap}_{\text{min}}(-B,[i_s,i_t]) < \mu_{\text{th}}$. Then the largest value of offline optimal attained within this bin $[i_s,i_t]$ is at-most $-B+ \mu_{\text{th}} + B/\sqrt{n_i}$ (recall $n_i := i_t - i_s + 1$ and TV within this bin is at-most $B/\sqrt{n_i}$ by Lemma \ref{lem:part-sq}). So $\text{gap}_{\text{min}}(B,[i_s,i_t]) \ge 2B - \mu_{\text{th}} - B/\sqrt{n_i}$. Our goal is to show that whenever $C_n$ obeys the constraint stated in the lemma, we must have
\begin{align}
    2B - \mu_{\text{th}} - B/\sqrt{n_i}
    &\ge \mu_{\text{th}}. \label{eq:gap}
\end{align}

Let $\ell_i$ be the length of a sub-interval of $[i_s,i_t]$ where the offline optimal solution assumes the form of Structure 1 or Structure 2. Due to Lemma \ref{lem:lamda-len}, we have
\begin{align}
    n_i \ge \ell_i \ge \frac{2 \lamda}{(G+B)} \ge \frac{2\phi}{(G+B)} \frac{n^{1/3}}{C_n^{1/3}}, \label{eq:len-lb}
\end{align}
where the last inequality follows due to the condition on $\lamda$ assumed in the current lemma. So a sufficient condition for Eq.\eqref{eq:gap} to be true is
\begin{align}
    2B \ge 2 \left(2 \sqrt{\frac{36(G+B)^3 C_n^{1/3} \log n}{ \phi n^{1/3}}}  \vee B\sqrt{\frac{(G+B)C_n^{1/3}}{2 \phi n^{1/3}}}\right).
\end{align}

Recall that by Assumption A1 in Section \ref{sec:square-game}, we have $G \ge B \ge 1$ WLOG. So the above maximum will be attained by the first term and can be further simplified as
\begin{align}
    2B \ge 4 \sqrt{\frac{36(G+B)^3 C_n^{1/3} \log n}{ \phi n^{1/3}}}.
\end{align}

The above condition is always satisfied whenever $C_n \le \left(\frac{B^2 \phi}{144 (B+G)^3 \log n} \right)^3 n$.

At this point, we have shown that $\text{gap}_{\text{min}}(-B,[i_s,i_t]) < \mu_{\text{th}} \implies \text{gap}_{\text{min}}(B,[i_s,i_t]) \ge \mu_{\text{th}}$ under the conditions of the lemma. Taking the contrapositive yields $\text{gap}_{\text{min}}(B,[i_s,i_t]) < \mu_{\text{th}} \implies \text{gap}_{\text{min}}(-B,[i_s,i_t]) \ge \mu_{\text{th}}$.
\end{proof}

\begin{figure}[h!]
\includegraphics[width=\linewidth,keepaspectratio=true]{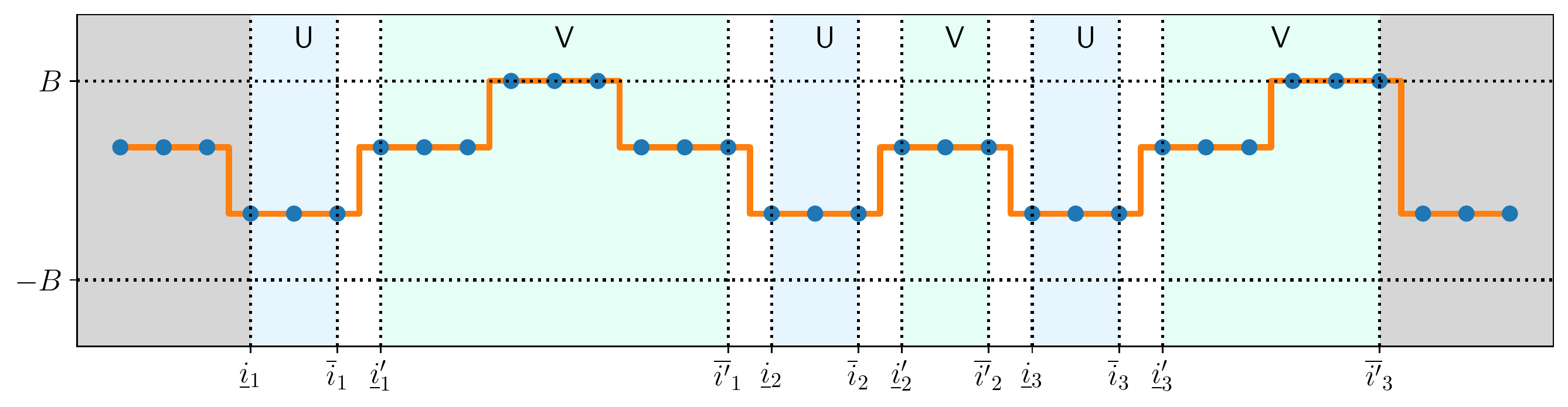}
    \caption{\emph{Refinement of a bin that satisfy Condition (A) in the proof of Lemma \ref{lem:high-lamda} with  $\text{gap}_{\text{min}}(-B,[i_s,i_t]) \ge \text{gap}_{\text{min}}(B,[i_s,i_t])$. Here we assign $i_s = \underbar i_1$ and $i_t = \bar{i}'_3$ The following pairs are formed in the proof of Lemma \ref{lem:high-lamda}: $\cP_i = ([\underbar i_1,\bar i_1],[\underbar i'_1,\bar i'_1]),([\underbar i_2,\bar i_2],[\underbar i'_2,\bar i'_2]),([\underbar i_3,\bar i_3],[\underbar i'_3,\bar i'_3])$. Blue dots represent the optimal sequence}}    
    \label{fig:splits-1d}
\end{figure}

\begin{restatable}{lemma}{lemHighLamda} (\textbf{high} $\bs \lamda$ \textbf{regime})\label{lem:high-lamda}
If the optimal dual variable $\lamda \ge \phi \frac{n^{1/3}}{C_n^{1/3}} = \Omega\left( \frac{n^{1/3}}{C_n^{1/3}} \right)$ for some constant $\phi > 0$ that doesn't depend on $n$ and $C_n$, we have the regret of FLH-OGD strategy bounded as
\begin{align}
    \sum_{t=1}^{n} (y_t - x_t)^2 - (y_t - u_t)^2 
    &= \tilde O(n^{1/3}C_n^{2/3} \vee 1),
\end{align}
where $x_t$ is the prediction of FLH-OGD at time $t$.
\end{restatable}
\begin{proof}
Throughout the proof, we consider only the regime where $C_n \le \left(\frac{B^2 \phi}{84 (B+G)^3 \log n} \right)^3 n = \tilde O(n)$. In the alternate regime where $C_n = \tilde \Omega(n)$, the trivial regret bound of $\tilde O(n)$ is near minimax optimal.

Reminiscent to the road-map in Section \ref{sec:road-sq}, it is useful to define the following condition:

\textbf{Condition (A):} Let a bin $[a,b]$ be given such that $C_{a \rightarrow b} \le B/\sqrt{b-a+1}$. It satisfies at-least one of the following criteria. (i)  $\text{gap}_{\text{min}}(B,[a,b]) \ge \text{gap}_{\text{min}}(-B,[a,b])$ and the optimal solution takes the form of Structure 1 in at-least one sub-interval $[r,s] \subseteq [a,b]$; or (ii) $\text{gap}_{\text{min}}(-B,[a,b]) \ge \text{gap}_{\text{min}}(B,[a,b])$ and the optimal solution takes the form of Structure 2 in at-least one sub-interval $[r,s] \subseteq [a,b]$.


Consider a bin $[i_s,i_t] \in \cP$ that satisfies Condition (A). We refine $[i_s,i_t]$ into a partition that contains smaller sub-intervals as follows:
\begin{align}
    \cP_i &:= \{ [i_s,\ubar i_1-1], [\ubar i_1, \bar i_1], [\ubar i'_1, \bar i'_1], \ldots,[\ubar i_{m^{(i)}}, \bar i_{m^{(i)}}], [\ubar i'_{m^{(i)}}, \bar i'_{m^{(i)}} := i_t]\}, \label{eq:part-refine}
\end{align}
such that:
\begin{enumerate}
    \item If $\text{gap}_{\text{min}}(B,[i_s,i_t]) > \text{gap}_{\text{min}}(-B,[i_s,i_t])$, then the offline optimal in the intervals $[\ubar i_j, \bar i_j]$, $j \in [m^{(i)}]$ takes the form of Structure 1. Further, let $k$ be the largest value in $[i_s,i_t]$ such that $u_{i_s:k}$ is constant. If $u_{i_s} > u_{i_s-1}$ and $u_{k} > u_{k+1}$, then we treat the first sub-interval in $\cP_i$ as empty by putting $\ubar i_1 = i_s$. Similarly let $k$ be smallest value in $[i_s,i_t]$ such that $u_{k:i_t}$ is constant. If $u_{k-1} < u_k$ and $u_{i_t} > u_{i_t+1}$ then we treat the last sub-interval in $\cP_i$ as empty by putting $\ubar i'_{m^{(i)}} = i_t+1$.

    \item If $\text{gap}_{\text{min}}(B,[i_s,i_t]) \le \text{gap}_{\text{min}}(-B,[i_s,i_t)$, then the offline optimal in the intervals $[\ubar i_j, \bar i_j]$, $j \in [m^{(i)}]$ takes the form of Structure 2. Further, let $k$ be the largest value in $[i_s,i_t]$ such that $u_{i_s:k}$ is constant. If $u_{i_s} < u_{i_s-1}$ and $u_{k} < u_{k+1}$, then we treat the first sub-interval in $\cP_i$ as empty by putting $\ubar i_1 = i_s$. Similarly let $k$ be smallest value in $[i_s,i_t]$ such that $u_{k:i_t}$ is constant. If $u_{k-1} > u_k$ and $u_{i_t} < u_{i_t+1}$ then we treat the last sub-interval in $\cP_i$ as empty by putting $\ubar i'_{m^{(i)}} = i_t+1$.

    \item In all sub-intervals $[\ubar i'_j, \bar i'_j]$, $j \in [m^{(i)}]$, the offline optimal sequence can be split into piece-wise monotonic sections with at-most 2 pieces.
\end{enumerate}

An illustration of this refinement scheme is given in Fig.\ref{fig:splits-1d}.

Let there be $m_1^{(i)}$ bins among $\{[\ubar i_1, \bar i_1], \ldots, [\ubar i_{m^{(i)}}, \bar i_{m^{(i)}}]\}$ which satisfy the property in Lemma \ref{lem:within1d}. Let their lengths be denoted by $\{\ell^{(1)}_{1^{(i)}}, \ldots, \{\ell^{(1)}_{m_1^{(i)}}\}$. These bins will be referred as \emph{Type 1} bins henceforth.

Similarly let there be $m_2^{(i)}$ bins among $\{ [\ubar i_1, \bar i_1], \ldots, [\ubar i_{m^{(i)}}, \bar i_{m^{(i)}}]\}$ which satisfy either Case 1 or Case 2 in Lemma \ref{lem:outside1d}. Let their lengths be denoted by $\{\ell^{(2)}_{1^{(i)}}, \ldots, \{\ell^{(2)}_{m_2^{(i)}}\}$. These bins will be referred as \emph{Type 2} bins henceforth.

Each bin in Type 1 and Type 2 can be paired with one adjacent bin (if non-empty) in $\cP_i$ where the optimal sequence displays a piece-wise monotonic behaviour with at-most 2 pieces. (For example the bin  $[\ubar i_1, \bar i_1]$ can be paired with $[\ubar i'_1, \bar i'_1]$ where in the later the optimal sequence displays a piece-wise monotonic behaviour. See Fig.\ref{fig:splits-1d} for example.) To see why this is true, consider the case  $\text{gap}_{\text{min}}(B,[i_s,i_t]) \le \text{gap}_{\text{min}}(-B,[i_s,i_t)$. By construction, the optimal solution must preclude the form of Structure 2 in the bin $[\ubar i'_k, \bar i'_k]$ where $k \in [m^{(i)}]$. This means the offline optimal can either take a non-increasing form in $[\ubar i'_k, \bar i'_k]$ or it can monotonically increase and then optionally monotonically decrease. In both the cases, it can be split into at-most 2 sections where the solution is purely monotonic. Similar arguments apply for the case $\text{gap}_{\text{min}}(B,[i_s,i_t]) > \text{gap}_{\text{min}}(-B,[i_s,i_t)$.

Similarly, if bin $[i_s,\ubar i_1-1]$ is non-empty then the offline optimal must assume a piece-wise monotonic structure with at-most 2 pieces. Then applying Lemma \ref{lem:mono-1d} to each of the 2 pieces separately and adding the regret bounds yields
\begin{align}
    \sum_{j=i_s}^{\ubar i_1 - 1} f_j(x_j) - f_j(u_j) = \tilde O(1). \label{eq:first-bin}
\end{align}

Note that $m_1^{(i)}+m_2^{(i)} = m^{(i)}$. Let the total regret contribution from Type 1 bins along with their pairs and Type 2 bins along with their pairs be referred as $R_1^{(i)}$ and $R_2^{(i)}$ respectively.

Since a sub-bin that is paired with a Type 1 or Type 2 bin can be split into at-most 2 sub-intervals where the optimal sequence is purely monotonic (see Fig. \ref{fig:splits-1d}), we can bound the regret within such sub-bins $[\ubar i'_k, \bar i'_k]$, $k \in [m^{(i)}]$ by at-most $62(G+B)^2 \log n$ by Lemma \ref{lem:mono-1d}. 

For a Type 2 bin $[a,b] \subseteq [i_s,i_t]$, we can have two possible configurations: If $\text{gap}_{\text{min}}(B,[i_s,i_t]) > \text{gap}_{\text{min}}(-B,[i_s,i_t])$ then $B-u_a \ge \text{gap}_{\text{min}}(B,[i_s,i_t]) \ge \mu_{\text{th}}$ where the first inequality follows by the definition of $\text{gap}_{\text{min}}(B,[i_s,i_t])$ and the last inequality follows by Lemma \ref{lem:large-margin}. Similarly If $\text{gap}_{\text{min}}(B,[i_s,i_t]) \le \text{gap}_{\text{min}}(-B,[i_s,i_t])$ then $B+u_a \ge \text{gap}_{\text{min}}(-B,[i_s,i_t]) \ge \mu_{\text{th}}$. With this observation and using the results of Lemma \ref{lem:outside1d}, we can bound the regret contribution from any Type 2 bin and its pair as:

\begin{align}
   R_2^{(i)} 
   &\le \sum_{j=1^{(i)}}^{m_2^{(i)}} \left( \left( 10 (G+B)^2 \log n - \ell^{(2)}_j \mu_{\text{th}}^2 \right) + 62(G+B)^2 \log n \right)\\
   &\le 72 m_2^{(i)} (G+B)^2 \log n - \mu_{\text{th}}^2 (\sum_{j=1^{(i)}}^{m_2^{(i)}} \ell^{(2)}_j ).
\end{align}
From Eq.\eqref{eq:len-lb}, we have $\ell^{(2)}_j \ge \frac{2\phi n^{1/3}}{(G+B)C_n^{1/3}}$ for $j \in \{1^{(i)},\ldots,m_2^{(i)}\}$. So we can continue as

\begin{align}
    R_2^{(i)}
    &\le 72 m_2^{(i)} (G+B)^2  \log n - \mu_{\text{th}}^2 \frac{2\phi n^{1/3}}{(G+B)C_n^{1/3}} m_2^{(i)}\\
    &= 0,
\end{align}
where the last line is obtained by plugging in the value of $\mu_{\text{th}}$ from Lemma \ref{lem:large-margin}.

So by refining every interval in $\cP$ that satisfy Condition (A) and summing the regret contribution from all Type 2 bins and their pairs across all refined intervals in $\cP$ yields

\begin{align}
    \sum_{i=1}^M R_2^{(i)}
    &\le 0, \label{eq:type2-sum}
\end{align}
where we recall that $M := |\cP| = O(n^{1/3}C_n^{2/3} \vee 1)$ and assign $R_2(i) = 0$ for intervals in $\cP$ that do not satisfy Condition (A).

For any Type 1 bin, its regret contribution can be bounded by Lemma \ref{lem:within1d}. So we have the regret contribution from Type 1 bins and their pairs bounded as

\begin{align}
    R_1^{(i)}
    &\le \sum_{j=1}^{m_1^{(i)}} \left( \left( 10 (G+B)^2 \log n - \frac{4\lamda^2}{\ell^{(1)}_{j^{(i)}}} \right) + 62(G+B)^2 \log n \right)\\
    &= 72 m_1^{(i)} (G+B)^2 \log n - 4 \lamda^2 \sum_{j=1}^{m_1^{(i)}} \frac{1}{\ell^{(1)}_{j^{(i)}}}\label{eq:type2reg}
\end{align}

By refining every interval in $\cP$ that satisfies Condition (A) and summing the regret contribution from all Type 2 bins and their pairs across all refined intervals in $\cP$ yields
\begin{align}
    \sum_{i=1}^M R_1^{(i)}
    &\le 72 (G+B)^2 \log n   \sum_{i=1}^M m_1^{(i)} - 4 \lamda^2  \sum_{i=1}^M \sum_{j=1}^{m_1^{(i)}} \frac{1}{\ell^{(1)}_{j^{(i)}}}\\
    &\le 72 (G+B)^2 M_1 \log n  - 4 \lamda^2 \frac{M_1^2}{n}, \label{eq:type1}
\end{align}
where in the last line:  a) we define $M_1 := \sum_{i=1}^M m_1^{(i)}$ with the convention that $m_1^{(i)} = 0$ if the bin $[i_s,i_t] \in \cP$ doesn't satisfy Condition (A); b) applied AM-HM inequality and noted that $\sum_{i=1}^M \sum_{j=1}^{m_1^{(i)}} \ell^{(1)}_{j^{(i)}} \le n$.

To further bound Eq.\eqref{eq:type1}, we consider two separate regimes as follows. 

Recall that $\lamda \ge \phi \frac{n^{1/3}}{C_n^{1/3}}$. So continuing from Eq.\eqref{eq:type1},
\begin{align}
    \sum_{i=1}^M R_1^{(i)}
    &\le 72 (G+B)^2 M_1 \log n  - 4 \phi^2 \frac{n^{2/3}}{C_n^{2/3}}  \frac{M_1^2}{n}\\
    &\le 0, \label{eq:type1-sum1}
\end{align}
whenever $M_1 \ge \frac{18(G+B)^2 \log n}{ \phi^2} n^{1/3}C_n^{2/3} = \tilde \Omega(n^{1/3}C_n^{2/3})$.

In the alternate regime where $M_1 \le \frac{18(G+B)^2 \log n}{ \phi^2} n^{1/3}C_n^{2/3} = \tilde O(n^{1/3}C_n^{2/3} \vee 1)$, we trivially obtain:
\begin{align}
 \sum_{i=1}^M R_1^{(i)} = \tilde O(n^{1/3}C_n^{2/3} \vee 1) \label{eq:type1-sum2} 
\end{align}

Putting everything together by combining the bounds in Eq.\eqref{eq:first-bin}, \eqref{eq:type2-sum}, \eqref{eq:type1-sum1} and \eqref{eq:type1-sum2}, we can bound the total regret contribution from the bins that satisfy Condition (A) as:
\begin{align}
    \sum_{i=1}^M R_1^{(i)} + R_2^{(i)} + \tilde O(1)
    &= \tilde O(n^{1/3}C_n^{2/3} \vee 1),
\end{align}
where we have assigned $R_1^{(i)} = R_2^{(i)} = 0$ for bins that don't satisfy Condition (A).

Throughout the proof till now, we have only considered bins $[i_s,i_t] \in \cP$ which satisfy Condition (A). Not meeting this criterion will only make the arguments easier as explained below.

If a bin $[i_s,i_t] \in \cP$  doesn't satisfy Condition (A), by taking a logical negation of Condition (A), we conclude that this can only happen if the optimal solution precludes the form of either Structure 1 or Structure 2 (or both) within some sub-interval of $[i_s,i_t]$. Consequently by applying similar arguments we used to handle the bins $[\ubar i'_k, \bar i'_k]$, $k \in [m^{(i)}]$, we can split the offline optimal sequence $u_{i_s:i_t}$ into at-most 2 piece-wise monotonic sections and use Lemma \ref{lem:mono-1d} to bound the regret in $[i_s,i_t]$ as $\tilde O(1)$. Since $|\cP| = O(n^{1/3}C_n^{2/3} \vee 1)$, we conclude that the total regret from all bins that don't satisfy Condition (A) is $\tilde O(n^{1/3}C_n^{2/3} \vee 1)$.
\end{proof}

\begin{proof} \textbf{of Theorem \ref{thm:main-sq}}.
The proof is now immediate from Lemmas \ref{lem:low-lamda} and \ref{lem:high-lamda}.
\end{proof}

\section{Proofs for Section \ref{sec:sc}} \label{app:sc}

\corSc*
\begin{proof}

Due to strong convexity, we have for any $\bs w_t \in \mathbb{R}^d$,

\begin{align}
    f_t(\bs x_t) - f_t(\bs w_t)
    &\le -\langle \grad f_t(\bs x_t), \bs w_t - \bs x_t \rangle - \frac{H}{2} \| \bs w_t - \bs x_t\|^2\\
    &=H \left( \langle \grad f_t(\bs x_t)/H, \bs x_t - \bs x_t \rangle + (1/2) \| \bs x_t - \bs x_t \|^2) \right) \\
    &\quad- H \left( \langle \grad f_t(\bs x_t)/H, \bs w_t - \bs x_t \rangle + (1/2) \| \bs w_t - \bs x_t \|^2)\right)\\
    &= \sum_{i=1}^d  H \left(\grad f_t(\bs x_t)[i] (\bs x_t[i] - \bs x_t[i])/H + (1/2) (\bs x_t[i] - \bs x_t[i])^2 \right)\\
    &\quad - H\left( \grad f_t(\bs x_t)[i] (\bs w_t[i] - \bs x_t[i])/H + (1/2) (\bs w_t[i] - \bs x_t[i])^2 \right)\\
    &= (H/2) \left( \sum_{i=1}^d \ell^{(i)}_t(\bs x_t[i]) - \ell^{(i)}_t(\bs w_t[i]), \right) \label{eq:sc}    
\end{align}
where the last line is obtained by completing the squares. Let $\bs u_t \in \mathbb{R}^d$ for $t \in [n]$ be defined as the offline optimal sequence corresponding to the optimization problem:
\begin{mini!}|s|[2]                   
    {\tilde {\bs u}_1,\ldots,\tilde {\bs u}_n,
    \tilde{z_1},\ldots,\tilde {\bs z}_{n-1}}                               
    {\sum_{t=1}^{n} \sum_{i=1}^d \ell_t^{(i)}(\tilde {\bs u}_t[i])}   
    {\label{eq:Example1}}             
    {}                                
    \addConstraint{\tilde {\bs z}_t}{=\tilde {\bs u}_{t+1} - \tilde {\bs u}_{t} \: \forall t \in [n-1],}    
    \addConstraint{\sum_{t=1}^{n-1} \|\tilde {\bs z}_t\|_1}{\le C_n, \label{eq:constr-1d}}  
\addConstraint{ \|\tilde{ \bs u}_t\|_\infty}{\le B \: \forall t \in [n],\label{eq:constr-3d}}
\end{mini!}

Let $C_n[i] = \sum_{t=2}^{n} |\bs u_t [i] - \bs u_{t-1}[i]|$ be its TV allocated to coordinate $i$. By Theorem 1, the FLH-OGD instance $i$ with learning rate $\zeta = 1/(2(2B+G_\infty/H)^2)$ attains the regret of $\tilde O\left(n^{1/3} (C_n[i])^{2/3} \vee 1\right)$ regret.  WLOG, let's assume that FLH-OGD instances for coordinates $i \in [k]$, $k \le d$ incurs $\tilde O\left(n^{1/3} (C_n[i])^{2/3} \right)$ regret wrt losses $\ell_t^{(i)}$  and the regret incurred by FLH-OGD instances for coordinates $k > k'$ is $O(\log n)$. Let $R_n(\bs w_{1:n}) := \sum_{t=1}^n f_t(\bs x_t) - f_t(\bs w_t)$ and $R'_n(\bs w_{1:n}) := (H/2) \left( \sum_{t=1}^n \sum_{i=1}^d \ell^{(i)}_t(\bs x_t[i]) - \ell^{(i)}_t(\bs w_t[i]) \right)$. From Eq.\eqref{eq:sc} $R_n(\bs w_{1:n}) \le R_n'(\bs w_{1:n})$. We have,
\begin{align}
    R_n(\bs w_{1:n})
    &\le R'_n(\bs w_{1:n})\\
    &\le \sup_{\substack{\bs w_1,\ldots,\bs w_n \in \cD \\ \sum_{t=2}^n \|\bs w_t - \bs w_{t-1} \|_1 \le C_n}} R'_n(\bs w_{1:n})\\
    &=  R'_n(\bs u_{1:n})\\
    &= (d-k) \tilde O(1) + \sum_{i=1}^{k} \tilde O\left(n^{1/3} (C_n[i])^{2/3} \right)\\
    &\le (d-k) \tilde O(1) + \tilde O\left(n^{1/3}  (k)^{1/3} \left( \sum_{i=1}^{k} C_n[i] \right) ^{2/3} \right), \label{eq:fin-sc}
\end{align}
where the last line follows by Holder's inequality $\bs x^T \bs y \le \| \bs x\|_3 \| \bs y\|_{3/2}$, where we treat $\bs x$ as just a vector of ones in $\mathbb{R}^{k}$. The above expression can be further upper bounded by\\ $\tilde O\left (2d \vee 2d^{1/3}n^{1/3} C_n^{2/3}\right)$.
\end{proof}

\section{Proofs for Section \ref{sec:ec}} \label{app:ec}

We start by inspecting the KKT conditions.
\begin{lemma} \label{lem:kkt-ec-d} (\textbf{characterization of offline optimal}) 
Consider the following convex optimization problem (where $\tilde{\bs z}_1,...,\tilde{\bs z}_{n-1}$ are introduced as dummy variables).
\begin{mini!}|s|[2]                   
    {\tilde{ \bs u}_1,\ldots,\tilde{ \bs u}_n,\tilde{ \bs z}_1,\ldots,\tilde{ \bs z}_{n-1}}                               
    {\sum_{t=1}^n f_t(\tilde {\bs u}_t)}   
    {\label{eq:Example1}}             
    {}                                
    \addConstraint{\tilde{ \bs z}_t}{=\tilde{ \bs u}_{t+1} - \tilde{ \bs u}_{t} \: \forall t \in [n-1],}    
    \addConstraint{\sum_{t=1}^{n-1} \|\tilde{ \bs z}_t\|_1}{\le C_n, \label{eq:constr-ec-1d}}  
    \addConstraint{\| \tilde {\bs u}_t\|_\infty}{\le B \: \forall t \in [n],\label{eq:constr-ec-2d}}
\end{mini!}

Let $\bs u_1,\ldots,\bs u_n,\bs z_1,\ldots,\bs z_{n-1} \in \mathbb{R}^d$ be the optimal primal variables and let $\lambda \ge 0$ be the optimal dual variable corresponding to the constraint \eqref{eq:constr-ec-1d}. Further, let $\bs \gamma_t^+, \bs \gamma_t^- \in \mathbb{R}^d$ with $\bs \gamma_t^+ \ge \bs 0$ and $\bs \gamma_t^- \ge \bs 0$ be the optimal dual variables that correspond to constraint \eqref{eq:constr-ec-2d}. Specifically for $k \in [d]$, $\bs \gamma_t^+[k]$ corresponds to the dual variable for the constraint $\bs u_t[k] \le B$ induced by the relation \eqref{eq:constr-ec-2d}. Similarly $\bs \gamma_t^-[k]$ corresponds to the constraint $-B \le \bs u_t[k]$. By the KKT conditions, we have

\begin{itemize}
    \item \textbf{stationarity: } $\grad f_t({\bs u}_t) = \lambda \left ( \bs s_t - \bs s_{t-1} \right) +  \bs \gamma^-_t -  \bs \gamma^+_t$, where $\bs s_t \in \partial|\bs z_t|$ (a subgradient). Specifically, $\bs s_t[k]=\sign(\bs u_{t+1}[k]-\bs u_t[k])$ if $|\bs u_{t+1}[k]-\bs u_t[k]|>0$ and $\bs s_t[k]$ is some value in $[-1,1]$ otherwise. For convenience of notations later, we also define 
    $\bs s_n = \bs s_0 = \bs 0$.
    \item \textbf{complementary slackness: } (a) $\lamda \left(\sum_{t=2}^n \|\bs u_t - \bs u_{t-1}\|_1 - C_n \right) = 0$; (b)  $ \bs \gamma^-_t[k] ( \bs u_t[k] + B) = 0$ and $ \bs \gamma^+_t[k] ( \bs u_t[k] - B) = 0$ for all $t \in [n]$ and all $k \in [d]$.
\end{itemize}
\end{lemma}
The proof of the above lemma is similar to that of Lemma \ref{lem:kkt-sq} and hence omitted.

\noindent\textbf{Terminology.} We will refer to the optimal primal variables $\bs u_1,\ldots,\bs u_n$ in Lemma \ref{lem:kkt-ec-d} as the \emph{offline optimal sequence} in this section. We reserve the term \emph{FLH-ONS} for the instantiation of FLH with ONS as base learners with parameters as in Theorem \ref{thm:ec-d}.

\textbf{Notations.} For bin $[i_s,i_t] \in \cP$ we define: $n_i = i_t - i_s + 1$, $\bar {\bs u}_i = \frac{1}{n_i}\sum_{j=i_s}^{i_t} \bs u_j$, $\bs \Gamma_i^+ = \sum_{j=i_s}^{i_t} \bs \gamma_j^+$, $\bs \Gamma_i^- = \sum_{j=i_s}^{i_t} \bs \gamma_j^-$, $\Delta \bs s_i = \bs s_{i_t} - \bs s_{i_s-1}$, $C_i = \sum_{j=i_s+1}^{i_t} \|\bs u_j - \bs u_{j-1}\|_1$.

For any general bin $[a,b]$ define the quantities $n_{a \rightarrow b},\bar {\bs u}_{a \rightarrow b}, \bs \Gamma^+_{a \rightarrow b}, \bs \Gamma^-_{a \rightarrow b},\Delta \bs s_{a \rightarrow b},C_{a \rightarrow b}$ analogously as above.

The following is a direct extension for Lemma \ref{lem:part-sq}.
\begin{restatable}{lemma}{lempartec}\label{lem:key-part-multi}(\textbf{key partition}) 
Initialize $\cP \leftarrow \Phi$. Starting from time 1, spawn a new bin $[i_s,i_t]$ whenever $\sum_{j=i_s+1}^{i_t+1} \| \bs u_j -  \bs u_{j-1} \|_1 > B/\sqrt{n_i}$, where $n_i = i_t - i_s + 2$. Add the spawned bin $[i_s,i_t]$ to $\cP$.

Let $M:=|\cP|$. We have $M = O\left (1 \vee n^{1/3}C_n^{2/3} B^{-2/3} \right)$.
\end{restatable}

\begin{proposition} \label{prop:loss}
The losses $f_t$ defined in Eq.\eqref{eq:exp-surrogate} are:
\begin{itemize}
    \item $G^2$ gradient Lipschitz over the domain $\cD$ in Assumption B1
    
    \item Define $\gamma := 2GB\sqrt{\alpha d/2} + 1/\sqrt{2 \alpha}$. Then the losses $f_t$ are $\alpha' := 1/(2 \gamma^2)$ exp-concave across $\cD$.
    
    \item $f_t$ are $G' := 2\alpha G^2 B\sqrt{d} + G$ Lipschitz in L2 norm across $\cD$.
\end{itemize}
\end{proposition}
\begin{proof}
The first two statements have been already proved in Section \ref{sec:ec}. For the last statement we have that
\begin{align}
    \grad f_t(\bs x)
    &= (\alpha \grad \ell_t(\bs x_t)^T (\bs x - \bs x_t) + 1) \grad \ell_t(\bs x_t).
\end{align}

So by triangle inequality we obtain that $\| \grad f_t(\bs x) \|_2 \le 2\alpha G^2 B\sqrt{d} + G$.
\end{proof}

\begin{lemma} (\textbf{Strongly Adaptive regret}) (\citep{hazan2007logregret}, \citep{hazan2007adaptive}) \label{lem:stat-flh}
Consider any bin $[a,b]$ and a comparator $\bs w \in \cD$. Under Assumptions B1-2 in Section \ref{sec:ec}, the static regret of the FLH-ONS with losses $f_t$ obeys
\begin{align}
    \sum_{j=a}^b f_j(\bs x_j) - f_j(\bs w) 
    &\le 10d(8G^2B^2\alpha d + 1/\alpha) \log n
\end{align}
where $\bs x_j$ are predictions of FLH-ONS and $\gamma$ is as defined in Theorem \ref{thm:ec-d}.
\end{lemma}
\begin{proof}
Let $\alpha' = 1/(2 \gamma^2)$. The static regret of ONS is $5d(G'D + 1/\alpha') \log n$ for $\alpha'$ exp-concave losses (Theorem 2 in \citep{hazan2007logregret}) where $D$ is the diameter of the decision set. We have $D = 2B\sqrt{d}$ for the box decision set. the static regret of ONS in our setting is at-most $5d(2G'B\sqrt{d} + 1/\alpha')\log n$.

The regret of the FLH against any of its base experts is at-most $(4/\alpha') \log n$ for $\alpha'$ exp-concave losses (Theorem 3.2 in \citep{hazan2007adaptive}). Adding both these regret bounds, using Proposition \ref{prop:loss} and further upper bounding the sum results in the lemma.

\end{proof}

\begin{lemma}(\textbf{low} $\bs \lamda$ \textbf{regime})\label{lem:low-lamda-multi}
If the optimal dual variable $\lamda = O\left( \frac{d^{1.5} n^{1/3}}{C_n^{1/3}} \right)$, we have the regret of FLH-ONS strategy bounded as
\begin{align}
    \sum_{t=1}^{n} f_t(\bs x_t) - f_t(\bs u_t)
    &= \tilde O \left ( 10d(8G^2B^2\alpha d + 1/\alpha) (n^{1/3}C^{2/3} \vee 1) \right ),
\end{align}
where $\bs x_t$ is the prediction of FLH-ONS at time $t$.
\end{lemma}
\begin{proof}
Consider a bin $[i_s,i_t] \in \cP$. Note that for any $j \in [i_s,i_t]$ and $k \in [d]$, both $\bs \gamma_j^+[k]$ and $\bs \gamma_j^-[k]$ can't be simultaneously non-zero due to complementary slackness and the fact that $C_i \le B/\sqrt{n_i} < 2B$ by the construction in Lemma \ref{lem:key-part-multi}. For some fixed $\check{\bs u} \in \cD$, we have

\begin{align}
    \underbrace{\sum_{j=i_s}^{i_t} f_j(\bs x_j) - f_j(\check{\bs u})}_{T_{1,i}} + \underbrace{\sum_{j=i_s}^{i_t} f_j(\check{\bs u}) - f_j(\bs u_j)}_{T_{2,i}}.
\end{align}

By virtue of Lemma \ref{lem:stat-flh}, we have $T_{1,i} = \tilde O(d^{2})$. Due to gradient Lipschitzness in Proposition \ref{prop:loss}

\begin{align}
    T_{2,i}
    &\le \sum_{j=i_s}^{i_t} \langle \grad f_j(\bs u_j), \check{\bs u} - \bs u_j \rangle + \frac{G^2}{2} \| \check{\bs u} - \bs u_j \|_2^2. \label{eq:strong-smooth}
\end{align}

We construct $\check{\bs u}$ as follows: 
\begin{itemize}
    \item If there exists a $j \in [i_s,i_t]$ and $k \in [d]$ such that $\bs u_j[k] = B$, set $\check{\bs u}[k] = B$.

    \item If there exists a $j \in [i_s,i_t]$ and $k \in [d]$ such that $\bs u_j[k] = -B$, set $\check{\bs u}[k] = -B$.
    
    \item If the optimal solution doesn't touch either boundaries $\pm B$ in $[i_s,i_t]$ across a coordinate, set $\check{\bs u}[k] = \bs u_{i_s}[k]$.
\end{itemize}

It is easy to see that $\check{\bs u} \in \cD$ and $\|\check{\bs u} - \bs u_j \|_2 \le \|\check{\bs u} - \bs u_j \|_1 \le C_i$ for all $j \in [i_s,i_t]$. Using this observation along with the KKT conditions, we continue from  Eq.\eqref{eq:strong-smooth} as
\begin{align}
    T_{2,i}
    &\le G^2 n_i C_i^2 + \sum_{j=i_s}^{i_t} \langle \grad f_j(\bs u_j), \check{\bs u} - \bs u_j \rangle,\\
    &\le G^2 n_i C_i^2 + \sum_{j=i_s}^{i_t} \lamda \langle \bs s_j - \bs s_{j-1}, \check{\bs u} - \bs u_j \rangle + \langle \bs \gamma_j^- - \bs \gamma_j^+, \check{\bs u} - \bs u_j \rangle\\
    &\le_{(a)} G^2 B^2 + \lamda \langle \bs s_{i_s-1}, \bs u_{i_s} - \check{\bs u} \rangle - \lamda \langle \bs s_{i_t}, \bs u_{i_t} - \check{\bs u} \rangle + \lamda \sum_{j=i_s+1}^{i_t} \| \bs u_j - \bs u_{j-1}\|_1\\
    &\quad + \sum_{j=i_s}^{i_t} \sum_{k=1}^d \bs \gamma_j^-[k] (\check{\bs u}[k] - \bs u_j[k]) - \gamma_j^+[k] (\check{\bs u}[k] - \bs u_j[k])\\
    &\le_{(b)} G^2 B^2 + 3 \lamda C_i,
\end{align}
where line (a) is obtained by using that fact that $C_i \le B/\sqrt{n_i}$ and a rearrangement of the summations and line (b) is obtained by noting that $\bs \gamma_j^-[k] = 0$ when $\bs u_j[k] > -B$ via complementary slackness and $\check{\bs u}[k] - \bs u_j[k]$ is zero when $\bs u_j[k] = -B$ since by construction of $\check{\bs u}$: $\check{\bs u}[k] = -B$ if $\bs u_j[k] = -B$ for some $j \in [i_s,i_t]$. Similar arguments are applied to show the terms including $\bs \gamma_j^+$ also sums to zero. In line (b) we also used the fact that $\langle \bs s_{i_s-1}, \bs u_{i_s} - \check{\bs u} \rangle \le \| \bs s_{i_s-1}\|_\infty \| \bs u_{i_s} - \check{\bs u} \|_1 \le C_i$. Similarly $\langle \bs s_{i_t}, \bs u_{i_t} - \check{\bs u} \rangle \le C_i$

Hence summing $T_{1,i}$ and $T_{2,i}$ across all bins in $\cP$ yields
\begin{align}
    \sum_{i=1}^{M} T_{1,i} + T_{2,i}
    &\le_{(a)} \tilde O(M d^{2}) + \lamda C_n\\
    &\le \tilde O \left ( 10d(8G^2B^2\alpha d + 1/\alpha) (n^{1/3}C^{2/3} \vee 1) \right ),
\end{align}
where we recall that $M := |\cP| = O(1 \vee n^{1/3}C_n^{2/3})$ by Lemma \ref{lem:key-part-multi} and in line (a) we used $\sum_{i=1}^M C_i \le C_n$ and $\lamda = O(d^{1.5}n^{1/3}/C_n^{1/3})$ by the premise of the current Lemma.

\end{proof}

\begin{definition} 
For a bin $[a,b]$, the offline optimal is said to be \textbf{piece-wise maximally monotonic in} $\mathbf{[a,b]}$ \textbf{with} $\bs m$ \textbf{pieces} \textbf{across some coordinate} $\mathbf{k \in [d]}$, if we can split $[a,b]$ into $m$ disjoint consecutive bins $[a_1,b_1],\ldots,[a_m,b_m]$ such that the offline optimal sequence within each $[a_i,b_i]$  is purely monotonic across coordinate $k'$. Further, right-extending any interval $[a_i,b_i]$ to $[a_i,b_i+1]$ if $b_i+1 \in [a,b]$ makes $\bs u_{a_i:b_i+1}[k']$ non-monotonic. The sections $[a_i,b_i]$ for $i \in [m]$ are termed \textbf{maximally monotonic sections}.

\end{definition}

\begin{figure}[h!]
	\centering
	\fbox{
		\begin{minipage}{14 cm}
		    \code{generateGhostSequence}: Inputs- (1) offline optimal sequence (2) two numbers $k_{\text{fix}} \in [d] \cup \{ 0 \}$ and $u_{\text{fix}} \in [-B,B]$ (3) an interval $[a,b] \subseteq [n]$ where the offline optimal is piece-wise maximally monotonic with at-most 4 pieces across any coordinate $k \in [d]$.
            \begin{enumerate}

                \item Initialize $\cQ \leftarrow \Phi$.
                
                \item For each coordinate $k \in [d]$:
                \begin{enumerate}
                
                    \item If $k$ is same as $k_{\text{fix}}$, then set $\check{\bs u}_t[k] = u_{\text{fix}}$ for all $t \in [a,b]$. Goto Step 2.
                    
                    \item If the optimal solution is constant across coordinate $k$, set $\check{\bs u}_t[k] = \bs u_a[k]$ for all $t \in [a,b]$. Goto Step 2
                    
                    \item If the optimal solution monotonically increases (decreases) first across coordinate $k$, then:
                    
                    \begin{enumerate}
                        \item Split $[a,b]$ into at-most 7 sub-bins -- $[\ubar r_i, \bar r_i]$, $i \in [7]$ -- with the following properties: 
                        
                        \begin{itemize}
                            \item $\ubar r_1 = a$. $\ubar r_2$ is the largest value in $[a,b]$ such that $\bs u_{\ubar r_1 : \ubar r_2}$ is monotonically increasing (decreasing) and $\bs u_{\ubar r_2}[k] \underset{(<)}{>} \bs u_{\ubar r_2-1}[k]$. Set $\bar r_1 = \ubar r_2 - 1$.
                            
                            \item $\bar r_2$ is the largest value in $[a,b]$ such that $\bs u_{\ubar r_2 : \bar r_2}[k]$ is constant.
                            
                            \item If $\bar r_2 = b$, then set $[\ubar r_i, \bar r_i]$, $i \in [3,7]$ to be empty. Goto Step 2(c)(ii).
                            
                            \item $\ubar r_3 = \bar r_2 + 1$.
                            
                            \item If $\bs u_{\ubar r_3:b}[k]$ is a constant, set $\bar r_3 = b$. Set $[\ubar r_i, \bar r_i]$, $i \in [4,7]$ to be empty. Goto Step 2(c)(ii).
                            
                            \item $\ubar r_4$ is the largest point in $[\ubar r_3,b]$ such that $\bs u_{\ubar r_3:\ubar r_4}[k]$ is monotonically decreasing (increasing) and $u_{\ubar r_4}[k] \underset{(>)}{<} u_{\ubar r_4-1}[k]$. Set $\bar r_3 = \ubar r_4 - 1$.
                            
                            \item $\bar r_4$ is the largest point in $[\ubar r_4,b]$ such that $\bs u_{\ubar r_4 : \bar r_4}[k]$ is constant.
                            
                            \item If $\bar r_4 = b$, Set $[\ubar r_i, \bar r_i]$, $i \in [5,7]$ to be empty. Goto Step 2(c)(ii).
                            
                            \item $\ubar r_5 = \bar r_4+1$. 
                            
                            \item If $\bs u_{\ubar r_5:b}[k]$ is constant, then set $\bar r_5 = b$ and $[\ubar r_i, \bar r_i]$, $i \in [6,7]$ to be empty. Goto Step 2(c)(ii).
                            
                            \item $\ubar r_6$ is the largest point such that $\bs u_{\ubar r_5:\ubar r_6}[k]$ is monotonically increasing (decreasing) and $\bs u_{\ubar r_6}[k] \underset{(<)}{>} \bs u_{\ubar r_6-1}[k]$. Set $\bar r_5 = \ubar r_6-1$.
                            
                            \item $\bar r_6$ is the largest point in $[\ubar r_6,b]$ such that $\bs u_{\ubar r_6:\bar r_6}$ is constant. 
                            
                            \item If $\bar r_6 = b$, set $[\ubar r_7, \bar r_7]$ as empty. Goto Step 2(c)(ii).
                            
                            \item Set $\ubar r_7 = \bar r_6+1$ and $\bar r_7 = b$.
                        \end{itemize}
                        
                        \item Assign $\check{\bs u}_t[k] = \bs u_{\ubar r_1}[k]$ for all $t \in [\ubar r_1, \bar r_1]$; $\check{\bs u}_t[k] = \bs u_{\ubar r_2}[k]$ for all $t \in [\ubar r_2, \bar r_2]$ if non-empty; $\check{\bs u}_t[k] = \bs u_{\ubar r_3}[k]$ for all $t \in [\ubar r_3, \bar r_3]$ if non-empty; $\check{\bs u}_t[k] = \bs u_{\ubar r_4}[k]$ for all $t \in [\ubar r_4, \bar r_4]$ if non-empty; $\check{\bs u}_t[k] = \bs u_{\ubar r_5}[k]$ for all $t \in [\ubar r_5, \bar r_5]$ if non-empty; $\check{\bs u}_t[k] = \bs u_{\ubar r_6}[k]$ for all $t \in [\ubar r_6, \bar r_6]$ if non-empty; $\check{\bs u}_t[k] = \bs u_{\bar r_7}[k]$ for all $t \in [\ubar r_7, \bar r_7]$ if non-empty;

                    \end{enumerate}
                \end{enumerate}
                \item Return $\{\check{\bs u}_a,\ldots, \check{\bs u}_b\}$.
            \end{enumerate}
		\end{minipage}
	}
	\caption{\emph{\code{generateGhostSequence} procedure. If line 2(c) is replaced by ``If the optimal solution monotonically decreases first across coordinate $k$, then'', then we propagate that change by replacing the phrases increasing/decreasing and $>/<$ in the lines below 2(c)(i) by the bracketed statements next to it.}}
	\label{fig:ghost}
\end{figure}

\begin{lemma} \label{lem:ghost}
The sequence returned at Step 3 of \code{generateGhostSequence} in Fig.\ref{fig:ghost} has the following properties:
\begin{enumerate}
    \item[Property 1] The elements in the sequence changes only at-most $3d$ times. i.e, $\sum_{j=a+1}^b \mathbb{I}(\check{\bs u}_j \neq \check{\bs u}_{j-1}) \le 7d$, where $\mathbb{I}(\cdot)$ is the indicator function.
    \item[Property 2] Every member of the sequence lie in the box decision set $\cD$.
    
    \item[Property 3]  For any $j \in [a,b]$, $\sum\limits_{\substack{k=1 \\ k \neq k_{\text{fix}}}}^d |\check{\bs u}_j[k] - \bs u_j[k]| \le C_{a \rightarrow b}$, where $C_{a \rightarrow b}$ is the TV of the offline optimal in bin $[a,b]$.
\end{enumerate}
\end{lemma}
\begin{proof}
Observe that in the procedure detailed in Fig.\ref{fig:ghost}, we split the bin $[a,b]$ into at-most 7 bins across any coordinate. The value of the comparator across that coordinate stays unchanged in each of the new sub-bins. This implies that number of distinct comparators in $\{\check{\bs u}_a, \ldots, \check{\bs u}_b \}$ is at-most $7d$. It is also easy to see that each $\check{\bs u}_j,\:j\in [a,b]$ stays inside the decision set $\cD$.

Note that for any $j \in [a,b]$ and any $k \in [d] \setminus \{ k_{\text{fix}}\}$, $\check{\bs u}_j[k]$ coincides with the value of $\bs u_{j'}[k]$ for some $j' \in [a,b]$. This implies that $|\check{\bs u}_j[k] - \bs u_j[k]| \le C_{a \rightarrow b}[k]$ for any $j \in [a,b]$, where $C_{a \rightarrow b}[k]$ is the TV of the optimal solution across coordinate $k$ in bin $[a,b]$. So $\sum\limits_{\substack{k=1 \\ k \neq k_{\text{fix}}}}^d |\check{\bs u}_j[k] - \bs u_j[k]| \le \sum\limits_{\substack{k=1 \\ k \neq k_{\text{fix}}}}^d  C_{a \rightarrow b}[k]  \le C_{a \rightarrow b}$. Thus Property 3 is true.
\end{proof}

\begin{lemma} (\textbf{monotonic bins}) \label{lem:mono-d}
Consider a bin $[a,b]$ with length $\ell$ where the offline optimal sequence is piece-wise maximally monotonic in $[a,b]$ across any coordinate with at-most 4 pieces. Let the TV of the optimal solution within bin $[a,b]$ denoted by $C_{a \rightarrow b}$ be at-most $B/\sqrt{\ell}$. Then we have the regret of FLH-ONS strategy in this bin bounded as
\begin{align}
    \sum_{j=a}^b f_j(\bs x_j) - f_j(\bs u_j)
    &\le 70d^2(8G^2B^2\alpha d + G^2B^2 + 1/\alpha) \log n,
\end{align}
where $\bs x_j$ are the predictions of the FLH-ONS lagorithm.
\end{lemma}
\begin{proof}

We first construct a useful sequence of comparators:\\
$\check{\bs u}_{a:b} =$ \code{generateGhostSequence} $(\bs u_{1:n}, k_{\text{fix}} = 0, u_{\text{fix}} = 0, [a,b])$. 

We remark that as $k_{\text{fix}} = 0 \notin [d]$, the condition in Step 2(a) of Fig.\ref{fig:ghost} is never satisfied.

Next, we employ a two term regret decomposition as follows
\begin{align}
    \underbrace{\sum_{j=a}^{b} f_j(\bs x_j) - f_j(\check{\bs u}_j)}_{T_1} + \underbrace{\sum_{j=a}^{b} f_j(\check{\bs u}_j) - f_j(\bs u_j)}_{T_2}.
\end{align}

By noting that there are only at-most $7d$ change points in the comparator sequence (see Lemma \ref{lem:ghost}), we can sum up the SA regret guarantee from Lemma \ref{lem:stat-flh} against each of the constant sections of $\check{\bs u}_{a:b}$ to obtain
\begin{align}
    T_1 
    &\le 70d^2(8G^2B^2\alpha d + 1/\alpha) \log n. \label{eq:stat-mono}
\end{align}

To bound $T_2$ we use gradient Lipschitzness in Proposition \ref{prop:loss} and look at a coordinate-wise decomposition.
\begin{align}
    T_2
    &\le \sum_{j=a}^b \langle \grad f_j(\bs u_j), \check{\bs u}_j - \bs u_j \rangle + \frac{G^2}{2} \| \check{\bs u}_j - \bs u_j \|_2^2\\
    &\le \frac{\ell G^2 C_{a \rightarrow b}^2}{2} + \sum_{k=1}^d \sum_{j=a}^b \grad f_j(\bs u_j)[k] (\check{\bs u}_j[k] - \bs u_j[k]), \label{eq:mono-eq1}
\end{align}
where in the last line we used that fact that $\| \check{\bs u}_j - \bs u_j \|_2^2 \le \| \check{\bs u}_j - \bs u_j \|_1^2  \le C_{a \rightarrow b}^2$ by Property 3 of Lemma \ref{lem:ghost}, where $C_{a \rightarrow b}$  is the TV of the optimal solution within bin $[a,b]$.

Since $C_{a \rightarrow b} \le B\sqrt{\ell}$, we have the first term in Eq.\eqref{eq:mono-eq1} bounded by $\frac{G^2 B^2}{2}$. Next we proceed to bound the second term in Eq.\eqref{eq:mono-eq1} coordinate-wise. Consider a coordinate $k \in [d]$. We have two cases:

\textbf{Case 1:} When the optimal solution across coordinate $k$ in bin $[a,b]$ has a structure described in Step 2(b) of the \code{generateGhostSequence} procedure of Fig.\ref{fig:ghost}. In this case $\check{\bs u}_j[k] = \bs u_j[k] = \bs u_a[k]$ for $j \in [a,b]$. So
\begin{align}
    \sum_{j=a}^b \grad f_j(\bs u_j)[k] (\check{\bs u}_j[k] - \bs u_j[k]) = 0.
\end{align}

\textbf{Case 2:} When the optimal solution across coordinate $k$ in bin $[a,b]$ has a structure described in Step 2(c) of the \code{generateGhostSequence} procedure of Fig.\ref{fig:ghost}. In this case, we can write
\begin{align}
    \sum_{j=a}^b \grad f_j(\bs u_j)[k] (\check{\bs u}_j[k] - \bs u_j[k])
    &= \sum_{i=1}^7 \sum_{j = \ubar r_i}^{\bar r_i} \grad f_j(\bs u_j)[k] (\check{\bs u}_j[k] - \bs u_j[k]), \label{eq:mono-eq2}
\end{align}
where $[\ubar r_i, \bar r_i]$, $i \in [7]$ are as defined in \code{generateGhostSequence} of Fig.\ref{fig:ghost}.

From Step 2(c)(ii) we have for each $i \in \{2,4,6 \}$, $\check{\bs u}_j[k] = \bs u_j[k] = \bs u_{\ubar r_i}[k]$ for all $j \in [\ubar r_i, \bar r_i]$ if non-empty. So $\sum_{j=\ubar r_i}^{\bar r_i} \grad f_j(\bs u_j)[k] (\check{\bs u}_j[k] - \bs u_j[k]) = 0$ for each $i \in \{2,4,6 \}$.

Next we consider the interval $[\ubar r_1,\bar r_1]$. If within bin $[\ubar r_1, \bar r_1]$, the optimal solution across coordinate $k$ is constant, then $\sum_{j=\ubar r_1}^{\bar r_1} \grad f_j(\bs u_j)[k] (\check{\bs u}_j[k] - \bs u_j[k]) = 0$. Otherwise let $[\ubar r_1, \bar r_1] = [\ubar r_1, p] \cup [p+1,\bar r_1]$ such that the optimal solution is constant in $[\ubar r_1, p]$ and non-decreasing (non-increasing) within $[p+1,\bar r_1]$ across coordinate $k$. Recall from Fig.\ref{fig:ghost} that $\ubar r_1 = a$. Since $\check{\bs u}_j[k] = \bs u_a[k]$ for all $j \in [\ubar r_1, p]$ we get $\sum_{j=\ubar r_1}^{p} \grad f_j(\bs u_j)[k] (\check{\bs u}_j[k] - \bs u_j[k]) = 0$. Further note that due to the presence of bins $[\ubar r_1,p]$ and $[\ubar r_2, \bar r_2]$ the solution $\bs u_j[k]$ for $j \in [p+1,\bar r_1]$ will never touch the boundaries $\pm B$. So by the KKT conditions and using $\check{\bs u}_j[k] = \bs u_{a}[k]$ for $j \in [p+1, \bar r_1]$, we have

\begin{align}
    \sum_{j=p+1}^{\bar r_1} \grad f_j(\bs u_j)[k] (\check{\bs u}_j[k] - \bs u_j[k])
    &= \sum_{j=p+1}^{\bar r_1} \lamda (\bs s_j[k] - \bs s_{j-1}[k]) (\bs u_a[k] - \bs u_j[k])\\
    &= \lamda \left(\bs s_p[k] (\bs u_{p+1}[k] - \bs u_a[k]) - \bs s_{\bar r_1}[k] (\bs u_{\bar r_1}[k] - \bs u_a[k]) \right)\\
    &\quad + \lamda \sum_{j=p+2}^{\bar r_1} |\bs u_j[k] - \bs u_{j-1}[k]|\\
    &= 0 \label{eq:mono-zero},
\end{align}
where the last line is obtained as follows: Observe that $\bs s_p[k] = \bs s_{\bar r_1}[k] = 1 \: (\text{or } -1)$ and $\bs s_p[k] \bs u_{p+1}[k] - \bs s_{\bar r_1}[k] \bs u_{\bar r_1}[k] = -C_{p+1 \rightarrow \bar r_1}$ due to monotonicity of $\bs u_{p+1 : \bar r_1}$

By using similar arguments we used to show Eq.\eqref{eq:mono-zero}, it can be proved that 
\begin{align}
    \sum_{j=\ubar r_i}^{\bar r_i} \grad f_j(\bs u_j)[k] (\check{\bs u}_j[k] - \bs u_j[k])
    &= \sum_{j=\ubar r_i}^{\bar r_i} \grad f_j(\bs u_j)[k] (\bs u_{\ubar r_i}[k] - \bs u_j[k])\\
    &=0,
\end{align}
for $i \in \{3,5\}$.

Further, by using similar arguments we used to handle $[\ubar r_1, \bar r_1]$, it can be shown that 
\begin{align}
    \sum_{j=\ubar r_7}^{\bar r_7} \grad f_j(\bs u_j)[k] (\check{\bs u}_j[k] - \bs u_j[k])
    &=0.
\end{align}

Thus overall by combining Case 1 and 2 and continuing from Eq.\eqref{eq:mono-eq1}, we have $T_2 \le G^2 B^2/2$. Thus the total regret
\begin{align}
    T_1 + T_2
    &\le 70d^2(8G^2B^2\alpha d + 1/\alpha) \log n + G^2 B^2/2\\
    &\le 70d^2(8G^2B^2\alpha d + G^2B^2 + 1/\alpha) \log n,
\end{align}
which concludes the proof.

\end{proof}

\begin{definition} \label{def:struct-gap} We introduce the following definitions for convenience.

\begin{itemize}
    \item For a bin $[a,b] \subseteq \{2,\ldots,n-1 \}$, the offline optimal solution is said to assume Structure 1 across coordinate $k$ if $\bs u_j[k] = \bs u_a[k] \in (-B,B)$ for all $j \in [a,b]$ and $\bs u_b[k] > \bs u_{b+1}[k]$ and $\bs u_a[k] > \bs u_{a-1}[k]$.

    \item For a bin $[a,b] \subseteq \{2,\ldots,n-1 \}$, the offline optimal solution is said to assume Structure 2 across coordinate $k$ if $\bs u_j[k] = \bs u_a[k] \in (-B,B)$ for all $j \in [a,b]$ and $\bs u_b[k] < \bs u_{b+1}[k]$ and $\bs u_a[k] < \bs u_{a-1}[k]$.
    
    \item A bin $[r,s]$ is said to contain Structure 1 and Structure 2 if across some coordinate $k$, the offline optimal solution assumes the form of Structure 1 in an interval $[a,b] \subset [r,s]$ and Structure 2 in some interval $[a',b'] \subset [r,s]$ with $[a,b] \cap [a',b'] = \Phi$.
    
    \item For a bin $[a,b]$, we define $\text{GAP}_{\text{min}}(\beta,[a,b])[k] := \min_{j \in [a,b]} |\bs u_j[k] - \beta|$, where $\beta \in \mathbb{R}$.
\end{itemize}

\end{definition}

Next we provide a lemma analogous to Lemma \ref{lem:within1d}.
\begin{lemma} \label{lem:within-multi}
Consider a bin $[a,b]$ with length $\ell$ where the TV of the offline optimal obeys $C_{a \rightarrow b} \le B/\sqrt{\ell}$. Assume that for some coordinate $k' \in [d]$, $\bs u_{a:b}[k']$ takes the form of Structure 1 or Structure 2. Further suppose that across all coordinates, the offline optimal solution is piece-wise maximally monotonic in $[a,b]$ with at-most 4 pieces. If $  \left |\bs u_a[k'] - \frac{1}{\ell G^2} \sum_{j=a}^b \grad f_j(\bs u_j)[k'] \right| \le B$, then
\begin{align}
    \sum_{j=a}^b f_j(\bs x_j) - f_j(\bs u_j)
    &\le 70d^2(8G^2B^2\alpha d + G^2B^2 + 1/\alpha) \log n - \frac{2 \lamda^2}{\ell G^2},
\end{align}
where $\bs x_j$ are the predictions of FLH-ONS.
\end{lemma}
\begin{proof}
Let $k_{\text{fix}} =  k'$ and $u_{\text{fix}} = \bs u_a[k'] - \frac{1}{\ell G^2} \sum_{j=a}^b \grad f_j(\bs u_j)[k']$.
Consider a comparator sequence\\
$\check{\bs u}_{a:b} =$ \code{generateGhostSequence}$(\bs u_{1:n}, k_{\text{fix}}, u_{\text{fix}}, [a,b] )$. We use a two term regret decomposition

\begin{align}
    \underbrace{\sum_{j=a}^b f_j(\bs x_j) - f_j(\check{\bs u}_j)}_{T_1} + \underbrace{\sum_{j=a}^b f_j(\check{\bs u}_j) - f_j(\bs u_j)}_{T_2}. \label{eq:two-term-within}
\end{align}

By Properties 1 and 2 in Lemma \ref{lem:ghost}, we know that the comparator $\check{\bs u}_{a:b}$ changes only at-most $7d$ times and every single point in the sequence belongs to $\cD$. Hence by strong adaptivity (Lemma \ref{lem:stat-flh}), we have
\begin{align}
    T_1 
    &\le 70d^2(8G^2B^2\alpha d + 1/\alpha) \log n. \label{eq:stat-within}
\end{align}

Further via gradient Lipschitzness in Proposition \ref{prop:loss},
\begin{align}
    T_2
    &\le \sum_{j=a}^b \langle \grad f_j(\bs u_j), \check{\bs u}_j - \bs u_j \rangle + \frac{G^2}{2} \| \check{\bs u}_j - \bs u_j \|_2^2\\
    &= \sum_{j=a}^b \left(\grad f_j(\bs u_j)[k']  (\check{\bs u}_j[k'] - \bs u_j[k']) + \frac{G^2}{2}  (\check{\bs u}_j[k'] - \bs u_j[k'])^2\right) \\
    &\quad + \sum_{\substack{k = 1\\ k \neq k'}}^d \sum_{j=a}^b \left(\grad f_j(\bs u_j)[k]  (\check{\bs u}_j[k] - \bs u_j[k]) + \frac{G^2}{2}  (\check{\bs u}_j[k] - \bs u_j[k])^2\right)\\
    &\le \frac{G^2 B^2}{2} + \sum_{\substack{k = 1\\ k \neq k'}}^d \sum_{j=a}^b \grad f_j(\bs u_j)[k]  (\check{\bs u}_j[k] - \bs u_j[k])\\
    &\quad + \sum_{j=a}^b \left(\grad f_j(\bs u_j)[k']  (\check{\bs u}_j[k'] - \bs u_j[k']) + \frac{G^2}{2}  (\check{\bs u}_j[k'] - \bs u_j[k'])^2\right),
\end{align}
where in the last line we have used the facts that $\sum_{\substack{k = 1\\ k \neq k'}}^d (\check{\bs u}_j[k] - \bs u_j[k])^2 \le \left( \sum_{\substack{k = 1\\ k \neq k'}}^d | \check{\bs u}_j[k] - \bs u_j[k]| \right)^2 \le C^2_{a \rightarrow b} \le B^2/\ell$ by Property 3 of Lemma \ref{lem:ghost} and the TV constraint assumed in the premise of the current lemma.

Since the optimal solution across any coordinate is piece-wise maximally monotonic with at-most 4 pieces, by following the same arguments used in Case 1 and 2 in the proof of Lemma \ref{lem:mono-d}, we can write
\begin{align}
    \sum_{j=a}^b \grad f_j(\bs u_j)[k]  (\check{\bs u}_j[k] - \bs u_j[k])
    &= 0, \label{eq:within-mono}
\end{align}
for any $k \neq k'$.

Recall that $\bs u_j[k'] = \bs u_a[k'] \in (-B,B)$ for all $j \in [a,b]$. Further by our construction, $\check{\bs u}_j[k'] = \bs u_a[k'] - \frac{1}{\ell G^2} \sum_{j=a}^b \grad f_j(\bs u_j)[k']$, for all $j \in [a,b]$. The key observation is to realize that $(\check{\bs u}_j[k'] - \bs u_j[k'])$ stays at a constant value for all $j \in [a,b]$. So we have

\begin{align}
    \sum_{j=a}^b \grad f_j(\bs u_j)[k']  (\check{\bs u}_j[k'] - \bs u_j[k'])
    &= (\check{\bs u}_a[k'] - \bs u_a[k']) \sum_{j=a}^b \grad f_j(\bs u_j)[k']\\
    &= \frac{-1}{\ell G^2} \left(\sum_{j=a}^b \grad f_j(\bs u_j)[k'] \right)^2.\label{eq:withinin-d1}
\end{align}

Further we have,
\begin{align}
    \sum_{j=a}^b \frac{G^2}{2}  (\check{\bs u}_j[k'] - \bs u_j[k'])^2
    &= \frac{1}{2 \ell G^2} \left( \sum_{j=a}^b \grad f_j(\bs u_j)[k'] \right)^2. \label{eq:withinin-d2}
\end{align}

Combining Eq.\eqref{eq:withinin-d1} and \eqref{eq:withinin-d2}, we get
\begin{align}
    \sum_{j=a}^b \grad f_j(\bs u_j)[k']  (\check{\bs u}_j[k'] - \bs u_j[k']) + \frac{G^2}{2}  (\check{\bs u}_j[k'] - \bs u_j[k'])^2
    &= \frac{-1}{2\ell G^2} \left( \sum_{j=a}^b \grad f_j(\bs u_j)[k'] \right)^2\\
    &=_{(a)}  \frac{-1}{2\ell G^2} \left( \lamda \Delta \bs s_{a \rightarrow b}[k']\right)^2\\
    &={(b)} \frac{-2 \lamda^2}{\ell G^2},
\end{align}
where line (a) is due to the KKT conditions and the fact that $\bs u_j[k'] \in (-B,B)$ thus making $\bs \gamma^+_j[k'] = \bs \gamma^-_j[k'] = 0$ and line (b) is due to the fact that $|\Delta \bs s_{a \rightarrow b}[k']| = 2$ for Structure 1 and Structure 2.

Hence overall we have shown that $T_2 \le \frac{G^2 B^2}{2} - \frac{2 \lamda^2}{\ell G^2}$. Combining with Eq.\eqref{eq:stat-within} we conclude that the total regret of the FLH-ONS strategy within the bin $[a,b]$ is bounded by
\begin{align}
    T_1 + T_2
    &\le 70d^2(8G^2B^2\alpha d + 1/\alpha) \log n + \frac{G^2 B^2}{2} - \frac{2 \lamda^2}{\ell G^2}\\
    &\le 70d^2(8G^2B^2\alpha d + G^2B^2 + 1/\alpha) \log n - \frac{2 \lamda^2}{\ell G^2}.
\end{align}

\end{proof}

\begin{lemma} \label{lem:outside-multi}
Consider a bin $[a,b]$ with length $\ell$ where the TV of the offline optimal obeys $C_{a \rightarrow b} \le B/\sqrt{\ell}$. Assume that for some coordinate $k' \in [d]$, $\bs u_{a:b}[k']$ takes the form of Structure 1 or Structure 2. Further suppose that across all coordinates, the offline optimal solution is piece-wise maximally monotonic in $[a,b]$ with at-most 2 pieces.

\textbf{Case 1:} When $\bs u_{a:b}[k']$ takes the form of Structure 1 and  $\bs u_a[k'] - \frac{1}{\ell G^2} \sum_{j=a}^b \grad f_j(\bs u_j)[k'] \ge B$, then 
\begin{align}
    \sum_{j=a}^b f_j(\bs x_j) - f_j(\bs u_j)
    &\le 70d^2(8G^2B^2\alpha d + G^2B^2 + 1/\alpha) \log n -  \frac{\ell G^2}{2} (B - \bs u_a[k'])^2,
\end{align}
and

\textbf{Case 2:} When $\bs u_{a:b}[k']$ takes the form of Structure 2 and  $\bs u_a[k'] - \frac{1}{\ell G^2} \sum_{j=a}^b \grad f_j(\bs u_j)[k'] \le -B$, then 
\begin{align}
    \sum_{j=a}^b f_j(\bs x_j) - f_j(\bs u_j)
    &\le 70d^2(8G^2B^2\alpha d + G^2B^2 + 1/\alpha) \log n -  \frac{\ell G^2}{2} (B + \bs u_a[k'])^2,
\end{align}
where $\bs x_j$ are the predictions of FLH-ONS.
\end{lemma}
\begin{proof}
We consider Case 1. The arguments for the alternate case are similar. We proceed in a similar way as in the proof of Lemma \ref{lem:within-multi}. Let $k_{\text{fix}} =  k'$ and $u_{\text{fix}} = B$.
Consider a comparator sequence $\check{\bs u}_{a:b} =$ \code{generateGhostSequence}$(\bs u_{1:n}, k_{\text{fix}}, u_{\text{fix}}, [a,b] )$. We use a two term regret decomposition as in Eq.\eqref{eq:two-term-within}. Using similar argumets as in the proof of Lemma \ref{lem:within-multi}, we have
\begin{align}
    T_1 
    &\le 70d^2(8G^2B^2\alpha d + 1/\alpha) \log n. \label{eq:stat-outside}
\end{align}

Bounding $T_2$ in a similar fashion as in the proof of Lemma \ref{lem:within-multi}, we have
\begin{align}
    T_2
    &\le \frac{G^2 B^2}{2} + \sum_{j=a}^b \left(\grad f_j(\bs u_j)[k']  (\check{\bs u}_j[k'] - \bs u_j[k']) + \frac{G^2}{2}  (\check{\bs u}_j[k'] - \bs u_j[k'])^2\right), \label{eq:t2-outside}
\end{align}
where we have used Eq.\eqref{eq:within-mono} for bounding the cross terms for coordinates $k \neq k'$

The main difference is in how we handle the last term of Eq.\eqref{eq:t2-outside}. Recall that $\check{\bs u}_j[k'] = B$ and $\bs u_j[k'] = \bs u_a[k']$ for all $j \in [a,b]$. So
\begin{align}
    \sum_{j=a}^b \left(\grad f_j(\bs u_j)[k']  (\check{\bs u}_j[k'] - \bs u_j[k']) + \frac{G^2}{2}  (\check{\bs u}_j[k'] - \bs u_j[k'])^2\right)
    &= \frac{G^2 \ell}{2} (B - \bs u_a[k'])^2 - 2\lamda (B-\bs u_a[k']), \label{eq:la-term-outside}
\end{align}
where the last line is obtained via the KKT conditions and the fact that $\Delta \bs s_{a \rightarrow b}[k'] = -2$ for Case 1. (Recall that $|\bs u_a[k']| < B$ by the definition of Structure 1. So by complementary slackness $\bs \gamma^+_j[k'] = \gamma^-_j[k'] = 0$.)
 
By the premise of the lemma for Case 1, we have $\bs u_a[k'] - \frac{1}{\ell G^2} \sum_{j=a}^b \grad f_j(\bs u_j)[k'] \ge B$. Again by using the KKT conditions and noting that $\Delta s_{a \rightarrow b}[k'] = -2$, we conclude that
\begin{align}
    \lamda \ge \frac{(B-\bs u_a[k']) \ell G^2}{2}.
\end{align}

Plugging this lower bound for $\lamda$ to Eq.\eqref{eq:la-term-outside} and noting that $(B-\bs u_a[k']) \ge 0$, we get
\begin{align}
   \sum_{j=a}^b \left(\grad f_j(\bs u_j)[k']  (\check{\bs u}_j[k'] - \bs u_j[k']) + \frac{G^2}{2}  (\check{\bs u}_j[k'] - \bs u_j[k'])^2\right)
   &\le \frac{-\ell G^2}{2} (B - \bs u_a[k'])^2.
\end{align}

Hence overall, we conclude that
\begin{align}
    T_1 + T_2
    &\le 70d^2(8G^2B^2\alpha d + 1/\alpha) \log n + \frac{G^2 B^2}{2} - \frac{\ell G^2}{2} (B - \bs u_a[k'])^2\\
    &\le 70d^2(8G^2B^2\alpha d + G^2B^2 + 1/\alpha) \log n -  \frac{\ell G^2}{2} (B - \bs u_a[k'])^2.
\end{align}

\end{proof}

\begin{figure}[h!]
	\centering
	\fbox{
		\begin{minipage}{14 cm}
		    \code{fineSplit}: Input - (1) offline optimal sequence $\bs u_{1:n}$ (2) an interval $[r,s] \subseteq [n]$. Across some coordinate $k \in [d]$, the offline optimal solution must take the form of both Structure 1 and 2 or either one of them at-least two times within some appropriate sub-intervals of $[r,s]$.

            \begin{enumerate}
                \item Initialize $\mathcal Q \leftarrow \Phi$, $\mathcal Q' \leftarrow \Phi$.

                \item For each coordinate $k \in [d]$ across which the optimal solution takes the form of Structures 1 and 2 or either one of them at-least two times within some appropriate sub-intervals of $[r,s]$:
                
                \begin{enumerate}
                    \item if $\text{GAP}_{\text{min}}(B,[r,s])[k] > \text{GAP}_{\text{min}}(-B,[r,s])[k]$ then add intervals $[a,b] \subset [r,s]$ where the offline optimal across coordinate $k$ assumes the form of Structure 1 to $\mathcal Q$.
                    
                    \item if $\text{GAP}_{\text{min}}(B,[r,s])[k] \le \text{GAP}_{\text{min}}(-B,[r,s])[k]$ then add intervals $[a,b] \subset [r,s]$ where the offline optimal across coordinate $k$ assumes the form of Structure 2 to $\mathcal Q$.                        
                \end{enumerate}
                
                \item For each bin $[a,b] \in \mathcal Q$ if there exists another interval $[p,q] \in \mathcal Q$ with $[p,q] \subseteq [a,b]$, then remove $[a,b]$ from $\mathcal Q$.
                
                \item Sort intervals in $\mathcal Q$ in increasing order of the left endpoints. (i.e $[a,b] < [p,q]$ if $a < p$).
                
                \item Starting from the first bin, for each bin $[a,b] \in \mathcal Q$:
                
                \begin{enumerate}
                    \item if there exists an interval $[p,q] \in \mathcal Q$ such that $a < p$ and $b < q$, then remove $[p,q]$ from $\mathcal Q$
                \end{enumerate}
                
                \item Add disjoint and maximally continuous intervals that are the subsets of $[r,s] \setminus \{ \cup_{[a,b] \in \mathcal Q} [a,b]\}$ to $\mathcal Q'$ such that the interval $[r,s]$ can be fully covered by disjoint intervals from $\mathcal Q$ and $\mathcal Q'$.
                
                \item Return ($\cQ, \cQ'$).

            \end{enumerate}
		\end{minipage}
	}
	\caption{\emph{\code{fineSplit} procedure.}}
	\label{fig:split}
\end{figure}

\begin{lemma} \label{lem:split-mono1}
Suppose \code{fineSplit} is invoked with input $[r,s]$ such that $C_{r \rightarrow s} \le B/\sqrt{s-r+1}$. The offline optimal solution within any bin $[a,b] \in \cQ$ at Step 7 of \code{fineSplit} procedure in Fig.\ref{fig:split} is piece-wise maximally monotonic in $[a,b]$ with at-most 4 pieces across any coordinate $k \in [d]$. Further there exists a coordinate $k \in [d]$ that satisfy one of the following conditions:
\begin{enumerate}
    \item The offline optimal within bin $[a,b]$ takes the form of Structure 1 across coordinate $k$ and $B-\bs u_a[k] \ge \text{GAP}_{\text{min}}(B,[r,s])[k] \ge \text{GAP}_{\text{min}}(-B,[r,s])[k]$.
    
    \item The offline optimal within bin $[a,b]$ takes the form of Structure 2 across coordinate $k$ and $B+\bs u_a[k] \ge \text{GAP}_{\text{min}}(-B,[r,s])[k] \ge \text{GAP}_{\text{min}}(B,[r,s])[k]$.    
\end{enumerate}
\end{lemma}
\begin{proof}
We start by a basic observation.

FACT 1: Note that $C_{r \rightarrow s} \le B/\sqrt{s-r+1} \le B$. So the $\bs u_{r:s}[k']$ cannot touch both $B$ and $-B$ boundaries.

Consider a bin $[a,b] \in \cQ$. By the construction of \code{fineSplit}, there exists a coordinate $k \in [d]$ across which the optimal solution stays constant within $[a,b]$ and assumes the form of Structure 1 or  2. For the sake of contradiction, let's assume that for some $k' \in [d]$, with $k' \neq k$, the optimal solution is maximally monotonic in $[a,b]$ with at-least 5 pieces across the coordinate $k'$. This can happen only when the optimal solution increases (decreases) then decreases (increases) then increases (decreases) then decreases (increases) and finally increase (decrease) again within bin $[a,b]$ and evolve arbitrarily there on-wards. Combined with FACT 1, such a behaviour can result in one of the following configurations across  the coordinate $k'$:

\begin{itemize}
    \item Both Structure 1 and Structure 2 are formed. 
    
    \item Only Structure 2 is formed at-least two times. This means that if $[x,y] \subset [a,b]$ is a maximally monotonic section with $\bs u_{x:y}[k']$ increasing, then $\bs u_y[k'] = B$. Then $\text{GAP}_{\text{min}}(-B,[r,s])[k'] > \text{GAP}_{\text{min}}(B,[r,s])[k']=0$.
    
    \item Only Structure 1 is formed at-least two times. This means that if $[x,y] \subset [a,b]$ is a maximally monotonic section with $\bs u_{x:y}[k']$ decreasing, then $\bs u_y[k'] = -B$. Then $\text{GAP}_{\text{min}}(B,[r,s])[k'] > \text{GAP}_{\text{min}}(-B,[r,s])[k']=0$.        
\end{itemize}

In all of the above cases, at-least one sub-interval of $[a,b]$ will be added to $\cQ$ at Step 2(a) or 2(b). This would imply that at Step 3, the bin $[a,b]$ is removed from $\cQ$ and never added again resulting in a contradiction.

The last statement of the Lemma is immediate from Steps 2(a)-(b) of \code{fineSplit}.
\end{proof}

\begin{lemma} \label{lem:split-mono2}
Suppose \code{fineSplit} is invoked with input $[r,s]$ such that $C_{r \rightarrow s} \le B/\sqrt{s-r+1}$. The offline optimal solution within any interval $[p,q] \in \mathcal Q'$ at Step 7 of \code{fineSplit} procedure in Fig.\ref{fig:split} is piece-wise maximally monotonic in $[p,q]$ with at-most 4 pieces across any coordinate.
\end{lemma}
\begin{proof}
Consider a coordinate $k \in [d]$ and a bin $[p,q] \in \mathcal Q'$. We provide the arguments for the case when $\text{GAP}_{\text{min}}(-B,[r,s])[k] \ge \text{GAP}_{\text{min}}(B,[r,s])[k]$. The arguments for the complementary case are similar. We start by stating two facts. 

FACT 1: $\text{GAP}_{\text{min}}(-B,[p,q])[k] > 0$.

To see this, assume for the sake of contradiction that $\text{GAP}_{\text{min}}(-B,[p,q])[k] = 0$. Then this means that $\text{GAP}_{\text{min}}(-B,[p,q])[k] = \text{GAP}_{\text{min}}(B,[r,s])[k] =0$. So the optimal solution across coordinate $k$, $\bs u_{r:s}[k]$ must touch both $B$ and $-B$ at distinct time points in $[r,s]$. This would violate the TV constraint that $C_{p \rightarrow q} \le B/\sqrt{s-r+1} \le B$, thus yielding a contradiction.

FACT2: It is not the case that there exists two intervals $[p_1,q_1],[p_2,q_2] \subset [p,q]$ within which the offline optimal takes the form of Structure 2 across the coordinate $k \in [d]$.

Let's prove the above fact via contradiction. Assume that there exists  $[p_1,q_1],[p_2,q_2] \subset [p,q] \in \cQ'$ such that the offline optimal takes the form of Structure 2 within them across the coordinate $k \in [d]$. Then $[p_i,q_i]$ ($i=1,2$) must have been added to $\mathcal Q$ in step 2(b) of \code{fineSplit}. Since intervals in $\mathcal Q$ don't overlap with intervals in $\mathcal Q'$ due to Step 6, this would mean that the interval $[p_i,q_i]$ ($i=1,2$) got removed from $\mathcal Q$ later. 

\textbf{Case 1:} Consider the case where $[p_i,q_i]$ ($i=1,2$) has been removed at Step 5(a). This means that there exists an interval $[a,b] \subseteq [r,s]$ where the offline optimal has Structure 1 or 2 across some coordinate $k' \neq k$ and $[p_i,q_i] \cap [a,b] \ne \Phi$. Observe that $[a,b]$ is never removed from $\cQ$ since we are processing bins in sorted order at Step 4-5. This would contradict the fact that intervals in $\mathcal Q$ don't overlap with intervals in $\mathcal Q'$ due to Step 6.

\textbf{Case 2:} Consider the case where $[p_i,q_i]$ ($i=1,2$) has been removed at Step 3. This means that there exists an interval $[x,y] \subseteq [p_i,q_i]$ where the offline optimal assumes Structure 1 or 2 across some coordinate $k' \neq k$. If $[x,y]$ is present in the final $\cQ$ in Step 7, then this would again warrant a contradiction to the non-overlapping property between the intervals of $\mathcal Q$ and $\mathcal Q'$. If $[x,y]$ is removed at a later point through Step 5(a), by using similar arguments as in Case 1 yields a contradiction. Thus we conclude that the FACT 2 is true.

FACT 3:  It is not the case that there exists two intervals $[p_1,q_1],[p_2,q_2] \subset [p,q]$ within the offline optimal takes the form of Structure 1 in $[p_1,q_1]$ and Structure 2 in $[p_2,q_2]$ across the coordinate $k \in [d]$.

The above fact can be proven using similar arguments that are used in proving FACT 2.

In light of FACT 1, FACT 2 and FACT 3, we conclude the statement of the lemma.

\end{proof}

Next we introduce a structural lemma analogous to Lemma \ref{lem:lamda-len}.

\begin{lemma} ($\bs \lamda$\textbf{-length lemma}) \label{lem:lamda-len-multi}
Consider a bin $[a,b] \subseteq \{2,\ldots,n-1 \}$ with length $\ell$. Suppose that within this bin, the offline optimal solution sequence assumes the form of Structure 1 or Structure 2 across some coordinate $k \in [d]$, then $\lamda \le \frac{G_\infty \ell}{2}$, where $G_\infty$ is as in Assumption B2.
\end{lemma}
\begin{proof}[Proof Sketch]
The arguments for this proof are almost identical to that used for proving Lemma \ref{lem:lamda-len}. We outline the parts where there are differences.
We provide the arguments for Structure 2. Structure 1 can be handled similarly. Let the optimal sign assignments across coordinate $k$ be written as $\bs s_j[k] = -1 + \epsilon_j$ where $\epsilon_j \in [0,2]$ and $j \in [a,b]$.
From the KKT conditions, we can write:
\begin{align}
    \grad f_a(\bs u_a)[k] &= \lamda \epsilon_a\\
    \grad f_{a+1}(\bs u_{a+1})[k] &= \lamda (\epsilon_{a+1} - \epsilon_{a})\\
    &\vdots\\
    \grad f_{b-1}(\bs u_{b-1})[k]  &= \lamda (\epsilon_{b-1} - \epsilon_{b-1})\\
    \grad f_b(\bs u_b)[k] &= \lamda (2 - \epsilon_{b-1})
\end{align}
Define the vector $\bs z = [\epsilon_a,\epsilon_{a+1} - \epsilon_{a},\ldots,2 - \epsilon_{b-1}]^T$. As noted in the proof of Lemma \ref{lem:lamda-len}, we must have $\| \bs z\|_\infty > 0$. Let $j^*$ be such that $\|\bs z \|_\infty = |\bs z[j^*]|$. Then $\lamda = \grad f_{a+j^*-1}(\bs u_{a+j^*-1})[k]/\|\bs z \|_\infty$. From the optimization problem considered in the proof of Lemma \ref{lem:lamda-len}, we have $\| \bs z\|_\infty \ge 2/\ell$. Since $\| \grad f_j(\bs u_j)\|_\infty \le G_\infty$ for all $j \in [n]$ by Assumption B2, we have $\lamda =  \grad f_{a+j^*-1}(\bs u_{a+j^*-1})[k]/\|\bs z \|_\infty \le (G_\infty \ell)/2$.

\end{proof}

\begin{lemma} (\textbf{large margin bins}) \label{lem:large-margin-multi}
Assume that $\lamda \ge d^{1.5} \phi \frac{n^{1/3}}{c_n^{1/3}}$ for a constant $\phi = \sqrt{70(8G^2B^2\alpha + G^2B^2 + 1/\alpha)}$ that does not depend on $n$ and $C_n$. Consider a bin $[i_s,i_t] \in \cP$ within which the offline optimal solution takes the form of Structure 1 or Structure 2 (or both) across a coordinate $k \in [d]$ for some appropriate sub-intervals of $[i_s,i_t]$. Let $\mu_{\text{th}} = \sqrt{\frac{140d^{1.5}(8G^2B^2\alpha + G^2B^2 + 1/\alpha)  G_\infty C_n^{1/3} \log n }{G^2 \phi n^{1/3}}}$. Then\\ $\text{GAP}_{\text{min}}(-B,[i_s,i_t])[k] \vee \text{GAP}_{\text{min}}(B,[i_s,i_t])[k] \ge \mu_{\text{th}}$,\\ whenever $C_n \le \left( \frac{B^2 G^2 \phi}{560 d^{1.5} (8G^2B^2\alpha + G^2B^2 + 1/\alpha) G_\infty \log n} \right)^3 n = \tilde O(n)$.
\end{lemma}
\begin{proof}
Suppose $\text{GAP}_{\text{min}}(-B,[i_s,i_t])[k] < \mu_{\text{th}}$. Then the largest value of the optimal solution across coordinate $k$ attained within this bin $[i_s,i_t]$ is at-most $-B+ \mu_{\text{th}} + B/\sqrt{n_i}$ (recall $n_i := i_t - i_s + 1$ and $C_i \le B/\sqrt{n_i}$ due to Lemma \ref{lem:key-part-multi}). So $\text{GAP}_{\text{min}}(B,[i_s,i_t])[k] \ge 2B - \mu_{\text{th}} - B/\sqrt{n_i}$. Our goal is to show that whenever $C_n$ obeys the constraint stated in the lemma, we must have
\begin{align}
    2B - \mu_{\text{th}} - B/\sqrt{n_i}
    &\ge \mu_{\text{th}}. \label{eq:gap-multi}
\end{align}

Let $\ell_i$ be the length of a sub-interval of $[i_s,i_t]$ where the offline optimal solution assumes the form of Structure 1 or Structure 2. Due to Lemma \ref{lem:lamda-len-multi}, we have
\begin{align}
    n_i \ge \ell_i \ge \frac{2 \lamda}{G_\infty} \ge \frac{2 d^{1.5}\phi n^{1/3}}{G_\infty C_n^{1/3} }  \label{eq:len-lb-multi}
\end{align}
where the last inequality follows due to the condition assumed in the current lemma. So a sufficient condition for Eq.\eqref{eq:gap-multi} to be true is
\begin{align}
    2B \ge 2 \left(2 \sqrt{\frac{140d^{1.5}(8G^2B^2\alpha + G^2B^2 + 1/\alpha)  G_\infty C_n^{1/3} \log n }{G^2 \phi n^{1/3}}} \vee B\sqrt{\frac{G_\infty C_n^{1/3} } {2 d^{2.5/2} \phi n^{1/3}}}\right).
\end{align}

Recall that by Assumption B2, we have $G \wedge G_\infty \wedge B \ge 1$.  So the above maximum will be attained by the first term and can be further simplified as
\begin{align}
    2B \ge 4 \sqrt{\frac{140d^{1.5}(8G^2B^2\alpha + G^2B^2 + 1/\alpha)  G_\infty C_n^{1/3} \log n }{G^2 \phi n^{1/3}}}.
\end{align}

The above condition is always satisfied whenever $C_n \le \left( \frac{B^2 G^2 \phi}{560 d^{1.5} (8G^2B^2\alpha + G^2B^2 + 1/\alpha) G_\infty \log n} \right)^3 n$.

At this point, we have shown that $\text{GAP}_{\text{min}}(-B,[i_s,i_t])[k] < \mu_{\text{th}} \implies \text{GAP}_{\text{min}}(B,[i_s,i_t])[k] \ge \mu_{\text{th}}$ under the conditions of the lemma. Taking the contrapositive yields $\text{GAP}_{\text{min}}(B,[i_s,i_t])[k] < \mu_{\text{th}} \implies \text{GAP}_{\text{min}}(-B,[i_s,i_t])[k] \ge \mu_{\text{th}}$.
\end{proof}

\begin{lemma} (\textbf{high} $\bs \lamda$ \textbf{regime})\label{lem:high-lamda-multi}
Suppose the optimal dual variable $\lamda \ge d^{1.5}\phi \frac{n^{1/3}}{C_n^{1/3}} = \Omega\left( \frac{n^{1/3}}{C_n^{1/3}} \right)$ for \\
$\phi = \sqrt{70(8G^2B^2\alpha + G^2B^2 + 1/\alpha)}$ that does not depend on $n$ and $C_n$. We have the regret of FLH-ONS strategy bounded as
\begin{align}
    \sum_{t=1}^{n} f_t(\bs x_t) - f_t(\bs u_t)
    &=  \tilde O \left( 140 d^2(8G^2B^2\alpha d + G^2B^2 + 1/\alpha) (n^{1/3}C_n^{2/3} \vee 1) \right) \mathbb{I}\{ C_n > 1/n\}\\
    &+ \tilde O\left( d(8G^2B^2\alpha d + 1/\alpha \right) \mathbb{I}\{C_n \le 1/n \}),
\end{align}
where $\bs x_t$ is the prediction of FLH-ONS at time $t$ and $\mathbb{I}\{ \cdot \}$ is the boolean indicator function taking values in $\{0 ,1 \}$.
\end{lemma}
\begin{proof}
Throughout the proof we assume that $C_n \left( \frac{B^2 G^2 \phi}{560 d^{1.5} (8G^2B^2\alpha + G^2B^2 + 1/\alpha) G_\infty \log n} \right)^3 n$. Otherwise the trivial regret bound of $\tilde O(n)$ is near minimax optimal.

First we consider the regime where $C_n \ge 1/n$. It is useful to define the following annotated condition.

\textbf{Condition (A):} Let a bin $[r,s]$ be given. For some coordinate $k' \in [d]$, there exists disjoint intervals $[r_1,s_1], [r_2,s_2] \subset [r,s]$ that satisfy at-least one of the following: (i) $\bs u_{r_1:s_1}[k']$ has the form of Structure 1 and $\bs u_{r_2:s_2}[k']$ has the form of Structure 2; (ii) Both $\bs u_{r_1:s_1}[k']$ and $\bs u_{r_2:s_2}[k']$ have the form of Structure 1; (iii) Both $\bs u_{r_1:s_1}[k']$ and $\bs u_{r_2:s_2}[k']$ have the form of Structure 2. 

The above condition is basically the prerequisite for the \code{fineSplit} procedure of Fig.\ref{fig:split}.

Let $[i_s,i_t] \in \cP$ be a bin that satisfy Condition (A) for a coordinate $k' \in [d]$. Here $\cP$ is the partition obtained in Lemma \ref{lem:key-part-multi}.

Let $(\cQ, \cQ')$ be the collections of intervals obtained by invoking the \code{fineSplit} procedure with the bin $[i_s,i_t]$ as input. Let's write $\cQ \cup \cQ' \cup \{ \Phi \}$ as a collection of disjoint consecutive intervals as follows:
\begin{align}
    \cQ \cup \cQ' \cup \{ \Phi \} &:= \{[i_s,\ubar i_1-1], [\ubar i_1, \bar i_1], [\ubar i'_1, \bar i'_1], \ldots,[\ubar i_{m^{(i)}}, \bar i_{m^{(i)}}], [\ubar i'_{m^{(i)}}, \bar i'_{m^{(i)}}] \}, \label{eq:part-refine}
\end{align}
with $ \bar i'_{m^{(i)}} = i_t$.

Here we follow the convention that the bins $[\ubar i_p, \bar i_p] \in \cQ$ and $[\ubar i'_p, \bar i'_p] \in \cQ' \cup \{ \Phi \}$ for all $p \in [m^{(i)}]$. Similar to the proof of Lemma \ref{lem:high-lamda}, for enforcing this convention, we may have to set either of the bins $[i_s,\ubar i_1 -1]$ or $[\ubar i'_{m^{(i)}}, \bar i'_{m^{(i)}}]$ to  be empty. More precisely, if $i_s$ belongs to some interval in $\cQ$, then we set the first sub-interval $[i_s,\bar i_1-1]$ to be empty by setting $\bar i_1 = i_s$. Similarly, if $i_t$ belongs to some interval in $\cQ$, we treat the sub-interval $[\ubar i'_{m^{(i)}}, \bar i'_{m^{(i)}}]$ as empty by setting $\ubar i'_{m^{(i)}} = i_t+1$. Further some of the intervals: $[\ubar i'_k, \bar i'_k]$, $k \in [m^{(i)}]$ can be empty. For example if $\ubar i_{k+1}=\bar i_k+1$, then $[\ubar i'_k, \bar i'_k]$ is treated as empty.

Note that if the first sub-interval $[i_s,\ubar i_1 -1]$ is non-empty then it must belong to $\cQ'$ according to our convention. By Lemma \ref{lem:split-mono2} and Lemma \ref{lem:mono-d},
\begin{align}
    \sum_{j=i_s}^{\ubar i_1-1} f_j(\bs x_j) - f_j(\bs u_j)
    &= \tilde O\left (70d^2(8G^2B^2\alpha d + G^2B^2 + 1/\alpha) \log n \right). \label{eq:mono-left}
\end{align}

We proceed to bound the regret in $[\ubar i_1, \bar i'_{m^{(i)}}]$. Let $\cP_1^{(i)}$ denote the collection of bins among $\cQ = \{ [\ubar i_1, \bar i_1], \ldots, [\ubar i_{m^{(i)}}, \bar i_{m^{(i)}}]\}$ which satisfy the property in Lemma \ref{lem:within-multi}. Let $|\cP_1^{(i)}| := m_1^{(i)}$ and their lengths be denoted by $\{\ell^{(1)}_{1^{(i)}}, \ldots, \ell^{(1)}_{m_1^{(i)}}\}$. These bins will be referred as \emph{Type 1} bins henceforth.

Similarly let $\cP_2^{(i)}=\cQ \setminus \cP_1^{(i)}$ which satisfy either of the properties in Lemma \ref{lem:outside-multi}. Let $|\cP_2^{(i)}| := m_2^{(i)}$ and their lengths be denoted by $\{\ell^{(2)}_{1^{(i)}}, \ldots, \ell^{(2)}_{m_2^{(i)}}\}$. These bins will be referred as \emph{Type 2} bins henceforth. A bin $[a,b] \in \cP_2^{(i)}$ satisfy at-least one of the following properties
\begin{enumerate}
    \item[P1:] For some coordinate $k \in [d]$, the offline optimal satisfy the condition of Case 1 in Lemma \ref{lem:outside-multi} and $B-\bs u_a[k] \ge \mu_{th}$.
    
    \item[P2:] For some coordinate $k \in [d]$, the offline optimal satisfy the condition of Case 2 in Lemma \ref{lem:outside-multi} and $B+\bs u_a[k] \ge \mu_{th}$.    
\end{enumerate}

To see this, let's inspect the way in which the bin $[a,b]$ has been added to $\cQ$ when we invoke \code{fineSplit} with the input bin $[i_s,i_t]$. If $[a,b]$ has been added via Step 2-(a), then we have $\text{GAP}_{\text{min}}(B,[i_s,i_t])[k] > \text{GAP}_{\text{min}}(-B,[i_s,i_t])[k]$ for a coordinate $k$. By Lemma \ref{lem:large-margin-multi} it holds that  $\text{GAP}_{\text{min}}(B,[i_s,i_t])[k] \ge \mu_{\text{th}}$ under the $C_n$ regime we consider. So $B - \bs u_a[k] \ge \text{GAP}_{\text{min}}(B,[i_s,i_t])[k] \ge \mu_{\text{th}}$ where the first inequality follows by the definition of GAP (see Definition \ref{def:struct-gap}). Further, observe that $\bs u_a[k'] - \frac{1}{\ell G^2} \sum_{j=a}^b \grad f_j(\bs u_j)[k'] < -B$ is never satisfied, where $\ell = b-a+1$. Otherwise it will imply that $-\frac{1}{\ell G^2} \sum_{j=a}^b \grad f_j(\bs u_j)[k']  = \frac{2\lamda}{\ell G^2}< -B - \bs u_a[k'] \le 0$ which is not true as $\lamda \ge 0$. We must also have $\bs u_a[k'] - \frac{1}{\ell G^2} \sum_{j=a}^b \grad f_j(\bs u_j)[k'] \notin [-B,B]$. Otherwise, bin $[a,b]$ would have been already added to $\cP_1^{(i)}$ and would have never present in $\cP_2^{(i)}$. So we conclude that property P1 follows. Property P2 can also be shown to be true using similar arguments when the bin $[a,b]$ has been added to $\cQ$ via Step 2-(b) of \code{fineSplit}.

Each bin  $[\ubar i_k, \bar i_k]$, $k \in [m^{(i)}]$ of Type 1 and Type 2 can be paired with an adjacent bin $[\ubar i'_k, \bar i'_k] \in \cQ' \cup \{ \Phi\}$, $k \in [m^{(i)}]$ which is either empty or the optimal sequence displays a piece-wise maximally monotonic behaviour in $[\ubar i'_k, \bar i'_k]$ across all coordinates as recorded in Lemma \ref{lem:split-mono2}. 

Note that $m_1^{(i)}+m_2^{(i)} = m^{(i)}$. Let the total regret contribution from Type 1 bins along with their pairs and Type 2 bins along with their pairs be referred as $R_1^{(i)}$ and $R_2^{(i)}$ respectively.

For a bin $[a,b] \in \cP_2^{(i)}$, in either of the cases covered by the properties P1 and P2, we have by Lemma \ref{lem:outside-multi} that
\begin{align}
    \sum_{j=a}^b f_j(\bs x_j) - f_j(\bs u_j)
    &\le 70d^2(8G^2B^2\alpha d + G^2B^2 + 1/\alpha) \log n -  \frac{\ell G^2}{2} \mu_{\text{th}}^2,
\end{align}
Let $[a',b'] \in \cQ'\cup \{ \Phi\}$ be the pair assigned to $[a,b]$. If it is non-empty, then due to Lemma \ref{lem:split-mono2} and Lemma \ref{lem:mono-d} the regret from the bin $[a',b']$ is at-most $70d^2(8G^2B^2\alpha d + G^2B^2 + 1/\alpha) \log n$.

So we can bound $R_2^{(i)}$ as
\begin{align}
   R_2^{(i)} 
   &\le \sum_{j=1^{(i)}}^{m_2^{(i)}} \left( \left( 70d^2(8G^2B^2\alpha d + G^2B^2 + 1/\alpha) \log n -  \frac{\ell^{(2)}_j G^2}{2} \mu_{\text{th}}^2 \right) + 70d^2(8G^2B^2\alpha d + G^2B^2 + 1/\alpha) \log n \right)\\
   &\le m_2^{(i)} 140d^2(8G^2B^2\alpha d + G^2B^2 + 1/\alpha) \log n - \frac{ G^2\mu_{\text{th}}^2}{2} (\sum_{j=1^{(i)}}^{m_2^{(i)}} \ell^{(2)}_j ),
\end{align}
From Eq.\eqref{eq:len-lb-multi}, we have $\ell^{(2)}_j \ge \frac{2 \phi d^{1.5}n^{1/3}}{G_\infty C_n^{1/3}}$ for $j \in \{1^{(i)},\ldots,m_2^{(i)}\}$ under the regime of $\lamda$ we consider. So we can continue as

\begin{align}
    R_2^{(i)}
    &\le 140 m_2^{(i)} d^2(8G^2B^2\alpha d + G^2B^2 + 1/\alpha) \log n - G^2 \mu_{\text{th}}^2 m_2^{(i)} \frac{\phi d^{1.5} n^{1/3}}{G_\infty C_n^{1/3}}\\
    &\le 140 m_2^{(i)} d^3(8G^2B^2\alpha + G^2B^2 + 1/\alpha) \log n - G^2 \mu_{\text{th}}^2 m_2^{(i)} \frac{\phi d^{1.5} n^{1/3}}{G_\infty C_n^{1/3}}\\
    &= 0,
\end{align}
where the last line is obtained by plugging in the value of $\mu_{\text{th}}$ as in Lemma \ref{lem:large-margin-multi}.

So by refining every interval in $\cP$ (recall that $\cP$ is from Lemma \ref{lem:key-part-multi}) that satisfy Condition (A) and summing the regret contribution from all Type 2 bins and their pairs across all refined intervals in $\cP$ yields

\begin{align}
    \sum_{i=1}^M R_2^{(i)}
    &\le 0,
\end{align}
where we recall that $M := |\cP| = O(n^{1/3}C_n^{2/3} \vee 1)$ and assign $R_2(i) = 0$ for intervals in $\cP$ that do not satisfy Condition (A).

For any Type 1 bin, its regret contribution can be bounded by Lemma \ref{lem:within-multi}. The regret contribution from its pair can be bounded by Lemma \ref{lem:mono-d} as before. So we have

\begin{align}
    R_1^{(i)}
    &\le \sum_{j=1}^{m_1^{(i)}} \left( \left(70d^2(8G^2B^2\alpha d + G^2B^2 + 1/\alpha) \log n - \frac{2\lamda^2}{\ell^{(1)}_{j^{(i)}} G^2} \right) + 70d^2(8G^2B^2\alpha d + G^2B^2 + 1/\alpha) \log n  \right)\\
    &=  140 m_1^{(i)} d^2(8G^2B^2\alpha d + G^2B^2 + 1/\alpha) \log n - \frac{2 \lamda^2}{G^2} \sum_{j=1}^{m_1^{(i)}} \frac{1}{\ell^{(1)}_{j^{(i)}}}.
\end{align}

So by refining every interval in $\cP$ that satisfy Condition (A) and summing the regret contribution from all Type 2 bins and their pairs across all refined intervals in $\cP$ yields
\begin{align}
    \sum_{i=1}^M R_1^{(i)}
    &\le   140 d^2(8G^2B^2\alpha d + G^2B^2 + 1/\alpha) \log n \sum_{i=1}^M m_1^{(i)} - \frac{2 \lamda^2}{G^2}  \sum_{i=1}^M \sum_{j=1}^{m_1^{(i)}} \frac{1}{\ell^{(1)}_{j^{(i)}}}\\
    &\le 140 d^2(8G^2B^2\alpha d + G^2B^2 + 1/\alpha) M_1 \log n   - \frac{2 \lamda^2}{G^2} \frac{M_1^2}{n},\\
    &\le 140 d^3(8G^2B^2\alpha  + G^2B^2 + 1/\alpha) M_1 \log n   - \frac{2 \lamda^2}{G^2} \frac{M_1^2}{n}
    \label{eq:type1-multi}
\end{align}
where in the last line:  a) we define $M_1 := \sum_{i=1}^M m_1^{(i)}$ with the convention that $m_1^{(i)} = 0$ if the $i^{\text{th}}$ bin in $\cP$ doesn't satisfy Condition (A); b) applied AM-HM inequality and noted that $\sum_{i=1}^M \sum_{j=1}^{m_1^{(i)}} \ell^{(1)}_{j^{(i)}} \le n$.

To further bound Eq.\eqref{eq:type1-multi}, we consider two separate regimes as follows. 

Recall that $\lamda \ge d^{1.5}\phi \frac{n^{1/3}}{C_n^{1/3}}$. So continuing from Eq.\eqref{eq:type1-multi},
\begin{align}
    140 d^3(8G^2B^2\alpha  + G^2B^2 + 1/\alpha) M_1 \log n  - 2 \lamda^2 \frac{M_1^2}{G^2 n}
    &\le 140 d^3(8G^2B^2\alpha  + G^2B^2 + 1/\alpha) M_1 \log n\\
    &\quad - 2 d^{2.5} \phi^2 \frac{n^{2/3}}{C_n^{2/3}}  \frac{M_1^2}{G^2 n}\\
    &\le 0,
\end{align}
whenever $M_1 \ge \frac{70(8G^2B^2\alpha + G^2B^2 + 1/\alpha)\log n}{\phi^2} n^{1/3}C_n^{2/3} = \tilde \Omega(n^{1/3}C_n^{2/3})$.

In the alternate regime where $M_1 \le \left(\frac{70(8G^2B^2\alpha + G^2B^2 + 1/\alpha)\log n}{\phi^2} n^{1/3}C_n^{2/3} \vee 1 \right) = \tilde O(n^{1/3}C_n^{2/3} \vee 1)$, we trivially obtain $\sum_{i=1}^M R_1^{(i)} = \tilde O \left( 140 d^2(8G^2B^2\alpha d + G^2B^2 + 1/\alpha) (n^{1/3}C_n^{2/3} \vee 1) \right)$.

The regret contribution from all sub-bins that starts at $i_s$ $i \in [M]$ which are not paired in Eq.\eqref{eq:part-refine} is only at-most $\tilde O(d^{2.5} (n^{1/3}C_n^{2/3} \vee 1))$ by adding the bound of Eq.\eqref{eq:mono-left} across all $O(n^{1/3}C_n^{2/3} \vee 1)$ bins in $\cP$.

Throughout the entire proof we have assumed that $m_1^{(i)}$ and $m_2^{(i)}$ are non-zero for some bin $[i_s,i_t] \in \cP$. Not meeting this criterion will only make the arguments easier as explained below.

We have shown that the total regret contribution from the refined bins $\sum_{i=1}^M R_1^{(i)} + R_2^{(i)} =  \tilde O(n^{1/3}C_n^{2/3} \vee 1)$, we trivially obtain $\sum_{i=1}^M R_1^{(i)} = \tilde O \left( 140 d^2(8G^2B^2\alpha d + G^2B^2 + 1/\alpha) (n^{1/3}C_n^{2/3} \vee 1) \right)$ under the conditions of the lemma, where we have taken $R_1^{(i)} = R_2^{(i)} =0$ if the $i^{\text{th}}$ bin $[i_s,i_t] \in \cP$ doesn't satisfy Condition (A) across any coordinate.

If a bin doesn't satisfy Condition (A) across any coordinate, then the offline optimal solution within that bin assumes a piece-wise maximally monotonic structure with at-most 4 pieces across any coordinate. By Lemma \ref{lem:mono-d}, the regret within such bins is $\tilde O\left( 70d^2(8G^2B^2\alpha d + G^2B^2 + 1/\alpha) \right)$. Since there can be at-most $O(n^{1/3}C_n^{2/3} \vee 1)$ such bins in $\cP$, the total regret contribution from those bins is again $\tilde O \left( 70d^2(8G^2B^2\alpha d + G^2B^2 + 1/\alpha) (n^{1/3}C_n^{2/3} \vee 1) \right)$. Now putting everything together yields the lemma.

If $C_n \le 1/n$, then we have
\begin{align}
    \sum_{t=1}^{n} f_t(\bs x_j) - f_t(\bs u_t)
    &\le \sum_{t=1}^{n} f_t(\bs x_j) - f_t(\bs u_1) + \sum_{t=1}^{n} f_t(\bs u_1) - f_t(\bs u_t)\\
    &\le_{(a)} \tilde O \left ( 10d(8G^2B^2\alpha d + 1/\alpha) \log n \right) + GnC_n\\
    &= \tilde O\left( d(8G^2B^2\alpha d + 1/\alpha \right)
\end{align}
where line (a) follows from the fact that $f_t$ is $G$ Lipschitz.
\end{proof}

\begin{proof} \textbf{of Theorem} \ref{thm:ec-d}. The proof is immediate from the results of Lemmas \ref{lem:low-lamda-multi} and \ref{lem:high-lamda-multi}.

\end{proof}

\section{Reparametrization of certain polytopes to box} \label{app:reparam}

\begin{proposition}
Consider an online problem with losses $f_t$ that are $\alpha$ exp-concave on the decision set $\cD = \{\bs x \in \mathbb{R}^d :  \bs c \le \bs {Ax} \le \bs b\}$ such that $\bs A$ is full rank and $\bs 0 < \bs b - \bs c$.

We can reparametrize this into an equivalent online learning problem with losses $\tilde f_t(\bs z) = f_t(\bs A^{-1} (\bs D^{-1}(\bs z + \bs 1) + \bs c))$ that are $\alpha$ exp-concave on the decision set $\tilde{\cD} = \{\ \bs z \in \mathbb{R}^d : \| \bs z\|_\infty \le 1 \}$, where $\bs D = \diag(2/(\bs b[1] = \bs c[1]),\ldots, 2/(\bs b[d] - \bs c[d]))$ and $\bs 1$ is the vector of ones in $\mathbb{R}^d$.

Further if the losses $f_t$ are $G$ Lipschitz in $\cD$, then the losses $\tilde f_t$ are $\| \bs A^{-1} \bs D^{-1}\|_{\text{op}} G$ Lispchitz in $\tilde{\cD}$.

\end{proposition}
\begin{proof}
We have,
\begin{align}
    &\bs c \le \bs{Ax} \le \bs b\\
    \iff& \bs 0 \le \bs{Ax} - \bs c \le \bs b - \bs c.
\end{align}
Then we have $\bs 0 \le \bs D(\bs{Ax} - \bs c) \le (2) \bs 1$. This equivalent to $-\bs 1 \le \bs D(\bs{Ax} - \bs c) - \bs 1 \le \bs 1$. By putting $\bs z = \bs D(\bs{Ax} - \bs c)  - \bs 1$ we can rewrite the original decision set as $\| \bs z\|_\infty \le 1$.

Since $\bs A$ is full rank, there is a one-one mapping between the original decision set $\cD$ and the new decision set $\tilde{\cD} := \{\bs z \in \mathbb{R}^d : \| \bs z\|_\infty \le 1 \}$. Given a $\bs z \in \tilde{\cD}$, we can find the corresponding point $\bs x \in \cD$ as $\bs x = \bs A^{-1} (\bs D^{-1}(\bs z + \bs 1) + \bs  c)$. So the losses in the new parametrization becomes $\tilde f_t(\bs z) = f_t(\bs A^{-1} (\bs D^{-1}(\bs z + \bs 1) + \bs  c))$.

Let $B := \bs A^{-1} \bs D^{-1}$ and $\bs d:= \bs A^{-1} \bs D^{-1} \bs 1 + \bs A^{-1} \bs c$ so that $\tilde f_t(\bs z) = f_t(\bs B \bs z + \bs d)$. Then we have
\begin{align}
    \grad \tilde f_t(\bs z) &= \bs B^T \grad f_t(\bs B \bs z + \bs d)\\
    &=  \bs B^T \grad f_t(\bs x),
\end{align}
for a point $\bs x = (\bs B \bs z + \bs d) \in \cD$.

Similarly
\begin{align}
    \grad ^2 \tilde f_t(\bs z) &= \bs B^T \grad^2 f_t(\bs B \bs z + \bs d) \bs B\\
    &=  \bs B^T  \grad^2 f_t(\bs x) \bs B.
\end{align}

From the above two equations we can easily verify that $\grad ^2 \tilde f_t(\bs z) \succcurlyeq \alpha  \grad \tilde f_t(\bs z) \grad \tilde f_t(\bs z) ^T$ as the functions $f_t$ itself are $\alpha$ exp-concave in $\cD$.

Further by Holder's inequality we have $\| \grad \tilde f_t(\bs z)\| \le \| \bs A^{-1} \bs D^{-1}\|_{\text{op}} \| \grad f_t(\bs x)\|_2 \le \| \bs A^{-1} \bs D^{-1}\|_{\text{op}} G$.

\end{proof}

\end{document}